\documentclass{article}

\usepackage{arxiv}

\usepackage[utf8]{inputenc}
\usepackage[T1]{fontenc}

\usepackage[numbers]{natbib}
\usepackage{url}

\usepackage{amsmath}
\usepackage{mathtools}
\usepackage{amssymb}
\usepackage{amsfonts}
\usepackage{amsthm}

\usepackage{float}
\usepackage{graphicx}
\usepackage{booktabs}
\usepackage{multirow}

\usepackage{algorithm}
\usepackage{algorithmic}

\usepackage{nicefrac}
\usepackage{microtype}
\usepackage{xcolor}

\newtheorem{theorem}{Theorem}
\newtheorem{lemma}{Lemma}
\newtheorem{proposition}{Proposition}
\newtheorem{corollary}{Corollary}
\newtheorem{assumption}{Assumption}

\newtheorem{remark}{Remark}
\newtheorem{lbproposition}[proposition]{Proposition}
\newtheorem{applemma}[lemma]{Lemma}
\newcommand{\runinlabel}[1]{\par\emph{#1}\enspace\ignorespaces}
\usepackage{hyperref}
\hypersetup{hidelinks}

\title{Convergence Rates of Gaussian Process Thompson Sampling for Efficient Global Optimization}

\author{%
\makebox[0.45\textwidth]{Haowei Wang\thanks{Equal contribution. These authors are listed in alphabetical order.}}\\
\normalfont National University of Singapore\\
\normalfont Singapore
\And
\makebox[0.45\textwidth]{Yilin Zheng\footnotemark[1]}\\
\normalfont National University of Singapore\\
\normalfont Singapore
\AND
\makebox[0.45\textwidth]{Songhao Wang}\\
\normalfont Southern University of Science and Technology\\
\normalfont China
\And
\makebox[0.45\textwidth]{Szu Hui Ng}\\
\normalfont National University of Singapore\\
\normalfont Singapore
}

\begin{document}
\maketitle

\begin{abstract}
Efficient global optimization aims to identify a global optimum of an expensive black-box function within a limited evaluation budget $N$. We investigate the convergence rates of Gaussian process Thompson sampling (GP-TS) under noiseless observations, assuming a fixed objective function in a reproducing kernel Hilbert space (RKHS). For Matérn kernels with smoothness parameter $\nu$ in dimension $d$, we establish a worst-case expected simple regret upper bound of $O(N^{-\nu/d}\sqrt{\log N})$ for both discrete GP-TS on sufficiently fine grids and its continuous counterpart. For the continuous implementation, we develop a dyadic chaining argument to control posterior sample paths over the entire uncountable input domain. We further establish matching lower bounds of $\Omega(N^{-5/2}\sqrt{\log N})$ for both implementations under a fixed one-dimensional Mat\'ern-$5/2$ kernel, demonstrating that the logarithmic penalty is intrinsic to standard GP-TS in this setting. Motivated by this limitation, we propose Gaussian process capped Thompson sampling (GP-CTS), which caps posterior sample paths using a valid RKHS norm upper bound while preserving optimism. This modification successfully eliminates the logarithmic penalty and achieves the minimax-optimal rate $\Theta(N^{-\nu/d})$ for both discrete and continuous implementations.

\end{abstract}


\section{Introduction}\label{sec:intro}
We consider efficient global optimization of an expensive deterministic black-box function, where the objective is to identify a near-optimal solution using a limited number of function evaluations \citep{JMLR:v12:bull11a}. In such problems, evaluating the objective requires running expensive computer simulations \citep{jones1998efficient,sacks1989design}. Many of these simulations are deterministic, making the observation model noiseless \citep{sacks1989design}. The goal is typically to identify a single high-quality solution as the final recommendation. Poor intermediate evaluations can therefore be acceptable if they help improve this recommendation, whereas cumulative regret penalizes the loss incurred at every evaluation \citep{srinivas2009gaussian}. When observations are noiseless, the best queried point can be identified directly from the observed values. Best-query simple regret is thus a natural performance measure for studying the convergence rates of efficient global optimization algorithms \citep{JMLR:v12:bull11a}.

Bayesian optimization (BO) provides a widely used framework for efficient global optimization by using a surrogate Gaussian process (GP) model to guide sequential evaluations \citep{srinivas2009gaussian,chowdhury2017kernelized,frazier2018tutorial}. Expected improvement (EI) \citep{jones1998efficient,JMLR:v12:bull11a,frazier2018tutorial} is one of the most widely studied acquisition functions. In particular, \citet{JMLR:v12:bull11a} investigated simple regret upper bounds for EI under reproducing kernel Hilbert space (RKHS) assumptions with noiseless observations. These results demonstrate how kernel smoothness affects the convergence rates.

Thompson sampling (TS) is another widely used acquisition strategy, with the associated algorithm called GP-TS \citep{chowdhury2017kernelized}. At each iteration, GP-TS samples from the current GP posterior and selects the maximizer as the next query point. We consider two implementations of GP-TS. The \textit{discrete implementation} \citep{chowdhury2017kernelized} draws a Gaussian random vector over a sufficiently fine discrete finite grid and selects the grid point corresponding to its largest coordinate. In contrast, the \textit{continuous implementation} \citep{wilson2021pathwise, roy2026frequentist} draws a continuous GP posterior sample path over the entire input domain and selects its maximizer. The simplicity and randomized sampling mechanism of GP-TS have made it useful in applications such as parallel BO \citep{kandasamy2018parallelised} and scalable optimization methods such as TuRBO \citep{eriksson2019scalable}.

Despite its practical use, the available simple regret bounds for GP-TS remain comparatively limited. Much of the existing theoretical analysis of GP-TS has focused on cumulative regret due to the convention in multi-armed bandits. Simple regret upper bounds can then be obtained by dividing cumulative regret bounds by the evaluation budget. Although this approach provides valid bounds, the resulting rates can be loose because cumulative regret also accounts for poor intermediate evaluations that do not affect the best queried point. This distinction is particularly important for smooth kernels, as discussed in the related work below. A direct analysis of simple regret is therefore needed to better characterize the convergence rates of GP-TS. This motivates the central question: 
\textit{What convergence rates can GP-TS achieve in efficient global optimization when performance is measured by best-query simple regret, and are these rates minimax-optimal?}

We address this question in the noiseless frequentist RKHS setting, where the objective is a fixed function in a bounded RKHS ball. Under a polynomial Kolmogorov-width condition with decay exponent $\alpha>0$ and regularity assumptions, we establish a general simple regret upper bound of $\widetilde O(N^{-\alpha})$, where $N$ denotes the evaluation budget.  Our analysis combines conditional optimism of posterior samples with a deterministic bound on posterior uncertainty along arbitrary subsequences of queries, building on the approximation geometry of \citet{DBLP:journals/corr/abs-2607-29245}. For the continuous implementation of GP-TS, the main challenge is to control posterior sample paths over the entire uncountable input domain. We address this challenge using a dyadic chaining argument. For Mat\'ern kernels with smoothness parameter $\nu$ in dimension $d$, we have $\alpha=\nu/d$. With fixed kernel parameters and sampling scale, we show that both discrete and continuous implementations achieve the worst-case expected simple regret upper bound $O(N^{-\nu/d}\sqrt{\log N})$ for all $\nu>0$. This bound recovers the full smoothness exponent $\nu/d$ and is near-optimal in terms of the minimax-optimal convergence rates \citep{JMLR:v12:bull11a}.

We next investigate whether the remaining $\sqrt{\log N}$ factor is an artifact of the upper bound analysis or an inherent limitation of GP-TS itself. Establishing a lower bound is challenging because the query points are determined by maxima of correlated Gaussian sample paths, whose behavior changes as observations are collected. To address this challenge, we construct a special family of hard objective functions and analyze Gaussian derivative processes to control sampled values near the true maximizer of the objective, then compare them with Gaussian overshoot over a suboptimal plateau. For a fixed one-dimensional Mat\'ern-$5/2$ kernel, we establish matching worst-case expected lower bounds of $\Omega(N^{-5/2}\sqrt{\log N})$ for both discrete and continuous implementations of GP-TS. The construction shows that Gaussian overshoot over a suboptimal plateau can repeatedly dominate sampled values near the true maximizer. For the discrete implementation, we further show that the discretization error is of smaller order than the lower bound. Therefore, in this setting, the additional $\sqrt{\log N}$ factor reflects the intrinsic behavior of GP-TS itself, instead of the artifact of the upper bound analysis. In particular, the result identifies a logarithmic loss relative to the minimax-optimal convergence rates provided by \citet{JMLR:v12:bull11a}.

This limitation motivates controlling excessive positive Gaussian overshoot in our lower bound analysis, while retaining the optimism used in our upper bound analysis. We thus propose Gaussian process capped Thompson sampling (GP-CTS), which caps each sample path by a deterministic upper envelope obtained from a known valid RKHS norm upper bound. The cap preserves optimism and yields simple regret upper bounds of $O(N^{-\alpha})$ for both discrete and continuous implementations, without a logarithmic factor. For Mat\'ern kernels, GP-CTS thus attains the expected minimax-optimal simple regret bound $\Theta(N^{-\nu/d})$. 
Note that sample clipping also appears in MOTS in finite-armed stochastic bandits \citep{jin2021mots}. Whereas MOTS clips arm-wise samples, we must control overestimation induced by maximizing correlated GP sample paths over a continuous domain. 

Our contributions are summarized as follows:
\begin{itemize}

\item[1).] \textbf{Simple regret upper bounds for GP-TS.} We establish direct simple regret upper bounds for GP-TS. For Matérn kernels with any smoothness parameter $\nu>0$, we obtain a worst-case expected simple regret upper bound of $O(N^{-\nu/d}\sqrt{\log N})$. Compared with previously reported bounds for GP-TS-based methods under noiseless observations, our results improve the logarithmic dependence when $1<\nu\le d$ and the polynomial convergence rate when $\nu>d$. Moreover, our analysis covers not only the commonly studied discrete implementation but also the exact continuous implementation of GP-TS.


\item[2).] \textbf{Matching lower bounds and an intrinsic limitation of GP-TS.}  For a fixed one-dimensional Mat\'ern-$5/2$ kernel, we establish matching lower bounds of $\Omega(N^{-5/2}\sqrt{\log N})$ for both discrete and continuous implementations of GP-TS. Together with our upper bounds, these results establish the tight worst-case expected simple regret bounds $\Theta(N^{-5/2}\sqrt{\log N})$. The lower bound construction identifies Gaussian overshoot over a suboptimal plateau as a mechanism causing the logarithmic loss, even when discretization error is negligible or no finite discrete grid is used.


\item[3).] \textbf{Eliminating the logarithmic penalty through capped posterior sampling. } We propose Gaussian process capped Thompson sampling (GP-CTS), which caps posterior sample paths using a known valid RKHS norm upper bound. The cap preserves optimism while controlling positive sample deviations, yielding an $O(N^{-\nu/d})$ worst-case expected simple regret upper bound for Mat\'ern kernels. Together with the minimax lower bound of \citet{JMLR:v12:bull11a}, this establishes the minimax-optimal convergence rates $\Theta(N^{-\nu/d})$ for both discrete and continuous implementations of GP-CTS. 


\end{itemize}

\subsection{Related Work}\label{subsec:related-work}

\textbf{Convergence rates in efficient global optimization.}
 \citet{JMLR:v12:bull11a} considered noiseless observations and established simple regret upper bounds for EI under fixed GP models and showed that a modified EI algorithm can attain the optimal polynomial convergence rate for Mat\'ern kernels. More recently, \citet{DBLP:journals/corr/abs-2607-29245} established the optimal simple regret upper bound $O(N^{-\nu/d})$ for fixed-prior EI under Mat\'ern RKHS assumptions. Other GP based algorithms, including maximum variance reduction and phased elimination, have also been analyzed in the noiseless setting \citep{iwazaki2025improved}. These results provide important benchmarks for studying the convergence of GP-TS. However, convergence guarantees for other acquisition strategies do not determine the rates achieved by GP-TS itself.

\begin{table}[H]
\centering
\small
\setlength{\tabcolsep}{3.5pt}
\begin{tabular}{@{}llcc@{}}
\toprule
 Simple regret
 & Mat\'ern kernel & \citet{takeno2026improved} & Ours \\
\midrule
Upper bound
 & $1<\nu<d$
 & $O(N^{-\nu/d}(\log N)^{\nu/d+1/2})$
 & $O(N^{-\nu/d}\sqrt{\log N})$ \\

 & $\nu=d>1$
 & $O(N^{-1}(\log N)^{5/2})$
 & $O(N^{-1}\sqrt{\log N})$ \\

 & $\nu>d$
 & $O(N^{-1}\sqrt{\log N})$
 & $O(N^{-\nu/d}\sqrt{\log N})$ \\
\midrule
Lower bound
 & $\nu=5/2,\ d=1$
 & ---
 & $\Omega(N^{-5/2}\sqrt{\log N})$ \\
\bottomrule
\end{tabular}
\caption{Comparison of worst-case expected simple regret bounds under noiseless observations. The previous bounds are for GP-BTS with fixed delay $Q$ and finite decision sets, with the logarithmic-factor qualification discussed
below. The last column reports our bounds for standard GP-TS.}
\label{tab:gp-ts-regret-comparison}
\end{table}

\textbf{Frequentist regret analysis of GP-TS.}
\citet{chowdhury2017kernelized} established cumulative regret upper bounds for GP-TS in the frequentist RKHS setting with noisy observations. Their analysis uses posterior sampling over finite decision sets and bounds cumulative regret through maximum information gain. More recently, \citet{takeno2026improved} studied GP Batched Thompson Sampling (GP-BTS) under both noisy and noiseless observations. In the noiseless setting, their simple regret upper bounds are obtained by dividing cumulative regret bounds by the evaluation budget. Among the available frequentist analyses of GP-TS based methods, their noiseless results provide a particularly relevant comparison with our work. Table~\ref{tab:gp-ts-regret-comparison} compares our simple regret bounds with those reported by \citet{takeno2026improved} for Mat\'ern kernels under noiseless observations. It is clear that our upper bounds improve the reported logarithmic factors when $1<\nu\le d$ and the polynomial convergence rate when $\nu>d$. Furthermore, we establish a matching lower bound for the fixed one-dimensional Mat\'ern-$5/2$ kernel, which is not provided in the previous work. The logarithmic factors reported by \citet{takeno2026improved} are subject to the qualification discussed in their Remark~2.2. \citet{roy2026frequentist} developed a frequentist analysis of GP-TS over continuous domains, focusing on cumulative regret under noisy observations. Their analysis relies on fractional posterior contraction and information gain. In contrast, we study best-query simple regret under noiseless observations for both discrete and continuous implementations, with our upper bound analysis based on conditional optimism and RKHS approximation geometry.

\section{Background}\label{sec:background}

We consider black-box optimization problems with noiseless observations over a compact domain $\mathcal X\subset\mathbb R^d$. For simplicity, throughout the paper we take $\mathcal X=[0,1]^d$. At round $n$, a sequential algorithm selects a query point $x_n\in\mathcal X$ and observes $y_n=f(x_n)$, $n=1,2,\ldots,N$, where $f:\mathcal X\to\mathbb R$ is an unknown but fixed black-box objective function. Without loss of generality, we consider maximization and write $f^*=\max_{x\in\mathcal X}f(x)$ and $x^*\in\operatorname*{arg\,max}_{x\in\mathcal X}f(x)$, where one maximizer $x^*$ is fixed if the maximizer is not unique. Our primary performance metric is the best-query simple regret
\begin{equation}\label{eq:simple-regret}
r_N^{\mathrm{best}}
=
f^*-\max_{1\le n\le N}f(x_n).
\end{equation}
Since all observations are exact, the best queried point can be identified directly from the observed function values. All probabilities and expectations in our theoretical analysis are taken with respect to the randomness of GP posterior sampling, for a fixed objective $f$.

\subsection{GP Posterior and its RKHS Geometry}
We use a zero-mean GP prior $\mathcal{GP}(0,k(\cdot,\cdot))$, where $k:\mathcal{X}\times\mathcal{X}\to \mathbb{R}$ is the kernel function. All kernel parameters remain fixed, and no positive nugget is added to the Gram matrix. For $X_n=(x_1,\dots,x_n)$, define $K_n=[k(x_i,x_j)]_{i,j=1}^n$, $\mathbf k_n(x)=(k(x_1,x),\ldots,k(x_n,x))^\top$, and $\mathbf y_n=(f(x_1),\ldots,f(x_n))^\top$. The well-known GP posterior formulas follow from Gaussian conditioning \citep{DBLP:books/lib/RasmussenW06}:
\begin{equation}\label{eq:gp-posterior}
\begin{aligned}
\mu_n(x)&=\mathbf k_n(x)^\top K_n^\dagger\mathbf y_n,\\
c_n(x,x')&=k(x,x')-\mathbf k_n(x)^\top K_n^\dagger\mathbf k_n(x'), \qquad s_n(x)=\sqrt{c_n(x,x)}.
\end{aligned}
\end{equation}
Here $K_n^\dagger$ denotes the Moore-Penrose pseudo-inverse, which equals $K_n^{-1}$ when $K_n$ is non-singular. We use the pseudo-inverse because GP-TS may query the same location more than once in our noiseless setting, which makes the Gram matrix $K_n$ singular.

For $f\in\mathcal H_k$, these posterior quantities also admit useful geometric interpretations in the RKHS $\mathcal{H}_k$ associated with kernel $k$. Let $k_x:=k(x,\cdot)$, $W_n:=\operatorname{span}\{k_{x_1},\ldots,k_{x_n}\}$, let $P_n:\mathcal H_k\to W_n$ denote the orthogonal projection operator, and let $I$ be the identity operator on $\mathcal H_k$. Then
\begin{equation}\label{eq:rkhs-geometry}
\mu_n=P_nf,\qquad
c_n(x,x')=\langle (I-P_n)k_x,(I-P_n)k_{x'}\rangle_{\mathcal H_k},\qquad
s_n(x)=\|(I-P_n)k_x\|_{\mathcal H_k}.
\end{equation}
In particular, once the RKHS norm of $f$ is bounded by $B>0$, the following deterministic bound holds for all $x\in \mathcal{X}$:
\begin{equation}\label{eq:deterministic-upper-bound}
|f(x)-\mu_n(x)|\le B s_n(x).
\end{equation}
The residual vector at round $n$, i.e., $(I-P_{n-1})k_{x_n}$, is also called the sequential {innovation} \citep{DBLP:journals/corr/abs-2607-29245} of the newly queried kernel section $k_{x_n}$, and its RKHS norm is exactly $s_{n-1}(x_n)$. For $n=0$, set $\mu_0=0$, $c_0=k$, $s_0(x)=\sqrt{k(x,x)}$, $W_0=\{0\}$, and $P_0=0$. We refer readers to Appendix~\ref{app:background} for the corresponding derivations, including singular Gram matrices.

\subsection{Kernel Approximation Geometry}\label{subsec:kernel-approximation}
Define $\mathcal K_k=\{k_x:x\in\mathcal X\}\subset\mathcal H_k$. For an integer $m\ge1$, the Kolmogorov $m$-width of $\mathcal K_k$ is defined as
\begin{equation}\label{eq:kolmogorov-width}
w_m(k,\mathcal X)
:=
\inf_{\substack{V\subset\mathcal H_k\text{ linear}\\\dim V\le m}}
\sup_{x\in\mathcal X}\operatorname{dist}_{\mathcal H_k}(k_x,V),
\end{equation}
which measures the smallest worst-case error achievable when the entire family of kernel sections is approximated by an $m$-dimensional linear subspace $V\subset\mathcal H_k$. We further define the canonical kernel pseudo-metric
\begin{equation}\label{eq:kernel-pseudometric}
d_k(x,x')
:=\|k_x-k_{x'}\|_{\mathcal H_k}
=\sqrt{k(x,x)+k(x',x')-2k(x,x')},
\end{equation}
which characterizes the local variation of kernel sections with respect to the input.

\subsection{Regularity Assumptions}
We state the following regularity assumptions. Each theoretical result specifies which of them is required.

\begin{assumption}\label{asmp:kernel-model}
The kernel \(k\) is continuous, symmetric, and positive semidefinite on \(\mathcal X\times\mathcal X\), and satisfies $\sup_{x\in\mathcal X}k(x,x)\le1$. The objective function satisfies $f\in\mathcal H_k$ and $\|f\|_{\mathcal H_k}\le B$ for some fixed $B>0$.
\end{assumption}
The kernel and $B$ are fixed independently of the evaluation budget $N$. This assumption implies that $f$ is continuous, the maximum $f^*$ exists, and $|f(x)|\le B$ as well as $0\le r_N^{\mathrm{best}}\le2B$.

\begin{assumption}\label{asmp:poly-kolmogorov-width}
There exist constants $A>0$ and $\alpha>0$ such that
\begin{equation}
    w_m(k,\mathcal{X})\le Am^{-\alpha},\quad m\ge 1.
\end{equation}
\end{assumption}
Assumption~\ref{asmp:poly-kolmogorov-width} characterizes the approximation complexity of the kernel section family $\mathcal K_k$. It does not require any space-filling, separation, or quasi-uniformity condition on the actual query points, and the approximating subspace $V$ need not be spanned by previously queried kernel sections. 

\begin{assumption}\label{asmp:canonical-metric}
There exist constants $L_k>0$ and $\zeta \in(0,1]$ such that
\begin{equation}
    d_k\left(x, x^{\prime}\right) \leq L_k\left\|x-x^{\prime}\right\|_2^\zeta, \quad x, x^{\prime} \in \mathcal{X} .
\end{equation}
\end{assumption}
Assumption~\ref{asmp:canonical-metric} is used to control finite-grid approximation errors and increments of continuous sample paths. Together with Assumption~\ref{asmp:kernel-model}, it implies
\begin{equation}\label{eq:Holder-control}
    |f(x)-f(x')|\le BL_k\|x-x'\|_2^\zeta.
\end{equation}

\begin{proposition}\label{proposition:Matern-kernel-asmps}
Let $k$ be a stationary isotropic Matérn kernel with fixed lengthscale $\ell>0$ and smoothness parameter $\nu>0$, restricted to $\mathcal{X}=[0,1]^d$ and normalized such that $k(x,x)\le 1,\ \forall x\in\mathcal{X}$. Then Assumption~\ref{asmp:poly-kolmogorov-width} holds with $\alpha=\nu/d$, and Assumption~\ref{asmp:canonical-metric} holds for any $0<\zeta<\min\{\nu,1\}$ and one may take $\zeta=1$ whenever $\nu>1$.  
\end{proposition}
Proposition~\ref{proposition:Matern-kernel-asmps} follows from classical estimates for kernel interpolation and from the spectral density of the Mat\'ern kernel, which satisfies $\rho(\omega)\asymp (1+\|\omega\|_2^2)^{-(\nu+d/2)}$. The proof is given in Appendix~\ref{app:prop-matern}.

\subsection{Gaussian Process Thompson Sampling}\label{subsec:gp-ts}
Let $\mathcal{F}_{n-1}$ denote the history available before round $n$, including all previous query points, exact observations, and algorithmic randomness. We distinguish two implementations of GP-TS according to the decision domain $\mathcal D$, which is fixed within a run. All GP posterior draws use fresh randomization and have the stated conditional law given the complete history. The sampling scale $v>0$ is fixed throughout the theoretical analysis, and $v=1$ corresponds to the unscaled working posterior.

\runinlabel{Discrete Implementation}
If $\mathcal{D}=\mathcal{G}=\{z_1,\ldots,z_M\}\subset\mathcal X$ is a nonempty deterministic finite set, GP-TS draws a finite-dimensional Gaussian random vector
\begin{equation}\label{eq:TS-discrete-sample}
\left(G_n(z_1),\ldots,G_n(z_M)\right)^\top\mid\mathcal F_{n-1}
\sim\mathcal N\left(\mu_{n-1}(\mathcal G),v^2c_{n-1}(\mathcal G,\mathcal G)\right)
\end{equation}
and selects its maximizing coordinate as the next query point. No continuous sample path is required. On $\mathcal G$, write $Z_n=(G_n-\mu_{n-1})/v$ for the corresponding centered residual vector.

\runinlabel{Continuous Implementation}
If $\mathcal D=\mathcal X$, GP-TS draws at round $n$ an exact continuous posterior sample path
\begin{equation}\label{eq:TS-continuous-sample}
Z_n\mid\mathcal F_{n-1}\sim\mathcal{GP}(0,c_{n-1}),
\qquad
G_n(x)=\mu_{n-1}(x)+vZ_n(x),\quad x\in\mathcal X,
\end{equation}
and selects a global maximizer of $G_n$.

To account for possible numerical errors when maximizing $G_n$, we allow a deterministic global optimization tolerance $\eta_n\ge0$ for either implementation and select a measurable point $x_n\in\mathcal D$ satisfying
\begin{equation}\label{eq:TS-selection}
G_n(x_n)\ge\sup_{x\in\mathcal D}G_n(x)-\eta_n.
\end{equation}
Exact maximization corresponds to $\eta_n=0$, which we always use for the discrete implementation because the sampled Gaussian vector can be maximized exactly. We use fixed measurable tie-breaking rules, and their existence for continuous sample paths on the compact domain is verified in Appendix~\ref{app:background}. Algorithm~\ref{alg:gp-ts} combines the two implementations in one procedure. For either implementation, $N$ counts the total evaluation budget.

\begin{algorithm}[tbp]
\caption{Gaussian-Process Thompson Sampling (GP-TS)}\label{alg:gp-ts}
\begin{algorithmic}[1]
\REQUIRE Domain $\mathcal X$; decision domain $\mathcal D=\mathcal G$ or $\mathcal X$; kernel $k$; fixed $v>0$; budget $N\ge1$.
\ENSURE A best queried point $\widehat x_N$.
\STATE Initialize $X_0=()$ and $\mathbf y_0=()$. Set $\mu_0=0$ and $c_0=k$.
\FOR{$n=1,\ldots,N$}
    \STATE Compute $(\mu_{n-1},c_{n-1},s_{n-1})$ from $(X_{n-1},\mathbf y_{n-1})$ using \eqref{eq:gp-posterior}.
    \IF{$\mathcal D=\mathcal G=\{z_1,\ldots,z_M\}$}
        \STATE Draw $g_n\sim\mathcal N(\mu_{n-1}(\mathcal G),v^2c_{n-1}(\mathcal G,\mathcal G))$ conditionally on $\mathcal F_{n-1}$.
        \STATE Set $G_n(z_j)=(g_n)_j$ for $j=1,\ldots,M$.
    \ELSE
        \STATE Draw a fresh continuous $Z_n\mid\mathcal F_{n-1}\sim\mathcal{GP}(0,c_{n-1})$.
        \STATE Set $G_n=\mu_{n-1}+vZ_n$ on $\mathcal X$.
    \ENDIF
    \STATE Select a measurable $x_n\in\mathcal D$ satisfying $G_n(x_n)\ge\sup_{x\in\mathcal D}G_n(x)-\eta_n$.
    \STATE Observe $y_n=f(x_n)$ and append $x_n$ to $X_{n-1}$ and $y_n$ to $\mathbf y_{n-1}$ to obtain $X_n,\mathbf y_n$.
\ENDFOR
\STATE Choose $j_N\in\operatorname*{arg\,max}_{1\le j\le N}y_j$ using a fixed tie-breaking rule.
\RETURN $\widehat x_N=x_{j_N}$.
\end{algorithmic}
\end{algorithm}

\section{Simple Regret Upper Bound of GP-TS}\label{sec:GP-TS-simple-regret}
In this section, we derive simple regret upper bounds for both the discrete and continuous implementations of GP-TS.

\subsection{Two Key Lemmas}
We first introduce two key lemmas that characterize, respectively, the decay of posterior uncertainty along arbitrary query subsequences and the frequency of optimistic posterior samples. The first lemma restates a consequence of the width-to-separation geometry of \citet{DBLP:journals/corr/abs-2607-29245}.

\begin{lemma}[Arbitrary subsequence innovation bound]\label{lemma:deterministic-bound-arbitrary-subseq}
Under Assumptions~\ref{asmp:kernel-model} and~\ref{asmp:poly-kolmogorov-width}, there is a constant $C_{\mathrm{sep}}>0$ depending only on constants $A$ and $\alpha$, such that for every $N\ge1$, every sequence $x_1,\ldots,x_N\in\mathcal X$, and every nonempty index set $I\subseteq\{1,\ldots,N\}$ with $|I|=m$,
\begin{equation}\label{eq:subsequence-innovation}
    \min_{n\in I}s_{n-1}(x_n)\le C_{\mathrm{sep}}m^{-\alpha}.
\end{equation}
One admissible constant is
\begin{equation}\label{eq:C-sep}
    C_{\mathrm{sep}}=e^{(1+2\alpha)/2}\sqrt{\max\{1,2A^2 3^{2\alpha}\}}.
\end{equation}

Consequently, for every $u>0$, we also have the following counting result
\begin{equation}\label{eq:innovation-count}
    \#\{n\le N:s_{n-1}(x_n)>u\}\le\left\lceil\left(\frac{C_{\mathrm{sep}}}{u}\right)^{1/\alpha}\right\rceil.
\end{equation}
\end{lemma}

Lemma~\ref{lemma:deterministic-bound-arbitrary-subseq} is deterministic and holds simultaneously for every subset of a realized algorithmic trajectory. Thus $I$ could be defined after all posterior samples $G_1,\ldots,G_N$ have been drawn. Its proof is given in Appendix~\ref{app:lemma-innovation}.

\begin{lemma}[Conditional lower-tail bound]\label{lemma:conditional-success-prob}
Let $(\mathcal F_n)$ be a filtration and let $O_n\in\{0,1\}$ be an $\mathcal F_n$-measurable indicator. If
\begin{equation}\label{eq:lemma2-if}
\mathbb P(O_n=1\mid\mathcal F_{n-1})\ge p \quad\text{a.s.}, \quad 0<p\le1,
\end{equation}
then, for every $N\ge1$,
\begin{equation}\label{eq:conditional-count}
    \mathbb P\left(\sum_{n=1}^N O_n<\frac{pN}{2}\right) \le e^{-pN/8}.
\end{equation}
\end{lemma}

Lemma~\ref{lemma:conditional-success-prob} follows from a conditional version of the standard Chernoff argument, which requires no independence across rounds, and its proof is given in Appendix~\ref{app:lemma-conditional}. For the analysis of upper bounds, we fix a deterministic comparator $x^\dagger\in\mathcal{X}$, taken as $x_{\mathcal G}^*$ in the discrete case and $x^*$ in the continuous case, and define the indicator $O_n=\mathbf{1}\{G_n(x^\dagger)\ge f(x^\dagger)\}$. Assumption~\ref{asmp:kernel-model}, together with the fresh conditional Gaussian sampling rule in Section~\ref{subsec:gp-ts}, implies \eqref{eq:lemma2-if} with $p=\Phi(-B/v)$, where $\Phi$ is the standard normal distribution function, as verified in \eqref{eq:app-optimism-prob}. Choosing $v=B$ makes $p$ the numerical constant $\Phi(-1)$. With $v=1$, $p$ remains strictly positive for every fixed RKHS ball, although it can be small when $B$ is large.

\subsection{Simple Regret Upper Bound for the Discrete Implementation}
Fix a nonempty deterministic finite set $\mathcal G\subset\mathcal X$ of cardinality $M$. Define

\begin{equation}\label{eq:discrete-comparator}
f_{\mathcal G}^*=\max_{x\in\mathcal G}f(x),
\quad
x_{\mathcal G}^*\in\operatorname*{arg\,max}_{x\in\mathcal G}f(x),
\quad
b_{\mathcal G}=f^*-f_{\mathcal G}^*.
\end{equation}

The bias $b_{\mathcal G}$ measures discretization of the continuous objective domain into a finite decision set.

\begin{theorem}[General upper bound for discrete implementation]\label{thm:simple-regret-discrete-general}
\leavevmode\par\nobreak\noindent
Under Assumptions~\ref{asmp:kernel-model} and~\ref{asmp:poly-kolmogorov-width}, run Algorithm~\ref{alg:gp-ts} on $\mathcal G$ for $N$ rounds with fixed $v>0$, fresh exact posterior draws, and the global selection rule \eqref{eq:TS-selection}. For $0<\delta<1$, define $p=\Phi(-B/v)$ and the Gaussian envelope coefficient $u_{N,M,\delta}:=\sqrt{2\log({4MN}/{\delta})}$ on $\mathcal{G}$. If $N\ge8\log(2/\delta)/p$, then, for every fixed $f$ in the radius-$B$ RKHS ball, with probability at least $1-\delta$,
    \begin{equation}\label{eq:ts-discrete-bound}
        r_N^{\mathrm{best}}\le b_{\mathcal G}+C_{\mathrm{sep}}(B+vu_{N,M,\delta}) \left(\frac{pN}{2}\right)^{-\alpha}+\max_{1\le n\le N}\eta_n.
    \end{equation}
\end{theorem}

\begin{proof}
Let $I_N=\{n\le N:G_n(x_{\mathcal G}^*)\ge f_{\mathcal G}^*\}$ be the set of optimistic rounds. For every $n\in I_N$, optimism at $x_{\mathcal G}^*$ and the selection rule \eqref{eq:TS-selection} give
\begin{equation}\label{eq:ts-one-round}
f_{\mathcal G}^*-f(x_n)
\le G_n(x_{\mathcal G}^*)-f(x_n)
\le G_n(x_n)-f(x_n)+\eta_n.
\end{equation}
The remaining positive sample deviation $G_n(x_n)-f(x_n)$ is evaluated at the query $x_n$, which depends on the current sample $G_n$ itself, so the Gaussian law at a fixed point cannot be applied directly. The finite decision set instead allows a uniform Gaussian envelope over all rounds and candidate points. Write $u=u_{N,M,\delta}$. Conditionally on $\mathcal F_{n-1}$, each $Z_n(x)$ with $x\in\mathcal G$ is a centered Gaussian variable with standard deviation $s_{n-1}(x)$, so the Gaussian tail bound gives $\mathbb P(|Z_n(x)|>us_{n-1}(x)\mid\mathcal F_{n-1})\le2e^{-u^2/2}$. A union bound over the $NM$ pairs $(n,x)$ shows that the envelope event $\mathcal E_{\mathrm{env}}=\{|Z_n(x)|\le us_{n-1}(x)\text{ for all }1\le n\le N\text{ and }x\in\mathcal G\}$ satisfies $\mathbb P_f(\mathcal E_{\mathrm{env}}^c)\le2NMe^{-u^2/2}=\delta/2$. On $\mathcal E_{\mathrm{env}}$, the bound \eqref{eq:deterministic-upper-bound} gives, for all $n\le N$ and $x\in\mathcal G$,
\begin{equation}\label{eq:discrete-envelope}
|G_n(x)-f(x)|\le|\mu_{n-1}(x)-f(x)|+v|Z_n(x)|\le(B+vu)s_{n-1}(x).
\end{equation}
This bound may be evaluated at the random point $x_n$ because it holds simultaneously on $\mathcal G$. By \eqref{eq:app-optimism-prob}, Lemma~\ref{lemma:conditional-success-prob}, and the budget condition, $\mathbb P_f(|I_N|<pN/2)\le e^{-pN/8}\le\delta/2$. The event $\mathcal E_{\mathrm{env}}\cap\{|I_N|\ge pN/2\}$ therefore has probability at least $1-\delta$. On this event, Lemma~\ref{lemma:deterministic-bound-arbitrary-subseq}, applied to $I_N$, provides a round $n_\circ\in I_N$ with $s_{n_\circ-1}(x_{n_\circ})\le C_{\mathrm{sep}}|I_N|^{-\alpha}\le C_{\mathrm{sep}}(pN/2)^{-\alpha}$. Since $r_N^{\mathrm{best}}\le b_{\mathcal G}+f_{\mathcal G}^*-f(x_{n_\circ})$, the bounds \eqref{eq:ts-one-round} and \eqref{eq:discrete-envelope} at $n=n_\circ$ and $x=x_{n_\circ}$ prove \eqref{eq:ts-discrete-bound}.
\end{proof}

Next we provide the Mat\'ern kernel specialization of Theorem~\ref{thm:simple-regret-discrete-general}. To control the discretization bias $b_{\mathcal{G}}$, we define the regular grid
\begin{equation}\label{eq:regular-grid}
    \mathcal Q_J=\{0,1/J,\ldots,1\}^d,\quad M_J=(J+1)^d,\quad h_J=\frac{\sqrt d}{J},
\end{equation}
which means that each coordinate axis is uniformly partitioned into $J$ subintervals.

Under Assumption~\ref{asmp:canonical-metric} and \eqref{eq:Holder-control}, we have $b_{\mathcal Q_J}\le BL_kh_J^\zeta$. For a given budget $N$, choose
\begin{equation}\label{eq:N-grid}
    J_N=\left\lceil\sqrt d\,N^{\alpha/\zeta}\right\rceil,\quad \mathcal G_N=\mathcal Q_{J_N}.
\end{equation}
Then
\begin{equation}\label{eq:grid-bias-size}
    b_{\mathcal G_N}\le BL_kN^{-\alpha},\quad|\mathcal G_N|\le C_dN^{\alpha d/\zeta},
\end{equation}
where one may take $C_d=(\sqrt d+2)^d$.

\begin{corollary}[Discrete Mat\'ern rate]\label{corollary:Matern-discrete}
For the Mat\'ern kernel in Proposition~\ref{proposition:Matern-kernel-asmps}, run Algorithm~\ref{alg:gp-ts} on the grid in \eqref{eq:N-grid} with $\alpha=\nu/d$, any fixed $v>0$, and exact maximization ($\eta_n=0$). Set $p=\Phi(-B/v)$. There exists a constant $C>0$, independent of $f$ and $\delta$, such that, for every fixed $f$ in the radius-$B$ RKHS ball and every $0<\delta<1$, whenever $N\ge2$ and $N\ge8\log(2/\delta)/p$,
\begin{equation}\label{eq:matern-discrete-hp}
\mathbb P_f\left(r_N^{\mathrm{best}}\le C N^{-\nu/d}\sqrt{\log(N/\delta)}\right)\ge1-\delta.
\end{equation}
Moreover, there exist constants $C'>0$ and an integer $N_0\ge2$, independent of the particular $f$, such that for all $N\ge N_0$,
\begin{equation}\label{eq:matern-discrete-expected}
\sup_{\|f\|_{\mathcal H_k}\le B}\mathbb E_f[r_N^{\mathrm{best}}]\le C' N^{-\nu/d}\sqrt{\log N}.
\end{equation}
\end{corollary}

The constants $C,C',N_0$ in Corollary~\ref{corollary:Matern-discrete} may depend on $B,v,k,d$ and the chosen $\zeta$. The proof is given in Appendix~\ref{app:cor-matern-discrete}. According to \citet{JMLR:v12:bull11a}, the minimax-optimal convergence rate over a fixed Mat\'ern RKHS ball under noiseless evaluations is $N^{-\nu/d}$, and Appendix~\ref{app:cor-matern-discrete} justifies the comparison under the best-query loss. Corollary~\ref{corollary:Matern-discrete} therefore matches the optimal polynomial exponent and is near-optimal up to a $\sqrt{\log N}$ factor.

\subsection{Simple Regret Upper Bound for the Continuous Implementation}
We now consider the continuous implementation of GP-TS with $\mathcal D=\mathcal X$. Unlike in the discrete case, the maximizer $x_n$ is now selected from an uncountable input domain using the current posterior sample path, so the uniform Gaussian envelope on a finite set used in Theorem~\ref{thm:simple-regret-discrete-general} is no longer directly applicable. We instead establish a uniform Gaussian envelope for the posterior residual process over the entire domain. We still use the regular grid defined in \eqref{eq:regular-grid}. Define
\begin{gather}
    u_{N, J, \delta}^{\mathrm{c}}=\sqrt{2 \log \frac{8 N M_J}{\delta}},\qquad \rho_\zeta=2^{-\zeta},\label{eq:continuous-envelope-coefficient}\\
    C_1(\zeta)=\frac1{1-\rho_\zeta},\qquad C_2(d,\zeta)=\sqrt{2(d+1)\log2}\frac{\rho_\zeta}{(1-\rho_\zeta)^2}.\label{eq:continuous-envelope-constants}
\end{gather}

\begin{lemma}[Continuous residual Gaussian envelope]\label{lemma:unif-envelope-continuous}
    Under Assumptions~\ref{asmp:kernel-model} and~\ref{asmp:canonical-metric}, let $Z_n$ be fresh continuous posterior residuals with the conditional law \eqref{eq:TS-continuous-sample}. For any integers $J,N\ge1$ and $0<\delta<1$, with probability at least $1-\delta/2$, simultaneously for all $1\le n\le N$ and $x\in\mathcal X$, the following uniform Gaussian envelope holds
    \begin{equation}\label{eq:continuous-envelope}
        |Z_n(x)|\le u_{N,J,\delta}^{\mathrm c}s_{n-1}(x)+L_kh_J^\zeta\left[C_1(\zeta)u_{N,J,\delta}^{\mathrm{c}}+C_2(d,\zeta)\right].
    \end{equation}
\end{lemma}
\runinlabel{Proof sketch}
The proof in Appendix~\ref{app:lemma-continuous-envelope} is a dyadic chaining argument over nested refinements of the grid $\mathcal Q_J$, which enter only the analysis and do not constrain the queries. Union bounds on the coarse grid and on the increments between successive refinements, whose variances are controlled by Assumption~\ref{asmp:canonical-metric}, are extended to every $x\in\mathcal X$ by path continuity, without a union bound over an uncountable set.

Using this Gaussian envelope, an argument similar to the discrete case gives the following result.

\begin{theorem}[General upper bound for continuous implementation]\label{thm:simple-regret-continuous-general}
\leavevmode\par\nobreak\noindent
    Under Assumptions~\ref{asmp:kernel-model}, \ref{asmp:poly-kolmogorov-width}, and~\ref{asmp:canonical-metric}, run Algorithm~\ref{alg:gp-ts} on $\mathcal X$ for $N$ rounds with fixed $v>0$, fresh exact continuous posterior draws, and the global selection rule \eqref{eq:TS-selection}. For $0<\delta<1$, set $p=\Phi(-B/v)$. If $N\ge8\log(2/\delta)/p$, then, for each integer $J\ge1$ and each fixed $f$ in the radius-$B$ RKHS ball, with probability at least $1-\delta$,
    \begin{equation}\label{eq:ts-continuous-bound}
            r_N^{\mathrm{best}}\le
            C_{\mathrm{sep}}(B+vu_{N,J,\delta}^{\mathrm c})
            \left(\frac{pN}{2}\right)^{-\alpha}
            +vL_kh_J^\zeta
            \left[C_1(\zeta)u_{N,J,\delta}^{\mathrm c}+C_2(d,\zeta)\right]
            +\max_{1\le n\le N}\eta_n.
    \end{equation}
\end{theorem}
\begin{proof}
Fix $J\ge1$ and abbreviate $u=u_{N,J,\delta}^{\mathrm c}$ and $D_J=L_kh_J^\zeta[C_1(\zeta)u+C_2(d,\zeta)]$. Let $I_N=\{n\le N:G_n(x^*)\ge f^*\}$. As in the proof of Theorem~\ref{thm:simple-regret-discrete-general}, $\mathbb P_f(|I_N|<pN/2)\le e^{-pN/8}\le\delta/2$. The intersection of $\{|I_N|\ge pN/2\}$ with the event of Lemma~\ref{lemma:unif-envelope-continuous} therefore has probability at least $1-\delta$. On this intersection, for every $n\in I_N$, optimism at $x^*$, the selection rule \eqref{eq:TS-selection}, the bound \eqref{eq:deterministic-upper-bound}, and the envelope \eqref{eq:continuous-envelope} give
\begin{equation}\label{eq:continuous-one-round}
f^*-f(x_n)\le G_n(x_n)-f(x_n)+\eta_n\le(B+vu)s_{n-1}(x_n)+vD_J+\eta_n.
\end{equation}
The envelope applies at the random point $x_n$ because it holds simultaneously on $\mathcal X$, which produces the additional chaining term $vD_J$. Lemma~\ref{lemma:deterministic-bound-arbitrary-subseq} provides a round $n_\circ\in I_N$ with $s_{n_\circ-1}(x_{n_\circ})\le C_{\mathrm{sep}}(pN/2)^{-\alpha}$, and \eqref{eq:continuous-one-round} at $n=n_\circ$ together with $r_N^{\mathrm{best}}\le f^*-f(x_{n_\circ})$ proves \eqref{eq:ts-continuous-bound}.
\end{proof}

Similar to the discrete case, we can derive the Mat\'ern kernel specialization of Theorem~\ref{thm:simple-regret-continuous-general}.
\begin{corollary}[Continuous Mat\'ern rate]\label{corollary:Matern-continuous}
   For the Mat\'ern kernel in Proposition~\ref{proposition:Matern-kernel-asmps}, run Algorithm~\ref{alg:gp-ts} on $\mathcal X$ with any fixed $v>0$, fresh exact continuous posterior draws, and exact maximization ($\eta_n=0$). The high-probability and worst-case expected bounds in Corollary~\ref{corollary:Matern-discrete} hold under the same respective budget conditions, with possibly different constants.
\end{corollary}

The proof of Corollary~\ref{corollary:Matern-continuous} is given in Appendix~\ref{app:cor-matern-continuous}. 


\begin{remark}[Source of the logarithmic factor]\label{subsec:upper-summary}
For Mat\'ern kernels, both the continuous and sufficiently fine discrete implementations achieve worst-case expected simple regret upper bounds that match the minimax-optimal rate up to a $\sqrt{\log N}$ factor. In both analyses, this factor enters only through the Gaussian envelope coefficient $u_{N,M,\delta}$ or $u_{N,J,\delta}^{\mathrm c}$, which controls the positive sample deviation $G_n(x_n)-f(x_n)$ at the selected query in \eqref{eq:ts-one-round}. Section~\ref{sec:log-factor-GP-TS} shows that this factor is necessary for a fixed one-dimensional Mat\'ern-$5/2$ kernel, and Section~\ref{sec:GP-CTS-simple-regret} removes it by capping this positive deviation.
\end{remark}

\section{Simple Regret Lower Bound of GP-TS}\label{sec:log-factor-GP-TS}

The upper bounds in Section~\ref{sec:GP-TS-simple-regret} leave open whether the factor $\sqrt{\log N}$ is necessary. This section shows that it is. For a fixed one-dimensional Mat\'ern-$5/2$ kernel, we construct objectives in the RKHS ball of radius $B$ on which both the discrete and the continuous implementations of GP-TS incur a simple regret of order $N^{-5/2}\sqrt{\log N}$ with probability bounded away from zero.

The lower bound comes from a competition between two regions of the domain. Each objective has a narrow peak close to the left boundary and vanishes on a plateau at the right end of the domain. On the plateau, a posterior sample takes its largest value among many Gaussian variables that are not too strongly correlated, and in round $n$ this largest value exceeds the posterior standard deviation by a factor of order $\sqrt{\log n}$. Near the peak, in contrast, the sample paths are twice continuously differentiable and are pinned at zero by an initial observation at the boundary, which keeps the sampled values small. When the height of the peak is comparable to the Gaussian overshoot on the plateau, the maximizer of every sample stays away from the peak, and no query approaches the true maximizer within the budget. All proofs are given in Appendix~\ref{app:log-factor}.

\subsection{Kernel, Decision Domains, and Hard Objectives}\label{subsec:discrete-lower-setup}
\runinlabel{Kernel}
Fix $B,v>0$ and consider
\begin{equation}\label{eq:lower-matern-kernel}
k(x,y)=\left(1+|x-y|+\frac{|x-y|^2}{3}\right)e^{-|x-y|},
\qquad x,y\in[0,1].
\end{equation}
This is the normalized Mat\'ern-$5/2$ kernel in dimension one with fixed lengthscale $\sqrt5$. It satisfies Assumption~\ref{asmp:kernel-model} with $k(x,x)=1$ and, by Proposition~\ref{proposition:Matern-kernel-asmps}, Assumption~\ref{asmp:poly-kolmogorov-width} with $\alpha=5/2$ and Assumption~\ref{asmp:canonical-metric} with $\zeta=1$.

\runinlabel{Decision domains}
The continuous implementation maximizes each sample over $[0,1]$. For the discrete implementation, let $(J_N^{\mathrm{lb}})$ be a deterministic sequence of positive integers and consider the regular decision grids and the resolution regime
\begin{gather}
\gamma_N=(J_N^{\mathrm{lb}})^{-1},\qquad
\mathcal G_N^{\mathrm{lb}}=\mathcal Q_{J_N^{\mathrm{lb}}}
=\{j\gamma_N:0\le j\le J_N^{\mathrm{lb}}\},\label{eq:lower-decision-grid}\\
\gamma_N=O(N^{-5/4}),\qquad
\log(J_N^{\mathrm{lb}}+1)=O(\log N).\label{eq:lower-resolution}
\end{gather}
For example, $J_N^{\mathrm{lb}}=\lceil N^{5/4}\rceil$ satisfies both conditions, and so does the finer choice $\lceil N^{5/2}\rceil$ obtained from \eqref{eq:N-grid} with $d=1$ and $\zeta=1$. The grid is fixed within each run and is the same for every objective in the RKHS ball. The first condition in \eqref{eq:lower-resolution} makes the discretization bias $b_{\mathcal G}(f):=f^*-\max_{x\in\mathcal G}f(x)$ negligible. For the kernel \eqref{eq:lower-matern-kernel}, every $f$ with $\|f\|_{\mathcal H_k}\le B$ satisfies $\|f''\|_\infty\le B$ by Proposition~\ref{proposition:lb-norm}. Since the grid contains both boundary points, Taylor's formula at an interior maximizer gives
\begin{equation}\label{eq:lower-grid-bias}
\sup_{\|f\|_{\mathcal H_k}\le B}b_{\mathcal G_N^{\mathrm{lb}}}(f)
\le\frac{B}{8}\gamma_N^2
=O(N^{-5/2}),
\end{equation}
which is of smaller order than the rate $N^{-5/2}\sqrt{\log N}$ of the lower bound. The second condition keeps the Gaussian envelope coefficient in Theorem~\ref{thm:simple-regret-discrete-general} of order $\sqrt{\log N}$, so that the upper bound of Section~\ref{sec:GP-TS-simple-regret} holds on the same grids. Appendix~\ref{app:thm-lower} verifies both statements.

\begin{figure}[t]
\centering
\includegraphics[width=\linewidth]{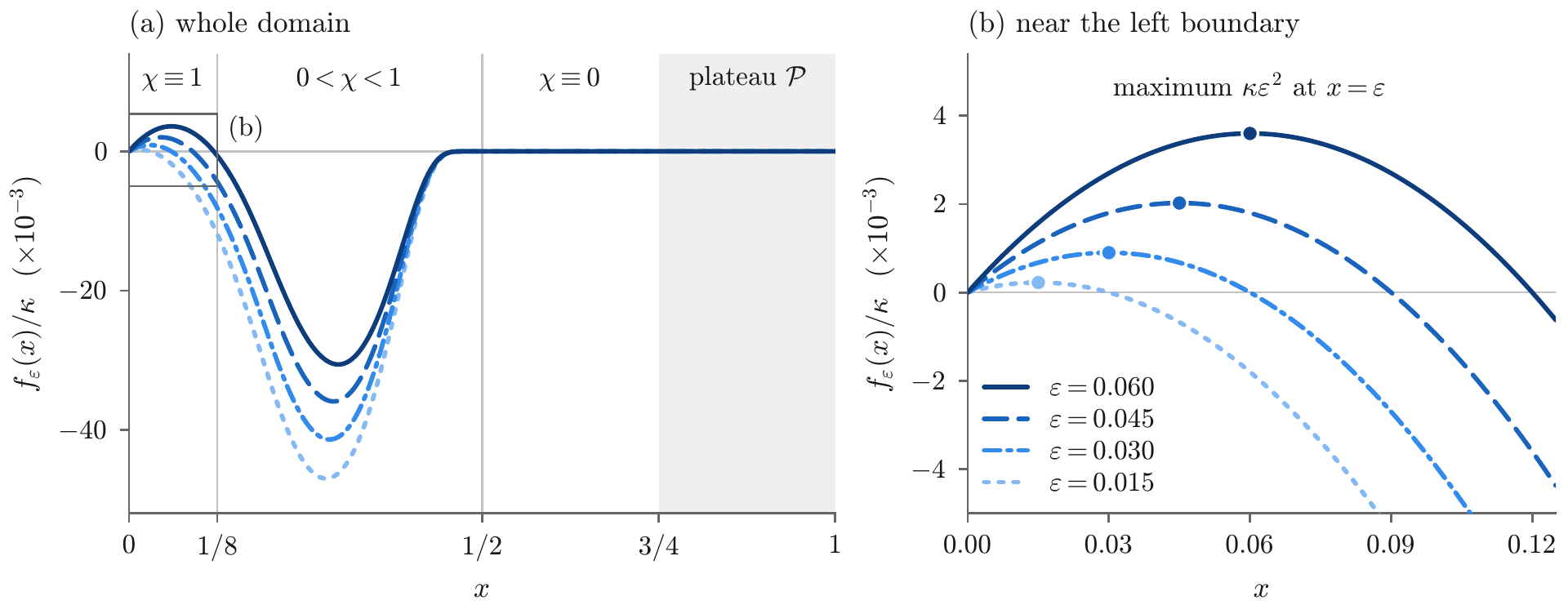}
\caption{The hard objectives $f_\varepsilon$ in \eqref{eq:lower-objective-family} for four values of $\varepsilon$ and one admissible cutoff $\chi$, in units of $\kappa$. (a) Each objective is non-positive on $[2\varepsilon,1]$ and vanishes on $[1/2,1]$, which contains the plateau $\mathcal P$. (b) On $[0,1/8]$ the objective is the parabola $\kappa[\varepsilon^2-(x-\varepsilon)^2]$, and the dots mark its maximum $\kappa\varepsilon^2$ at $x=\varepsilon$. The values of $\varepsilon$ are chosen only for visibility.}
\label{fig:hard-objectives}
\end{figure}

\runinlabel{Hard objectives} 
Let $\chi:\mathbb R\to[0,1]$ be a fixed smooth function with compact support that equals $1$ on $[0,1/8]$ and $0$ on $[1/2,1]$. The functions $g_1(x)=x\chi(x)$ and $g_2(x)=x^2\chi(x)$ are smooth with compact support, so their restrictions to $[0,1]$ belong to $\mathcal H_k$. Choose a fixed $\kappa>0$ such that $\kappa\bigl(2\|g_1\|_{\mathcal H_k}+\|g_2\|_{\mathcal H_k}\bigr)\le B/2$, and define the family of objectives
\begin{equation}\label{eq:lower-objective-family}
\{f_\varepsilon: 0<\varepsilon<1/16\},\qquad
f_{\varepsilon}(x)=\kappa\chi(x)(2\varepsilon x-x^2)
=\kappa\bigl(2\varepsilon g_1(x)-g_2(x)\bigr).
\end{equation}
On $[0,1/8]$, the cutoff $\chi$ equals one, so that $f_\varepsilon(x)=\kappa[\varepsilon^2-(x-\varepsilon)^2]$. Since $f_\varepsilon\le0$ on $[2\varepsilon,1]$, the function attains its unique maximum $f_\varepsilon^*=\kappa\varepsilon^2$ at $x=\varepsilon$. Moreover, $f_\varepsilon(0)=0$ and $f_\varepsilon$ vanishes on the plateau $\mathcal P=[3/4,1]$. Figure~\ref{fig:hard-objectives} illustrates the functions in this family with different $\varepsilon$. By construction, $\|f_\varepsilon\|_{\mathcal H_k}\le B/2$, so every $f_\varepsilon$ belongs to the RKHS ball of radius $B$. For a budget $N$, we write $f_N=f_{\varepsilon_N}$ and $\Delta_N=f_N^*=\kappa\varepsilon_N^2$, where $\varepsilon_N$ may depend on $N$ but is fixed before the algorithm is run.

\runinlabel{Sampling convention}
Let $\mathcal D_N$ denote the decision domain, which is either $\mathcal G_N^{\mathrm{lb}}$ or $[0,1]$ and is fixed within a run. Each implementation maximizes a fresh conditional sample exactly over $\mathcal D_N$, so that $\eta_n=0$ in \eqref{eq:TS-selection}. For the kernel \eqref{eq:lower-matern-kernel}, the posterior sample paths can be taken twice continuously differentiable by Proposition~\ref{proposition:lb-paths}, and we always use such paths. In order to treat both implementations with the same notation, we realize the Gaussian vector of the discrete implementation as the restriction to the grid of such a path, which does not change the algorithm because only the grid values determine the query. In both cases we therefore write
\begin{equation}\label{eq:lower-sampling}
G_n=\mu_{n-1}+vZ_n,\qquad
Z_n\mid\mathcal F_{n-1}\sim\mathcal{GP}(0,c_{n-1}),\qquad
x_n\in\operatorname*{arg\,max}_{x\in\mathcal D_N}G_n(x),
\end{equation}
where $\mathcal F_{n-1}$ is the complete history before round $n$, including the sampled paths of the previous rounds.

\subsection{Boundary Event and Additional Two Key Lemmas}\label{subsec:lower-key-lemmas}
\runinlabel{Boundary event}
Let $\mathcal E_{\partial}=\{x_1=0\}\in\mathcal F_1$ be the event that the first query is the left boundary point. Proposition~\ref{proposition:lb-boundary} shows that $\mathbb P(\mathcal E_{\partial})\ge q_{\partial}$ for both implementations, where the constant $q_{\partial}>0$ depends neither on the budget nor on the objective. On this event the first observation is $f_N(0)=0$, and exact posterior interpolation gives $G_n(0)=0$ for every $n\ge2$. A sample path that vanishes at zero can take a large value at a maximizer close to zero only if its second derivative is large, so bounds on the curvature of the samples translate into bounds on the sampled values near the peak. Both lemmas below are stated conditionally on $\mathcal E_{\partial}$.

The first lemma states that, from a fixed round onward, every Gaussian sample exceeds a level of order $n^{-5/2}\sqrt{\log n}$ somewhere on the plateau, although the objective vanishes there. We refer to this phenomenon as \textit{Gaussian overshoot}.

\begin{lemma}[Persistent plateau Gaussian overshoot]\label{lemma:lower-plateau}
For the kernel \eqref{eq:lower-matern-kernel}, let GP-TS operate on either $\mathcal D_N=[0,1]$ or a fixed grid sequence satisfying \eqref{eq:lower-decision-grid} and \eqref{eq:lower-resolution}. Let $f_N=f_{\varepsilon_N}$ satisfy $\kappa\varepsilon_N^2\le N^{-5/2}\sqrt{\log N}$. There exist a constant $b_*\in(0,1]$ and an integer $m_*\ge2$ such that, with the levels
\begin{equation}\label{eq:lower-overshoot-level}
\beta_n=b_*n^{-5/2}\sqrt{\log n},\qquad n\ge2,
\end{equation}
the following bound holds for all sufficiently large $N$:
\begin{equation}\label{eq:lower-plateau-event}
\mathbb P_{f_N}\left(
\max_{z\in\mathcal D_N\cap\mathcal P}G_n(z)>\beta_n
\ \text{ for every }\ m_*<n\le N
\,\middle|\,\mathcal E_{\partial}\right)\ge\frac78.
\end{equation}
\end{lemma}

\runinlabel{Proof sketch} The proof is given in Appendix~\ref{app:lemma-lower-plateau} and has three steps. First, the actual queries cover the plateau: after $m$ rounds, every point of $\mathcal P$ lies within a distance of order $m^{-1}(\log m)^{1/5}$ from the nearest observed point, except on an event of exponentially small probability. Second, the gaps between neighboring plateau observations provide comparison points whose number grows almost linearly in $n$, at which the posterior values, normalized by their natural scale, form a Gaussian vector whose covariance matrix has eigenvalues bounded above and below. Third, the maximum of such a Gaussian vector exceeds a multiple of $\sqrt{\log n}$ except on an event whose probability is summable in $n$.

The second lemma controls the second derivative, i.e., the curvature of the samples. A curvature bound of order $\sqrt{\log N}$ holds on the whole $[0,1]$ in every round. A curvature bound that does not grow with $N$ holds on a short interval $[0,h_N]$ as soon as this interval contains three distinct observations, namely the boundary point and two further points.

\begin{lemma}[Local observations and curvature]\label{lemma:lower-local}
Under the conditions of Lemma~\ref{lemma:lower-plateau}, there exist constants $A_{\mathrm{loc}},C_g,L,c>0$, with $L\ge\max\{1,\kappa\}$, such that the following holds for all sufficiently large $N$. Define
\begin{equation}\label{eq:lower-local-scales}
h_N=\frac{A_{\mathrm{loc}}}{\log N},\qquad
m_N=\lceil N^{1/4}\rceil,
\end{equation}
and let $\tau_N^{\mathrm{loc}}$ be the first round $n$ such that the queries $x_1,\ldots,x_n$ include two distinct points of $(0,h_N)$, so that $[0,h_N]$ contains three distinct observations on $\mathcal{E}_\partial$, with $\tau_N^{\mathrm{loc}}=\infty$ if there is no such round. Then
\begin{align}
\mathbb P_{f_N}(\tau_N^{\mathrm{loc}}>m_N\mid\mathcal E_{\partial})
&\le e^{-cN^{3/16}},\label{eq:lower-local-visits}\\
\mathbb P_{f_N}\left(\max_{2\le n\le N}\|G_n''\|_\infty
>C_g\sqrt{\log N}\,\middle|\,\mathcal E_{\partial}\right)
&\le N^{-2},\label{eq:lower-global-curvature}\\
\mathbb P_{f_N}\left(\tau_N^{\mathrm{loc}}\le m_N,
\max_{m_N<n\le N}\|G_n''\|_{\infty,[0,h_N]}>L
\,\middle|\,\mathcal E_{\partial}\right)
&\le N^{-2}.\label{eq:lower-local-curvature}
\end{align}
Here $\|\cdot\|_{\infty,[0,h_N]}$ denotes the supremum norm on $[0,h_N]$, and the derivatives refer to the sample paths in \eqref{eq:lower-sampling}, which serve only as auxiliary variables in the discrete implementation.
\end{lemma}

\runinlabel{Proof sketch} The proof is given in Appendix~\ref{app:lemma-lower-local}. For the arrival bound \eqref{eq:lower-local-visits}, as long as $(0,h_N)$ contains fewer than two observations, an interval without observations supplies a test point $z_n\in (0,h_N)$ at which the sample value $G_n(z_n)$ is large with a fixed conditional probability. On the uniform Gaussian envelope event, such a large $G_n(z_n)$ forces a large innovation $s_{n-1}(x_n)$ at the actual query, which Lemmas~\ref{lemma:deterministic-bound-arbitrary-subseq} and~\ref{lemma:conditional-success-prob} allow only in a limited number of rounds. The curvature bounds \eqref{eq:lower-global-curvature} and \eqref{eq:lower-local-curvature} follow from a dyadic chaining bound for the second derivative $Z_n^{\prime\prime}$ of the posterior residual, whose variance is at most one in general and of order $h_N$ on $[0,h_N]$ once this interval contains three distinct observations. The scales $h_N$ and $m_N$ are parameters of the analysis only.

\subsection{Simple Regret Lower Bounds for the Mat\'ern-$5/2$ Kernel}\label{subsec:lower-theorem}
We now choose the height of the peak so that it is dominated by the Gaussian overshoot on the plateau. For either implementation, fix $b_*$ from Lemma~\ref{lemma:lower-plateau} and $L$ from Lemma~\ref{lemma:lower-local} before choosing the objective parameter, and set
\begin{equation}\label{eq:lower-epsilon-choice}
\varepsilon_N^2=\frac{\beta_N}{16L}
=\frac{b_*}{16L}N^{-5/2}\sqrt{\log N},\qquad f_N=f_{\varepsilon_N}.
\end{equation}
Since $b_*\le1$ and $L\ge\kappa$, the height $\Delta_N=\kappa\varepsilon_N^2$ satisfies the condition $\kappa\varepsilon_N^2\le N^{-5/2}\sqrt{\log N}$ of both Lemma~\ref{lemma:lower-plateau} and Lemma~\ref{lemma:lower-local}. Define the neighborhood of the global maximizer $x^*=\varepsilon_N$
\begin{equation}\label{eq:lower-near-optimal-set}
\mathcal A_N=(\varepsilon_N/2,3\varepsilon_N/2).
\end{equation}
Outside $\mathcal A_N$, the objective gap $f_N^*-f_N(x)$ is at least $\kappa\varepsilon_N^2/4$. The simple regret is therefore at least $\kappa\varepsilon_N^2/4$ whenever no query falls in $\mathcal A_N$.

\begin{theorem}[Mat\'ern logarithmic lower bound]\label{thm:matern-log-lower}
Fix $B,v>0$ and the kernel \eqref{eq:lower-matern-kernel}. Run Algorithm~\ref{alg:gp-ts} with fresh exact posterior draws and exact maximization, either on $\mathcal D=[0,1]$ or on any deterministic sequence of regular grids satisfying \eqref{eq:lower-decision-grid} and \eqref{eq:lower-resolution}. There exist constants $c_{\mathrm{lb}},q_{\mathrm{lb}}>0$ and an integer $N_0$ such that, for every $N\ge N_0$, the deterministic objective $f_N$ in \eqref{eq:lower-epsilon-choice} belongs to the RKHS ball of radius $B$ and satisfies
\begin{equation}\label{eq:matern-log-lower-hp}
\mathbb P_{f_N}\left(r_N^{\mathrm{best}}
\ge c_{\mathrm{lb}}N^{-5/2}\sqrt{\log N}\right)
\ge q_{\mathrm{lb}}.
\end{equation}
Consequently, for both implementations,
\begin{equation}\label{eq:matern-tight-rate}
\sup_{\|f\|_{\mathcal H_k}\le B}\mathbb E_f[r_N^{\mathrm{best}}]
=\Theta\left(N^{-5/2}\sqrt{\log N}\right).
\end{equation}
The constants may depend on the implementation and on the fixed problem and resolution parameters, but not on $N$. The objective $f_N$ is fixed before the algorithm is run and may vary with the budget within the same RKHS ball.
\end{theorem}

\runinlabel{Proof sketch}
We show that, conditionally on $\mathcal E_{\partial}$, all $N$ queries avoid $\mathcal A_N$ with probability at least $1/2$. The first query is $x_1=0\notin\mathcal A_N$. The remaining rounds rely on one comparison, in which we set $\gamma_N=0$ for the continuous implementation. Suppose that $x_n\in(0,1)$ maximizes $G_n$ over $\mathcal D_N$ and that $|G_n''|\le\Lambda_n$ on $[0,x_n+\gamma_N]$. A grid maximizer dominates its two neighbors $x_n\pm \gamma_N$, and an interior maximizer of a $C^2$ sample path over $[0,1]$ satisfies the exact first order condition $G_n'(x_n)=0$. In both cases, Taylor's formula and $G_n(0)=0$ give
\begin{equation}\label{eq:lower-discrete-comparison}
|G_n'(0)|\le\Lambda_n\bigl(x_n+\tfrac12\gamma_N\bigr),
\qquad
G_n(x_n)\le\tfrac12\Lambda_n(x_n^2+\gamma_Nx_n).
\end{equation}
We divide the rounds into three stages and show that $x_n\notin\mathcal A_N$ in each stage. For the initial stage $2\le n\le m_*$, Lemma~\ref{lemma:lower-plateau} is not yet available, so we argue probabilistically. The first bound in~\eqref{eq:lower-discrete-comparison} forces $G_n'(0)$ to be nearly zero whenever $x_n$ lies close to the boundary. Gaussian anti-concentration of $G_n'(0)$ makes this unlikely over a sequence of shrinking protective radii, so $x_n\notin\mathcal A_N$ with probability $1-o(1)$ (Proposition~\ref{proposition:lb-initial-rounds}). For the intermediate stage $m_*<n\le m_N$ and last stage $m_N< n\le N$, we argue by contradiction on the good events. If $x_n\in\mathcal A_N$, the second bound in~\eqref{eq:lower-discrete-comparison} gives $G_n(x_n)<\beta_n$. Here $\Lambda_n=C_g\sqrt{\log N}$ for $m_*<n\le m_N$ by \eqref{eq:lower-global-curvature}, and $\Lambda_n=L$ for $n>m_N$ by \eqref{eq:lower-local-curvature}. This contradicts the plateau Gaussian overshoot of Lemma~\ref{lemma:lower-plateau}. Intersecting these events with $\mathcal E_{\partial}$ yields \eqref{eq:matern-log-lower-hp}. The matching upper bound follows from Theorem~\ref{thm:simple-regret-discrete-general} and \eqref{eq:lower-grid-bias} for the discrete implementation and from Corollary~\ref{corollary:Matern-continuous} for the continuous implementation. Appendices~\ref{app:lb-common} and~\ref{app:thm-lower} give the details. For the discrete implementation, the logarithmic lower bound persists although the uniform discretization bias in \eqref{eq:lower-grid-bias} is $o(N^{-5/2}\sqrt{\log N})$, so it is not caused by the discrete grid.

\begin{remark}[Suboptimality of exact GP-TS]
Theorem~\ref{thm:matern-log-lower} establishes the tight worst-case expected convergence rate $\Theta(N^{-5/2}\sqrt{\log N})$ for the fixed one-dimensional Mat\'ern-$5/2$ kernel, which exceeds the minimax-optimal rate $\Theta(N^{-5/2})$. The logarithmic loss occurs under exact posterior sampling and exact maximization in both implementations. The construction shows that positive Gaussian fluctuations over a suboptimal plateau can repeatedly dominate the sampled values near the true maximizer. Optimism alone therefore does not ensure the minimax-optimal convergence rate, because excessive positive sample deviations can cause a logarithmic loss.
\end{remark}

\begin{remark}[Choice of Mat\'ern-$5/2$ parameters]
The kernel, its lengthscale, and the dimension $d=1$ are fixed independently of the evaluation budget. The explicit RKHS norm and derivative structure of this kernel, together with the natural ordering of points in one dimension, make the overshoot mechanism transparent and tractable. We expect the approach to extend to other Mat\'ern kernels and to higher dimensions, and establishing such extensions remains open. Nevertheless, a single fixed kernel already suffices for concluding the limitation of GP-TS itself.
\end{remark}

\section{GP-CTS and Minimax-Optimal Convergence Rates}\label{sec:GP-CTS-simple-regret}
The analyses in Sections~\ref{sec:GP-TS-simple-regret} and~\ref{sec:log-factor-GP-TS} provide an important insight into the additional $\sqrt{\log N}$ factor in the convergence rates of GP-TS. The upper bound introduces this factor when controlling positive Gaussian deviations at the selected query, while the lower bound shows that Gaussian overshoot over a suboptimal region can indeed cause such a loss. These results suggest limiting excessive positive sample deviations while preserving optimism at a fixed comparator. Motivated by this observation, we propose Gaussian Process Capped Thompson Sampling (GP-CTS).

\subsection{Algorithmic Details of GP-CTS}\label{subsec:gp-cts-algorithm}
The deterministic RKHS inequality \eqref{eq:deterministic-upper-bound} provides an upper bound on the objective at every point. Using this bound to cap a posterior sample prevents excessively large sampled values without changing whether the sample is optimistic at a fixed comparator. To construct the cap, GP-CTS requires a known valid RKHS norm upper bound $\overline B\ge B>0$, which is fixed independently of the evaluation budget and the particular objective. The algorithm does not need the exact value of $\|f\|_{\mathcal H_k}$. For $n\ge0$, define the upper envelope
\begin{equation}\label{eq:cts-upper-envelope}
U_n(x)=\mu_n(x)+\overline B s_n(x),\qquad x\in\mathcal X.
\end{equation}
By \eqref{eq:deterministic-upper-bound}, $f(x)\le U_n(x)$ holds for every realized history and every $x\in\mathcal X$. At round $n$, draw $G_n$ using the same conditional sampling rule as in Section~\ref{subsec:gp-ts}, with the fixed scale $v>0$. Define the capped acquisition function
\begin{equation}\label{eq:GP-CTS}
\begin{aligned}
H_n(x)&=\min\{G_n(x),U_{n-1}(x)\}\\
&=\mu_{n-1}(x)+\min\{vZ_n(x),\overline B s_{n-1}(x)\},
\qquad x\in\mathcal D.
\end{aligned}
\end{equation}
The cap changes a sampled value only when it exceeds the deterministic upper envelope. GP-CTS then selects a measurable point $x_n\in\mathcal D$ satisfying
\begin{equation}\label{eq:CTS-selection}
H_n(x_n)\ge\sup_{x\in\mathcal D}H_n(x)-\eta_n.
\end{equation}
It then observes $y_n=f(x_n)$ and updates the GP posterior using \eqref{eq:gp-posterior}. For the discrete implementation, each coordinate of the joint Gaussian vector on $\mathcal G$ is capped by its corresponding value of $U_{n-1}$, and we set $\eta_n=0$ as in the discrete implementation of GP-TS. For the continuous implementation, $H_n$ is continuous whenever $G_n$ has continuous paths, because $\mu_{n-1}$ and $s_{n-1}$ are continuous. The measurable selection argument in Appendix~\ref{app:background} therefore also applies to $H_n$. Algorithm~\ref{alg:gp-cts} summarizes both implementations.

\begin{algorithm}[tbp]
\caption{Gaussian Process Capped Thompson Sampling (GP-CTS)}\label{alg:gp-cts}
\begin{algorithmic}[1]
\REQUIRE The inputs of Algorithm~\ref{alg:gp-ts}; known valid $\overline B\ge B$.
\ENSURE A best queried point $\widehat x_N$.
\STATE Run Algorithm~\ref{alg:gp-ts}, replacing the selection of $x_n$ in every round $n$ by the following two steps.
\STATE \hspace{1em}Set $U_{n-1}=\mu_{n-1}+\overline B s_{n-1}$ and $H_n=\min\{G_n,U_{n-1}\}$ on $\mathcal D$.
\STATE \hspace{1em}Select a measurable $x_n\in\mathcal D$ satisfying $H_n(x_n)\ge\sup_{x\in\mathcal D}H_n(x)-\eta_n$.
\end{algorithmic}
\end{algorithm}

\subsection{Theoretical Analysis}\label{subsec:gp-cts-analysis}
We first introduce a new lemma that shows the properties of the cap.

\begin{lemma}[Optimism preservation and positive sample deviations]\label{lemma:cts-optimism-overshoot}
\leavevmode\par\nobreak\noindent
Under Assumption~\ref{asmp:kernel-model}, let $\overline B\ge B$ and define $U_{n-1}$ and $H_n$ by \eqref{eq:cts-upper-envelope} and~\eqref{eq:GP-CTS}. For every fixed comparator $x^\dagger\in\mathcal D$,
\begin{equation}\label{eq:cts-optimism-event}
\{H_n(x^\dagger)\ge f(x^\dagger)\}
=\{G_n(x^\dagger)\ge f(x^\dagger)\}.
\end{equation}
Moreover, for every $x\in\mathcal D$, every realized history, and every sample realization,
\begin{equation}\label{eq:cts-overshoot}
H_n(x)-f(x)\le U_{n-1}(x)-f(x)
\le(B+\overline B)s_{n-1}(x).
\end{equation}
\end{lemma}

\begin{proof}
By \eqref{eq:deterministic-upper-bound} and $\overline B\ge B$, every $x\in\mathcal D$ satisfies $U_{n-1}(x)=\mu_{n-1}(x)+\overline Bs_{n-1}(x)\ge f(x)$. For real numbers $g$, $w$, and $a$ with $w\ge a$, the inequality $\min\{g,w\}\ge a$ holds if and only if $g\ge a$. Taking $g=G_n(x^\dagger)$, $w=U_{n-1}(x^\dagger)$, and $a=f(x^\dagger)$ proves \eqref{eq:cts-optimism-event}. Moreover, $H_n\le U_{n-1}$ and \eqref{eq:deterministic-upper-bound} give $H_n(x)-f(x)\le\mu_{n-1}(x)-f(x)+\overline Bs_{n-1}(x)\le(B+\overline B)s_{n-1}(x)$, which is \eqref{eq:cts-overshoot}.
\end{proof}

The event identity~\eqref{eq:cts-optimism-event} preserves the conditional optimism probability $p=\Phi(-B/v)$ under exact Gaussian sampling. The bound~\eqref{eq:cts-overshoot} is one-sided, since negative sample deviations remain uncapped, but only the positive deviation is needed in the regret comparison. In particular, the hard objectives of Section~\ref{sec:log-factor-GP-TS} vanish on the plateau, so \eqref{eq:cts-overshoot} gives $H_n(z)\le(B+\overline B)s_{n-1}(z)$ at every plateau point $z$, whereas the uncapped sample exceeds the posterior standard deviation by a factor of order $\sqrt{\log n}$ somewhere on the plateau. The cap therefore removes precisely the Gaussian overshoot of Lemma~\ref{lemma:lower-plateau}.

Unlike in Section~\ref{sec:GP-TS-simple-regret}, both implementations of GP-CTS can be treated in a single theorem, since the deterministic bound \eqref{eq:cts-overshoot} replaces the uniform Gaussian envelope and the remaining difference is only the discretization bias. With $\mathcal D=\mathcal G$ for the discrete implementation and $\mathcal D=\mathcal X$ for the continuous implementation, define
\begin{gather}
f_{\mathcal D}^*=\max_{x\in\mathcal D}f(x),\qquad
b_{\mathcal D}=f^*-f_{\mathcal D}^*,\qquad
\overline b_{\mathcal D}=\sup_{\|f\|_{\mathcal H_k}\le B}b_{\mathcal D}(f),\label{eq:cts-domain-bias}\\
p=\Phi(-B/v),\qquad
C_{\mathrm{cap}}=(B+\overline B)C_{\mathrm{sep}}(2/p)^\alpha.\label{eq:cts-constants}
\end{gather}

\begin{theorem}[General upper bound of GP-CTS for both implementations]\label{thm:simple-regret-capped-general}
\leavevmode\par\nobreak\noindent
Under Assumptions~\ref{asmp:kernel-model} and~\ref{asmp:poly-kolmogorov-width}, run Algorithm~\ref{alg:gp-cts} for $N$ rounds with fixed $v>0$, a known fixed $\overline B\ge B$, fresh exact conditional draws, and the measurable global selection rule \eqref{eq:CTS-selection}. On the continuous domain, assume that the posterior samples admit continuous paths and use those versions. For every fixed $f$ in the radius-$B$ RKHS ball and every integer $N\ge1$, on $\mathcal{D}=\mathcal{G}$ or on $\mathcal{D}=\mathcal{X}$,
\begin{equation}\label{eq:cts-hp}
\mathbb P_f\left(
 r_N^{\mathrm{best}}>b_{\mathcal D}+C_{\mathrm{cap}}N^{-\alpha}
 +\max_{1\le n\le N}\eta_n
\right)\le e^{-pN/8}.
\end{equation}
Consequently,
\begin{equation}\label{eq:cts-expected}
\sup_{\|f\|_{\mathcal H_k}\le B}\mathbb E_f[r_N^{\mathrm{best}}]
\le \overline b_{\mathcal D}+C_{\mathrm{cap}}N^{-\alpha}
+\max_{1\le n\le N}\eta_n+2Be^{-pN/8}.
\end{equation}
\end{theorem}

\begin{proof}
Lemma~\ref{lemma:deterministic-bound-arbitrary-subseq} holds for every sequence of queries, and Lemma~\ref{lemma:conditional-success-prob} and \eqref{eq:app-optimism-prob} use only the conditional Gaussian law of $G_n$, which GP-CTS shares with GP-TS. These results therefore apply to the trajectory of GP-CTS. Fix a maximizer $x^\dagger$ of $f$ over $\mathcal D$, which exists because $f$ is continuous and $\mathcal D$ is finite or compact, and let $I_N=\{n\le N:G_n(x^\dagger)\ge f(x^\dagger)\}$. By \eqref{eq:app-optimism-prob} and Lemma~\ref{lemma:conditional-success-prob}, $\mathbb P_f(|I_N|<pN/2)\le e^{-pN/8}$. By \eqref{eq:cts-optimism-event}, every $n\in I_N$ satisfies $H_n(x^\dagger)\ge f(x^\dagger)=f_{\mathcal D}^*$. The selection rule \eqref{eq:CTS-selection} and \eqref{eq:cts-overshoot} then give
\begin{equation}\label{eq:capped-one-round}
f_{\mathcal D}^*-f(x_n)\le H_n(x_n)-f(x_n)+\eta_n\le(B+\overline B)s_{n-1}(x_n)+\eta_n.
\end{equation}
The last inequality holds at every point and for every sample realization, so no envelope event is needed. On $\{|I_N|\ge pN/2\}$, Lemma~\ref{lemma:deterministic-bound-arbitrary-subseq} provides a round $n_\circ\in I_N$ with $s_{n_\circ-1}(x_{n_\circ})\le C_{\mathrm{sep}}(pN/2)^{-\alpha}$, and hence $r_N^{\mathrm{best}}\le b_{\mathcal D}+f_{\mathcal D}^*-f(x_{n_\circ})\le b_{\mathcal D}+C_{\mathrm{cap}}N^{-\alpha}+\max_{1\le n\le N}\eta_n$ by \eqref{eq:capped-one-round}. This proves \eqref{eq:cts-hp}. Since $0\le r_N^{\mathrm{best}}\le2B$, taking expectations and then the supremum over the RKHS ball gives \eqref{eq:cts-expected}.
\end{proof}

Since the continuous analysis no longer requires Lemma~\ref{lemma:unif-envelope-continuous}, Assumption~\ref{asmp:canonical-metric} is not used in Theorem~\ref{thm:simple-regret-capped-general}, although it is sufficient to establish continuous sample paths and to control finite-grid bias. Moreover, unlike in Theorems~\ref{thm:simple-regret-discrete-general} and~\ref{thm:simple-regret-continuous-general}, the confidence level enters only through a budget condition and not through the bound itself: with exact maximization on $\mathcal X$, \eqref{eq:cts-hp} gives $r_N^{\mathrm{best}}\le C_{\mathrm{cap}}N^{-\alpha}$ with probability at least $1-\delta$ whenever $N\ge8\log(1/\delta)/p$.

We next specialize the bound to Mat\'ern kernels. The same finite grids as in Section~\ref{sec:GP-TS-simple-regret} are sufficient because their uniform discretization bias is already of order $N^{-\nu/d}$.

\begin{corollary}[Mat\'ern minimax rate of GP-CTS for both implementations]\label{corollary:Matern-capped}
\leavevmode\par\nobreak\noindent
For the fixed Mat\'ern kernel in Proposition~\ref{proposition:Matern-kernel-asmps}, run Algorithm~\ref{alg:gp-cts} with fixed $v>0$, a known fixed $\overline B\ge B$, fresh exact conditional draws, and exact maximization. On $\mathcal D=\mathcal G_N$ from \eqref{eq:N-grid} with $\alpha=\nu/d$, or on $\mathcal D=\mathcal X$,
\begin{equation}\label{eq:matern-capped-rate}
\sup_{\|f\|_{\mathcal H_k}\le B}\mathbb E_f[r_N^{\mathrm{best}}]
=O(N^{-\nu/d}).
\end{equation}
More generally, the same upper bound holds for any deterministic sequence of nonempty finite decision sets whose uniform bias is $O(N^{-\nu/d})$. This matches the minimax worst-case expected simple regret upper bound over the fixed Mat\'ern RKHS ball.
\end{corollary}

The proof is given in Appendix~\ref{app:gp-cts}. The minimax lower bound used in this comparison is the classical bound for the function class \citep[Theorem~1]{JMLR:v12:bull11a}, recalled in Appendix~\ref{app:cor-matern-discrete}, instead of the lower bound specific to standard GP-TS in Section~\ref{sec:log-factor-GP-TS}.

\begin{remark}[Role of the norm upper bound]\label{rem:cts-summary}
For Mat\'ern kernels, GP-CTS attains the minimax-optimal convergence rate $\Theta(N^{-\nu/d})$ for both discrete and continuous implementations. In particular, for the fixed one-dimensional Mat\'ern-$5/2$ kernel \eqref{eq:lower-matern-kernel}, it attains the rate $N^{-5/2}$ on $[0,1]$ and on every grid satisfying \eqref{eq:lower-decision-grid} and \eqref{eq:lower-resolution}, which eliminates the $\sqrt{\log N}$ penalty that Theorem~\ref{thm:matern-log-lower} establishes for standard GP-TS in the same setting. This improvement requires a known valid RKHS norm upper bound $\overline B$, which is an additional input compared with standard GP-TS. A fixed conservative choice of $\overline B$ affects the constants but not the convergence rate, although it may influence empirical performance.
\end{remark}

\begin{remark}[A broader direction for general acquisition function design]
\leavevmode\par\nobreak\noindent
Our analysis highlights the importance of controlling acquisition overestimation after query selection. Maximization favors large acquisition values, so uncertainty guarantees at a predetermined location do not automatically control overestimation at the selected query. For GP-TS, this overestimation takes the form of the Gaussian overshoot in Section~\ref{sec:log-factor-GP-TS}, and GP-CTS controls it through a uniform cap while preserving comparator optimism. This suggests a potentially promising broader acquisition function design direction: assess how acquisition function maximization selects uncertainty fluctuations, and control the resulting overestimation without removing opportunities for optimism.
\end{remark}

\section{Numerical Experiments}\label{sec:experiments}
Although this paper is primarily theoretical, we complement the analysis with a numerical comparison of GP-TS and GP-CTS under finite evaluation budgets. The experiments examine whether the cap improves the best-query simple regret in practice.

\runinlabel{Setup}
We consider three benchmark objectives, namely Eggholder ($d=2$), Styblinski-Tang ($d=4$), and Hartmann ($d=6$), with evaluation budgets $N=200$, $500$, and $1000$. All three objectives are nonconvex and multi-modal, thus commonly evaluated in prior works \citep{surjanovic2013virtual}. The input domains are rescaled to $[0,1]^d$, the observations are noiseless, and every run starts without initial observations.

All methods use a Mat\'ern-$5/2$ kernel whose lengthscale $\ell$ is fixed before the runs and is not estimated online, in accordance with the fixed kernel in Assumption~\ref{asmp:kernel-model}. For each task, $\ell$ is a maximum likelihood estimate that is computed once from the objective values at $256$ scrambled Sobol points \citep{sobol1967distribution}. These points are neither used as observations nor counted in the budget. The remaining parameters are the sampling scale $v$ and the cap input $\overline B$, and the theory ties both of them to the RKHS norm bound $B$. We choose $v=B$ such that the optimism probability satisfies $p=\Phi(-1)$ in Lemma~\ref{lemma:conditional-success-prob}, and GP-CTS requires $\overline B\ge B$. However, the RKHS norm of a benchmark objective is unknown. We therefore shift and rescale every objective such that its values on a fixed finite candidate set $\mathcal G\subset[0,1]^d$ have unit RKHS norm. This normalization uses offline evaluations on $\mathcal G$, which are not counted in the budget either. All methods observe the normalized objective, for which $B=1$ is a valid norm bound on $\mathcal G$, and we accordingly set $v=\overline B=1$.

We compare GP-TS and GP-CTS, each in its discrete and its continuous implementation, with $50$ macro-replications per method and task. The discrete implementations follow Algorithms~\ref{alg:gp-ts} and~\ref{alg:gp-cts} exactly with the decision domain $\mathcal G$, which is a regular $161\times161$ grid for Eggholder and consists of $2^{14}$ and $2^{15}$ scrambled Sobol points for the other two tasks. Every round draws a joint Gaussian vector on $\mathcal G$ and maximizes the criterion exactly. The continuous implementations select their queries in $[0,1]^d$. In practice, we approximate the continuous implementations in Sections~\ref{sec:GP-TS-simple-regret} and~\ref{sec:GP-CTS-simple-regret} using approximate pathwise sampling based on Matheron's rule and multi-start L-BFGS-B optimizer over $[0,1]^d$. Appendix~\ref{app:experiments} provides all details.

\begin{figure}[t]
\centering
\definecolor{tsblue}{HTML}{0072B2}
\definecolor{ctsorange}{HTML}{D55E00}
\newcommand{\legsolid}[1]{\textcolor{#1}{\rule[0.55ex]{1.9em}{1.1pt}}}
\newcommand{\legdashed}[1]{\textcolor{#1}{\rule[0.55ex]{0.5em}{1.1pt}\hspace{0.2em}\rule[0.55ex]{0.5em}{1.1pt}\hspace{0.2em}\rule[0.55ex]{0.5em}{1.1pt}}}
{\footnotesize
\legsolid{tsblue}~GP-TS (discrete)\hspace{1.6em}%
\legdashed{tsblue}~GP-TS (continuous)\hspace{1.6em}%
\legsolid{ctsorange}~GP-CTS (discrete)\hspace{1.6em}%
\legdashed{ctsorange}~GP-CTS (continuous)}\\[3pt]
\includegraphics[width=0.328\linewidth]{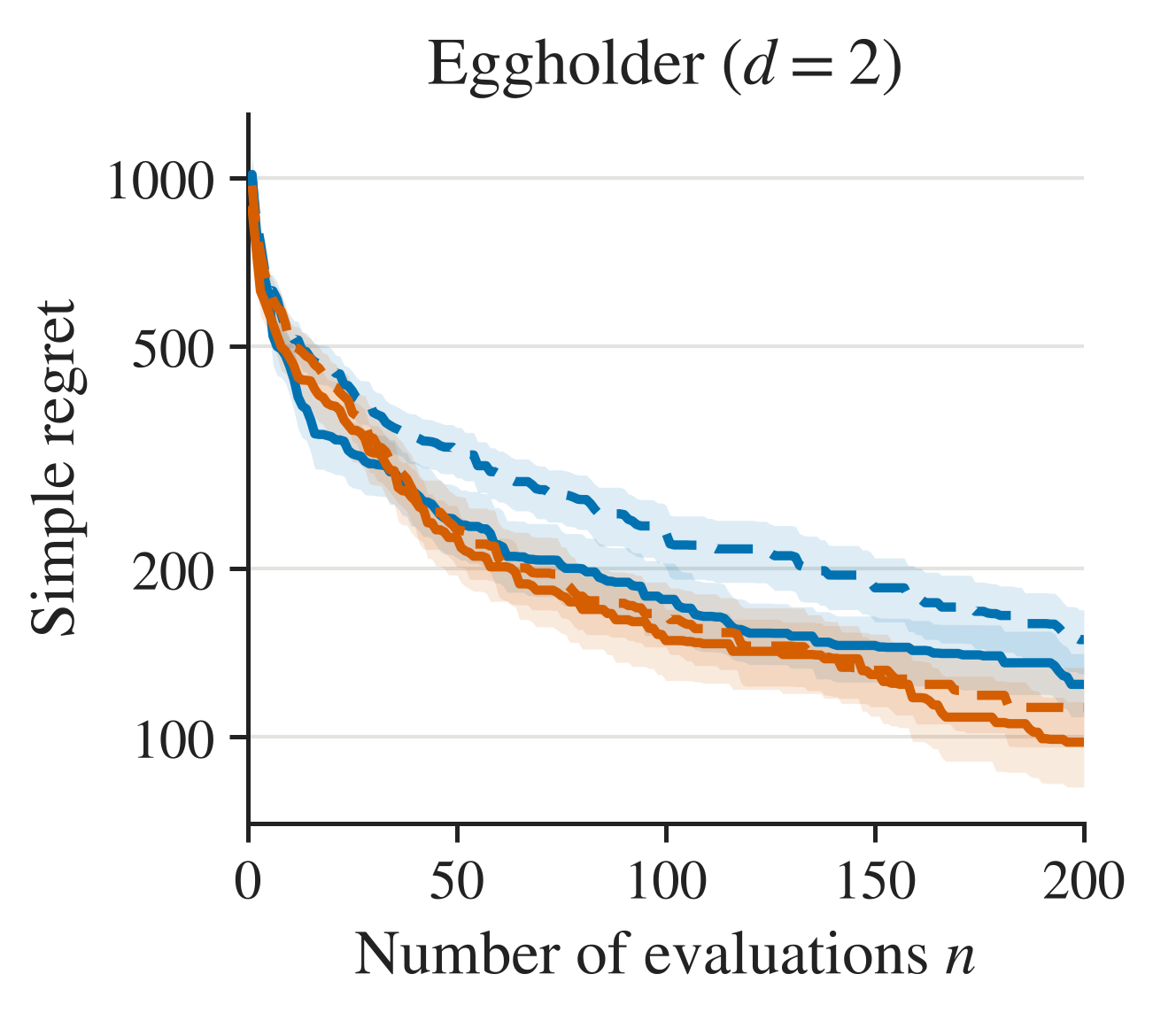}\hfill
\includegraphics[width=0.328\linewidth]{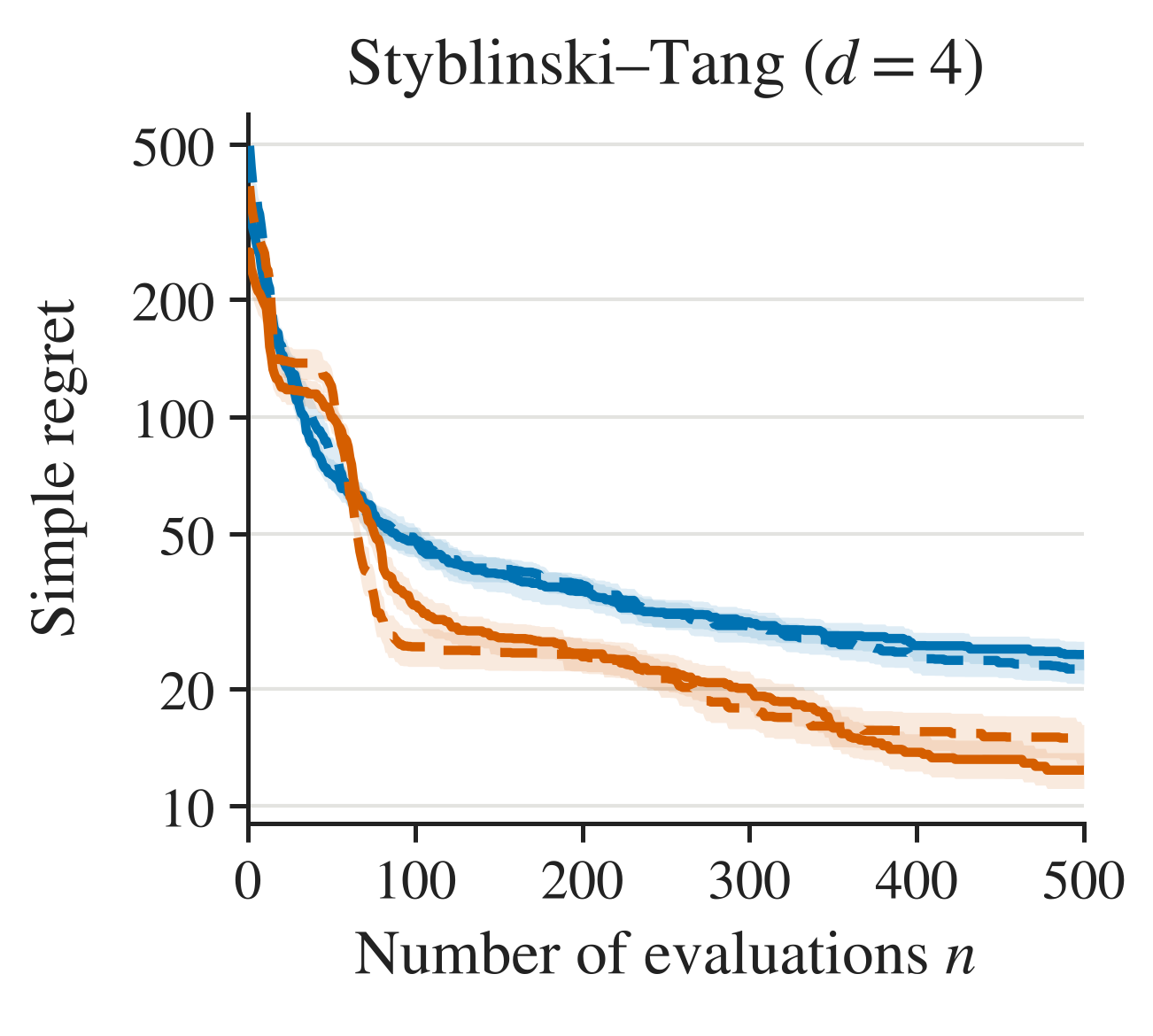}\hfill
\includegraphics[width=0.328\linewidth]{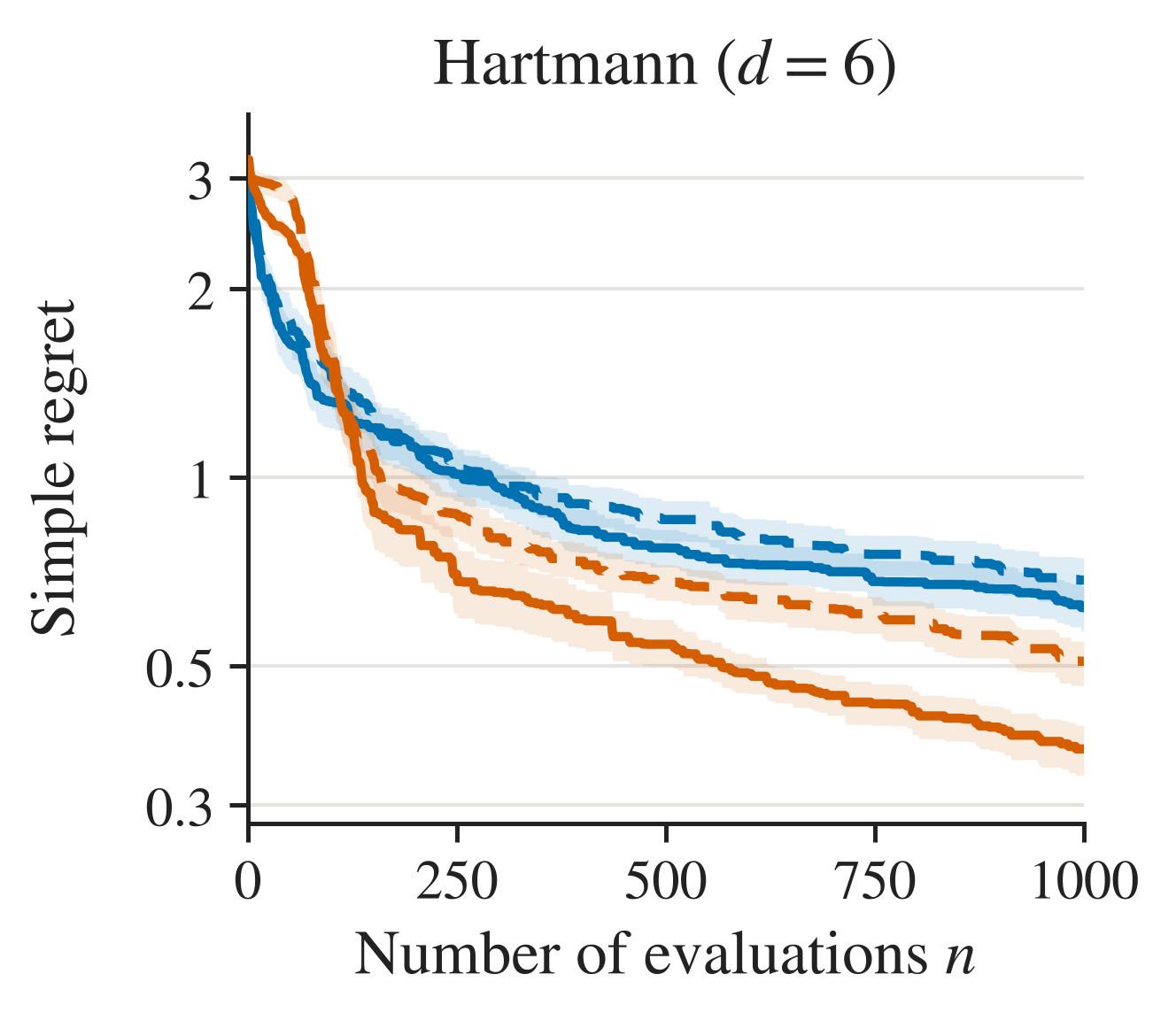}
\caption{Mean best-query simple regret over $50$ macro-replications, in the original units of each objective. The shaded regions are pointwise $95\%$ bootstrap intervals for the mean. Colors distinguish GP-TS and GP-CTS, and solid and dashed lines distinguish the discrete and the continuous implementations.}
\label{fig:regret}
\end{figure}

\runinlabel{Results}
Figure~\ref{fig:regret} reports the mean best-query simple regret of the four methods. GP-TS has the lower mean regret only during an initial phase, which ends within the first fifth of the budget on every task. Afterwards, GP-CTS attains the lower mean regret on all three tasks until the final budget, in the discrete and in the continuous implementations alike. At the final budget, the bootstrap intervals of the two algorithms are separated on Styblinski-Tang and Hartmann and overlap on Eggholder. Under finite evaluation budgets, the cap therefore improves the mean best-query simple regret of GP-TS on these tasks. Appendix~\ref{app:exp-evaluation} lists the numerical values. In the analysis, the cap acts by bounding the positive sample deviation at the selected query, as shown in Lemma~\ref{lemma:cts-optimism-overshoot}, and the regret curves are consistent with this effect.

\section{Conclusions}\label{sec:conclusions}
This paper investigates the convergence rates of Gaussian Process Thompson Sampling (GP-TS) in efficient global optimization under noiseless observations. Combining conditional optimism with a deterministic bound on posterior uncertainty along arbitrary subsequences of queries, we establish direct simple regret upper bounds for both discrete and continuous implementations of GP-TS. For Mat\'ern kernels with smoothness parameter $\nu$ in dimension $d$, we obtain the worst-case expected simple regret upper bound $O(N^{-\nu/d}\sqrt{\log N})$, recovering the full smoothness exponent $\nu/d$. For a fixed one-dimensional Mat\'ern-$5/2$ kernel, we establish matching lower bounds for both implementations. The analysis shows that positive Gaussian fluctuations over a suboptimal region can dominate sampled values near the true maximizer, so that the logarithmic factor is inherent to standard GP-TS in this setting instead of an artifact of the upper-bound analysis. Motivated by these findings, we propose Gaussian Process Capped Thompson Sampling (GP-CTS), which caps posterior sample paths using a deterministic RKHS upper envelope. Given a valid RKHS norm upper bound, the cap preserves optimism while controlling positive sample deviations, which removes the logarithmic factor and yields the minimax-optimal rate $\Theta(N^{-\nu/d})$ for Mat\'ern kernels in both implementations.

Several questions remain. Extending the logarithmic lower bounds beyond the present Mat\'ern parameters would clarify how widely the identified limitations of GP-TS exist. For GP-CTS, obtaining a valid norm upper bound without prior knowledge of the function class remains an additional challenge.

\section*{Acknowledgments and Disclosure of Funding}
{
We gratefully acknowledge the financial support from the Department of Industrial Systems Engineering and Management (ISEM), National University of Singapore. We have no relevant competing financial interests.
}

\appendix
\numberwithin{equation}{section}

\section{Proofs for Section~\ref{sec:background}}\label{app:background}
This appendix collects the facts about the posterior that are used throughout the paper and proves Proposition~\ref{proposition:Matern-kernel-asmps}.

\runinlabel{Posterior identities.}
For a non-singular Gram matrix, the identities \eqref{eq:rkhs-geometry} and the bound \eqref{eq:deterministic-upper-bound} are classical facts of kernel interpolation \citep[Chapter~11]{wendland2005scattered}, and \citet[Lemma~6]{JMLR:v12:bull11a} gives a closely related statement. They remain valid when queries are repeated and $K_n$ is singular. Indeed, if $a\in\ker K_n$, then $\|\sum_ia_ik_{x_i}\|_{\mathcal H_k}^2=a^\top K_na=0$, and taking the inner product of $\sum_ia_ik_{x_i}=0$ with $k_x$ and with $f$ gives $a^\top\mathbf k_n(x)=0$ and $a^\top\mathbf y_n=0$. Hence $\mathbf k_n(x)$ and $\mathbf y_n$ belong to $\operatorname{ran}K_n$, and $K_nK_n^\dagger\mathbf y_n=\mathbf y_n$. The function $g=\sum_i(K_n^\dagger\mathbf y_n)_ik_{x_i}\in W_n$ therefore interpolates the observations, so $f-g$ is orthogonal to $W_n$ and $\mu_n=g=P_nf$. The same argument with $k_x$ in place of $f$ gives
\begin{equation}\label{eq:app-projected-section}
P_nk_x=\sum_{i=1}^n\bigl(K_n^\dagger\mathbf k_n(x)\bigr)_ik_{x_i},
\qquad
\langle(I-P_n)k_x,(I-P_n)k_{x'}\rangle_{\mathcal H_k}
=k(x,x')-\mathbf k_n(x)^\top K_n^\dagger\mathbf k_n(x'),
\end{equation}
which is \eqref{eq:rkhs-geometry}. The reproducing property and the Cauchy-Schwarz inequality then give $|f(x)-\mu_n(x)|=|\langle f,(I-P_n)k_x\rangle_{\mathcal H_k}|\le Bs_n(x)$, which is \eqref{eq:deterministic-upper-bound}. In particular, $s_n(x_i)=0$ and $\mu_n(x_i)=f(x_i)$ at every observed point.

\runinlabel{Consequences.}
We record three consequences. First, for subspaces $V\subseteq W$ of $\mathcal H_k$ and $g\in\mathcal H_k$,
\begin{equation}\label{eq:app-projection-monotonicity}
\operatorname{dist}_{\mathcal H_k}(g,W)\le\operatorname{dist}_{\mathcal H_k}(g,V).
\end{equation}
Since $W_n\subseteq W_{n'}$ for $n\le n'$, additional observations never increase the posterior standard deviation. Second, a posterior residual $Z$ with conditional covariance $c_n$ given $\mathcal F_n$ satisfies
\begin{equation}\label{eq:app-increment-contraction}
\operatorname{Var}(Z(x)-Z(x')\mid\mathcal F_n)
=\|(I-P_n)(k_x-k_{x'})\|_{\mathcal H_k}^2\le d_k(x,x')^2,
\end{equation}
so conditioning does not increase the variance of an increment. Third, $|f(x)-f(x')|=|\langle f,k_x-k_{x'}\rangle_{\mathcal H_k}|\le Bd_k(x,x')$, which proves \eqref{eq:Holder-control} under Assumption~\ref{asmp:canonical-metric}. Since $k$ is continuous, $d_k(x,x')\to0$ as $x'\to x$, so $f$ is continuous and attains its maximum on $\mathcal X$. Moreover, $|f(x)|\le B\sqrt{k(x,x)}\le B$, and hence $0\le r_N^{\mathrm{best}}\le2B$.

\runinlabel{Sample paths and measurable selection.}
Let $\Xi\sim\mathcal{GP}(0,k)$. Under Assumption~\ref{asmp:canonical-metric}, the Gaussian increments satisfy $\mathbb E|\Xi(x)-\Xi(y)|^q=c_qd_k(x,y)^q\le c_qL_k^q\|x-y\|_2^{q\zeta}$ for every $q>0$, and the Kolmogorov-Chentsov theorem with $q>d/\zeta$ provides a version of $\Xi$ with continuous paths \citep[Theorem~3.23]{kallenberg2002foundations}. Given the history before round $n$, let $\Xi_n$ be a fresh copy of this version and define, as in pathwise conditioning \citep{wilson2020efficiently,wilson2021pathwise},
\begin{equation}\label{eq:app-exact-posterior-path}
Z_n(x)=\Xi_n(x)-\mathbf k_{n-1}(x)^\top K_{n-1}^\dagger\,\Xi_n(X_{n-1}).
\end{equation}
Conditionally on $\mathcal F_{n-1}$, this is a centered Gaussian process, and the identity 
\[
K_{n-1}^\dagger K_{n-1}K_{n-1}^\dagger=K_{n-1}^\dagger
\]
 shows that its covariance is $c_{n-1}$. Its paths are continuous, because the correction term is a finite linear combination of continuous kernel sections. This is the continuous posterior sample in \eqref{eq:TS-continuous-sample}. For the selection rule \eqref{eq:TS-selection}, a fixed ordering of a finite decision set resolves ties measurably. On $\mathcal D=\mathcal X$, the sample $G_n$ is continuous in $x$ and measurable in the randomness, so the measurable maximum theorem \citep[Theorem~18.19]{aliprantis2006infinite} provides a measurable maximizer $x_n$. The same argument applies to the capped sample $H_n$ of Section~\ref{sec:GP-CTS-simple-regret}, which is continuous whenever $G_n$ is.

\subsection{Proposition~\ref{proposition:Matern-kernel-asmps}: Mat\'ern Kernel Assumptions}\label{app:prop-matern}
\begin{proof}
We first verify Assumption~\ref{asmp:poly-kolmogorov-width}. For a finite set $X\subset\mathcal X$, let $h_X=\sup_{x\in\mathcal X}\min_{z\in X}\|x-z\|_2$ denote its fill distance, and let $s_X$ denote the posterior standard deviation given observations at the points of $X$. By \eqref{eq:rkhs-geometry} and the reproducing property, $s_X(x)$ is the largest value of $|u(x)|$ over the functions $u$ in the unit ball of $\mathcal H_k$ that vanish on $X$. The RKHS of the Mat\'ern kernel on $\mathcal X$ coincides with the Sobolev space of order $\nu+d/2$ on $(0,1)^d$, with equivalent norms \citep[Chapter~10]{wendland2005scattered}. Functions in this space that vanish on $X$ are uniformly small when $X$ fills the domain. More precisely, the error estimates of \citet{narcowich2003refined} and the sampling inequality of \citet{arcangeli2007extension} provide constants $C,h_0>0$, depending only on the kernel and $d$, such that
\begin{equation}\label{eq:app-power-function}
\sup_{x\in\mathcal X}s_X(x)\le Ch_X^\nu\qquad\text{whenever }h_X\le h_0.
\end{equation}
This is the classical bound on the power function of kernel interpolation, and \citet{JMLR:v12:bull11a} uses it in the same form for expected improvement. Since $s_X(x)=\operatorname{dist}_{\mathcal H_k}(k_x,\operatorname{span}\{k_z:z\in X\})$, applying \eqref{eq:app-power-function} to the regular grid $X=\mathcal Q_J$ from \eqref{eq:regular-grid}, whose fill distance is at most $h_J$, gives $w_m(k,\mathcal X)\le Ch_J^\nu$ whenever $M_J\le m$ and $h_J\le h_0$. The choice $J=\lfloor m^{1/d}\rfloor-1$ satisfies $M_J\le m$ and $J\ge m^{1/d}/2$ for $m\ge4^d$, and therefore $w_m(k,\mathcal X)\le C'm^{-\nu/d}$ for all sufficiently large $m$. For the remaining finitely many $m$, the zero subspace gives $w_m(k,\mathcal X)\le\sup_{x\in\mathcal X}\sqrt{k(x,x)}\le1$. Assumption~\ref{asmp:poly-kolmogorov-width} thus holds with $\alpha=\nu/d$.

For Assumption~\ref{asmp:canonical-metric}, let $\rho$ denote the spectral density of the kernel, so that $k(x,y)=\int_{\mathbb R^d}e^{i\omega\cdot(x-y)}\rho(\omega)\,d\omega$ and $\rho(\omega)\le C(1+\|\omega\|_2^2)^{-(\nu+d/2)}$ \citep{DBLP:books/lib/RasmussenW06}. Let $0<\zeta\le1$ with $\zeta<\nu$. The elementary inequality $|e^{iu}-1|\le2|u|^\zeta$ for real $u$ gives
\begin{equation}\label{eq:app-matern-metric}
d_k(x,y)^2
=\int_{\mathbb R^d}\bigl|e^{i\omega\cdot x}-e^{i\omega\cdot y}\bigr|^2\rho(\omega)\,d\omega
\le4\|x-y\|_2^{2\zeta}\int_{\mathbb R^d}\|\omega\|_2^{2\zeta}\rho(\omega)\,d\omega,
\end{equation}
and the last integral is finite because $\zeta<\nu$. This proves Assumption~\ref{asmp:canonical-metric} for every such $\zeta$, including $\zeta=1$ when $\nu>1$.
\end{proof}

\section{Proofs for Section~\ref{sec:GP-TS-simple-regret}}\label{app:gp-ts}

\subsection{Lemma~\ref{lemma:deterministic-bound-arbitrary-subseq}: Arbitrary Subsequence Innovation Bound}\label{app:lemma-innovation}
\begin{proof}
The proof follows the approximation geometry of \citet{DBLP:journals/corr/abs-2607-29245}.

\runinlabel{Step 1. Eigenvalue tails.}
Let $z_1,\ldots,z_m\in\mathcal X$ be arbitrary, and let $\lambda_1\ge\cdots\ge\lambda_m\ge0$ be the eigenvalues of the Gram matrix $K=[k(z_i,z_j)]_{i,j=1}^m$. Fix $1\le\ell<m$ and $\epsilon>0$. By Assumption~\ref{asmp:poly-kolmogorov-width}, there is a subspace $V\subset\mathcal H_k$ of dimension at most $\ell$ such that $\operatorname{dist}_{\mathcal H_k}(k_x,V)\le A\ell^{-\alpha}+\epsilon$ for every $x\in\mathcal X$. Let $P$ be the orthogonal projection onto $V$. The operator $S=\sum_{i=1}^m\langle\,\cdot\,,k_{z_i}\rangle_{\mathcal H_k}k_{z_i}$ has the same nonzero eigenvalues as $K$, and $\operatorname{tr}(PS)=\sum_i\|Pk_{z_i}\|_{\mathcal H_k}^2$. Since $P$ has rank at most $\ell$, Ky Fan's maximum principle \citep[Section~4.3]{horn2013matrix} gives $\operatorname{tr}(PS)\le\lambda_1+\cdots+\lambda_\ell$. Hence
\[
\sum_{j>\ell}\lambda_j\le\operatorname{tr}(S)-\operatorname{tr}(PS)
=\sum_{i=1}^m\|(I-P)k_{z_i}\|_{\mathcal H_k}^2
\le m(A\ell^{-\alpha}+\epsilon)^2,
\]
and letting $\epsilon\downarrow0$ yields
\begin{equation}\label{eq:app-gram-tail}
\sum_{j>\ell}\lambda_j\le mA^2\ell^{-2\alpha},\qquad1\le\ell<m.
\end{equation}

\runinlabel{Step 2. Determinant.}
For $j\ge2$, apply \eqref{eq:app-gram-tail} with $\ell=\lfloor j/2\rfloor$. Since $j-\ell\ge j/2$ and $\ell\ge j/3$,
\[
(j-\ell)\lambda_j\le\sum_{r=\ell+1}^j\lambda_r\le mA^2\ell^{-2\alpha},
\qquad\text{and hence}\qquad
\lambda_j\le2A^23^{2\alpha}mj^{-(1+2\alpha)}.
\]
Moreover, $\lambda_1\le\operatorname{tr}K\le m$. With $C_0=\max\{1,2A^23^{2\alpha}\}$, we therefore have $\lambda_j\le C_0mj^{-(1+2\alpha)}$ for $1\le j\le m$, and $m!\ge(m/e)^m$ gives
\begin{equation}\label{eq:app-gram-determinant}
\det K=\prod_{j=1}^m\lambda_j
\le\frac{(C_0m)^m}{(m!)^{1+2\alpha}}
\le\bigl(C_0e^{1+2\alpha}m^{-2\alpha}\bigr)^m
=\bigl(C_{\mathrm{sep}}^2m^{-2\alpha}\bigr)^m,
\end{equation}
where $C_{\mathrm{sep}}$ is the constant in \eqref{eq:C-sep}.

\runinlabel{Step 3. Innovations.}
Let $\widetilde s_j=\operatorname{dist}_{\mathcal H_k}(k_{z_j},\operatorname{span}\{k_{z_1},\ldots,k_{z_{j-1}}\})$, where the empty span is $\{0\}$. The Gram-Schmidt procedure gives $\det K=\prod_{j=1}^m\widetilde s_j^{\,2}$, and the smallest factor is at most the geometric mean, so
\begin{equation}\label{eq:app-ordered-innovation}
\min_{1\le j\le m}\widetilde s_j\le(\det K)^{1/(2m)}\le C_{\mathrm{sep}}m^{-\alpha}.
\end{equation}
Now let $I=\{n_1<\cdots<n_m\}$ and $z_j=x_{n_j}$. Since $\operatorname{span}\{k_{z_1},\ldots,k_{z_{j-1}}\}\subseteq W_{n_j-1}$, the monotonicity \eqref{eq:app-projection-monotonicity} gives $s_{n_j-1}(x_{n_j})\le\widetilde s_j$, and \eqref{eq:app-ordered-innovation} proves \eqref{eq:subsequence-innovation}. For \eqref{eq:innovation-count}, let $I=\{n\le N:s_{n-1}(x_n)>u\}$ and $m=|I|$. If $m>0$, then $u<\min_{n\in I}s_{n-1}(x_n)\le C_{\mathrm{sep}}m^{-\alpha}$, so $m<(C_{\mathrm{sep}}/u)^{1/\alpha}$.
\end{proof}

\subsection{Lemma~\ref{lemma:conditional-success-prob}: Conditional Lower-Tail Bound and Optimism}\label{app:lemma-conditional}
\begin{proof}
This is the Chernoff argument, applied conditionally. Let $S_n=\sum_{j=1}^nO_j$ and $\theta=\log2$. Since $\mathbb P(O_n=1\mid\mathcal F_{n-1})\ge p$ and $e^{-\theta}=1/2$, we have $\mathbb E\bigl[e^{-\theta O_n}\mid\mathcal F_{n-1}\bigr]\le1-p\,(1-e^{-\theta})=1-p/2\le e^{-p/2}$.
The tower property gives $\mathbb Ee^{-\theta S_N}\le e^{-p/2}\,\mathbb Ee^{-\theta S_{N-1}}$, and iteration yields $\mathbb Ee^{-\theta S_N}\le e^{-pN/2}$. By Markov's inequality,
\begin{equation}\label{eq:app-bernoulli-tail}
\mathbb P\Bigl(S_N<\frac{pN}{2}\Bigr)
\le e^{\theta pN/2}\,\mathbb Ee^{-\theta S_N}
\le\exp\Bigl(-\frac{1-\log2}{2}\,pN\Bigr)
\le e^{-pN/8},
\end{equation}
which is \eqref{eq:conditional-count}.
\end{proof}

\runinlabel{Optimism probability.}
Let $x^\dagger$ be a comparator that is fixed before the algorithm is run, and write $a=\mu_{n-1}(x^\dagger)$ and $s=s_{n-1}(x^\dagger)$. Conditionally on $\mathcal F_{n-1}$, the sampled value $G_n(x^\dagger)$ is Gaussian with mean $a$ and standard deviation $vs$, and $f(x^\dagger)-a\le Bs$ by \eqref{eq:deterministic-upper-bound}. If $s>0$, then
\begin{equation}\label{eq:app-optimism-prob}
\mathbb P\bigl(G_n(x^\dagger)\ge f(x^\dagger)\mid\mathcal F_{n-1}\bigr)
=1-\Phi\Bigl(\frac{f(x^\dagger)-a}{vs}\Bigr)
\ge\Phi(-B/v)=p.
\end{equation}
If $s=0$, then $G_n(x^\dagger)=a=f(x^\dagger)$ almost surely, and the probability equals one. The indicators $O_n=\mathbf 1\{G_n(x^\dagger)\ge f(x^\dagger)\}$ therefore satisfy \eqref{eq:lemma2-if}. Both $x^*$ and $x_{\mathcal G}^*$ are admissible comparators, whereas the query $x_n$ is not, because it depends on the sample $G_n$.

\subsection{Corollary~\ref{corollary:Matern-discrete}: Discrete Mat\'ern Rate}\label{app:cor-matern-discrete}
\begin{proof}
Every $x\in\mathcal X$ lies within distance $h_J$ of a point of $\mathcal Q_J$, so \eqref{eq:Holder-control} gives $b_{\mathcal Q_J}\le BL_kh_J^\zeta$ for every $f$ in the RKHS ball. The choice \eqref{eq:N-grid} implies $h_{J_N}\le N^{-\alpha/\zeta}$ and $J_N+1\le(\sqrt d+2)N^{\alpha/\zeta}$, which proves \eqref{eq:grid-bias-size}. Consequently, for $N\ge2$,
\begin{equation}\label{eq:app-grid-log}
u_{N,|\mathcal G_N|,\delta}^2
\le2\log\frac{4C_d}{\delta}+2\Bigl(1+\frac{\alpha d}{\zeta}\Bigr)\log N
\le C\log(N/\delta).
\end{equation}
Substituting \eqref{eq:grid-bias-size} and \eqref{eq:app-grid-log} into Theorem~\ref{thm:simple-regret-discrete-general} with $\eta_n=0$ and $\alpha=\nu/d$ proves \eqref{eq:matern-discrete-hp}. The terms without a logarithmic factor are absorbed because $\log(N/\delta)\ge\log2$.

For the expected regret, choose $\delta_N=N^{-(\alpha+1)}$. The budget condition of the high-probability bound becomes $N\ge(8/p)[\log2+(\alpha+1)\log N]$, which holds for all $N\ge N_0$, where $N_0\ge2$ depends only on $p$ and $\alpha$. Using $r_N^{\mathrm{best}}\le2B$ on the exceptional event, we obtain
\begin{equation}\label{eq:app-raw-expectation}
\mathbb E_f[r_N^{\mathrm{best}}]
\le CN^{-\alpha}\sqrt{\log(N/\delta_N)}+2B\delta_N
\le C'N^{-\alpha}\sqrt{\log N}.
\end{equation}
The constants and $N_0$ do not depend on $f$, so taking the supremum over the RKHS ball proves \eqref{eq:matern-discrete-expected}.
\end{proof}

\runinlabel{Minimax benchmark.}
\citet[Theorem~1]{JMLR:v12:bull11a} shows that, over a fixed Mat\'ern RKHS ball, the smallest worst-case expected simple regret achievable with $N$ noiseless evaluations is of order $N^{-\nu/d}$. The infimum is taken over all sequential strategies, which may be randomized and may recommend an arbitrary point of $\mathcal X$. Algorithms that return the best queried point form a subclass, so the lower bound of order $N^{-\nu/d}$ applies to the best-query simple regret as well. Corollary~\ref{corollary:Matern-capped} shows that this order is attained under the best-query loss. The minimax-optimal rate for $r_N^{\mathrm{best}}$ is therefore $N^{-\nu/d}$, and Corollary~\ref{corollary:Matern-discrete} matches it up to the factor $\sqrt{\log N}$.

\subsection{Lemma~\ref{lemma:unif-envelope-continuous}: Continuous Residual Envelope}\label{app:lemma-continuous-envelope}
We first state the dyadic chaining bound for a single Gaussian process. Lemma~\ref{lemma:unif-envelope-continuous} follows by applying it to the posterior residual of every round. Appendix~\ref{app:log-factor} applies it again to a single round and to the second derivative of a posterior sample.

\begin{applemma}[Chaining bound]\label{lemma:app-chaining}
Let $U$ be a centered Gaussian process on $[0,1]^d$ with continuous paths, and let $\sigma(x)$ denote the standard deviation of $U(x)$. Assume that there are constants $L>0$ and $\zeta\in(0,1]$ such that $\operatorname{Var}(U(x)-U(x'))\le L^2\|x-x'\|_2^{2\zeta}$ for all $x,x'\in[0,1]^d$. Then, for every integer $J\ge1$ and every $0<\delta<1$, with $u=\sqrt{2\log(4M_J/\delta)}$ and the constants in \eqref{eq:continuous-envelope-constants},
\begin{equation}\label{eq:app-chaining-bound}
\mathbb P\Bigl(|U(x)|\le u\,\sigma(x)+Lh_J^\zeta\bigl[C_1(\zeta)u+C_2(d,\zeta)\bigr]
\ \text{ for every }x\in[0,1]^d\Bigr)\ge1-\delta.
\end{equation}
\end{applemma}
\begin{proof}
We bound $U$ on a coarse grid, then its increments along dyadic refinements of this grid, and finally extend the bound to the whole domain.

\runinlabel{Nested grids.}
For $j\ge0$, let $\mathcal Q_{J,j}=\mathcal Q_{J2^j}$ be the regular grid of mesh $1/(J2^j)$, so that $\mathcal Q_{J,0}=\mathcal Q_J$ and $|\mathcal Q_{J,j}|\le M_J2^{dj}$. Let $\pi_jx$ be the point of $\mathcal Q_{J,j}$ obtained by rounding every coordinate of $x$ down to this grid. These maps satisfy
\begin{equation}\label{eq:app-dyadic-maps}
\begin{gathered}
\pi_{j-1}(\pi_jx)=\pi_{j-1}x,\qquad
\|z-\pi_{j-1}z\|_2\le h_J2^{-j}\ \text{ for }z\in\mathcal Q_{J,j},\\
\|x-\pi_0x\|_2\le h_J,\qquad
\pi_jx\to x.
\end{gathered}
\end{equation}

\runinlabel{Coarse grid.}
The Gaussian tail bound and a union bound over $\mathcal Q_J$ give $\mathbb P\bigl(|U(z)|>u\,\sigma(z)\text{ for some }z\in\mathcal Q_J\bigr)\le2M_Je^{-u^2/2}=\delta/2$.

\runinlabel{Refinements.}
For $j\ge1$, set $a_j=\sqrt{u^2+2(d+1)j\log2}$. By the assumption on the increments and \eqref{eq:app-dyadic-maps}, the increment $U(z)-U(\pi_{j-1}z)$ with $z\in\mathcal Q_{J,j}$ has standard deviation at most $Lh_J^\zeta2^{-j\zeta}$. A union bound over $\mathcal Q_{J,j}$ therefore gives
\begin{equation}\label{eq:app-dyadic-failure}
\mathbb P\bigl(|U(z)-U(\pi_{j-1}z)|>Lh_J^\zeta2^{-j\zeta}a_j\text{ for some }z\in\mathcal Q_{J,j}\bigr)
\le2M_J2^{dj}e^{-a_j^2/2}=\frac\delta2\,2^{-j}.
\end{equation}
The failure probabilities of the levels $j\ge1$ sum to at most $\delta/2$.

\runinlabel{Extension to the continuum.}
The exceptional events above have total probability at most $\delta$. On their complement, telescoping along $\pi_0x,\pi_1x,\ldots$ and the continuity of the paths give, for every $x\in[0,1]^d$,
\begin{equation}\label{eq:app-chaining-sum}
|U(x)-U(\pi_0x)|
\le Lh_J^\zeta\sum_{j=1}^\infty2^{-j\zeta}a_j
\le Lh_J^\zeta\Bigl[u\,\frac{\rho_\zeta}{1-\rho_\zeta}+C_2(d,\zeta)\Bigr],
\end{equation}
where the second inequality uses $a_j\le u+j\sqrt{2(d+1)\log2}$. No union bound over an uncountable set is taken. Moreover, $\sigma(\pi_0x)\le\sigma(x)+Lh_J^\zeta$ by the triangle inequality for standard deviations. Combining these bounds with $|U(\pi_0x)|\le u\,\sigma(\pi_0x)$ and $1+\rho_\zeta/(1-\rho_\zeta)=C_1(\zeta)$ proves \eqref{eq:app-chaining-bound}.
\end{proof}

\begin{proof}[Proof of Lemma~\ref{lemma:unif-envelope-continuous}]
Fix $1\le n\le N$. Conditionally on $\mathcal F_{n-1}$, the residual $Z_n$ is a centered Gaussian process with continuous paths and standard deviation $s_{n-1}$, and \eqref{eq:app-increment-contraction} together with Assumption~\ref{asmp:canonical-metric} gives $\operatorname{Var}(Z_n(x)-Z_n(x')\mid\mathcal F_{n-1})\le L_k^2\|x-x'\|_2^{2\zeta}$. We apply Lemma~\ref{lemma:app-chaining} conditionally, with $L=L_k$ and with $\delta/(2N)$ in place of $\delta$. Its coefficient is then $\sqrt{2\log(8NM_J/\delta)}=u_{N,J,\delta}^{\mathrm c}$, and the bound \eqref{eq:continuous-envelope} for round $n$ fails with conditional probability at most $\delta/(2N)$. Taking expectations and a union bound over the $N$ rounds proves the lemma.
\end{proof}

\subsection{Corollary~\ref{corollary:Matern-continuous}: Continuous Mat\'ern Rate}\label{app:cor-matern-continuous}
\begin{proof}
Proposition~\ref{proposition:Matern-kernel-asmps} gives $\alpha=\nu/d$ and an admissible $\zeta$. In Theorem~\ref{thm:simple-regret-continuous-general}, choose $J=J_N$ from \eqref{eq:N-grid}. This grid enters only the analysis. Then $h_{J_N}^\zeta\le N^{-\alpha}$, and the computation in \eqref{eq:app-grid-log} gives $u_{N,J_N,\delta}^{\mathrm c}\le C\sqrt{\log(N/\delta)}$ for $N\ge2$. With $\eta_n=0$, both terms in \eqref{eq:ts-continuous-bound} are therefore at most a constant multiple of $N^{-\alpha}\sqrt{\log(N/\delta)}$, which proves the high-probability bound. The expected bound follows as in \eqref{eq:app-raw-expectation} with $\delta_N=N^{-(\alpha+1)}$.
\end{proof}

\section{Proofs for Section~\ref{sec:log-factor-GP-TS}}\label{app:log-factor}
This appendix proves Lemmas~\ref{lemma:lower-plateau} and~\ref{lemma:lower-local} and Theorem~\ref{thm:matern-log-lower}.

\runinlabel{Setting and notation.}
Throughout this appendix, $k$ is the Mat\'ern-$5/2$ kernel \eqref{eq:lower-matern-kernel}. We write $k(x,y)=\varphi(x-y)$ with the even profile
\begin{equation}\label{eq:lb-profile}
\varphi(t)=\left(1+|t|+\frac{t^2}{3}\right)e^{-|t|},\qquad t\in\mathbb R.
\end{equation}
The objective is $f_N=f_{\varepsilon_N}$ from \eqref{eq:lower-objective-family}, with $\Delta_N=f_N^*=\kappa\varepsilon_N^2$. The decision domain $\mathcal D_N$ is either $[0,1]$ or the grid $\mathcal G_N^{\mathrm{lb}}$ in \eqref{eq:lower-decision-grid}, whose mesh is $\gamma_N$. In order to treat both implementations at once, we set $\gamma_N=0$ for the continuous implementation. With this convention, every point of $[0,1]$ lies within distance $\gamma_N/2$ of a point of $\mathcal D_N$ in both cases, and \eqref{eq:lower-resolution} provides a constant $C_\gamma$ such that $\gamma_N\le C_\gamma N^{-5/4}$ for all $N$. The algorithm is represented as in \eqref{eq:lower-sampling}. The sequence of actual queries is $X_n=(x_1,\ldots,x_n)$, and $\mathcal F_n$ is the history after round $n$, which contains the sampled paths, the queries, and the exact observations of the first $n$ rounds. All probabilities and expectations are taken with respect to the randomness of the algorithm for the fixed objective $f_N$, and we abbreviate $\mathbb P=\mathbb P_{f_N}$ and $\mathbb E=\mathbb E_{f_N}$. The boundary event is $\mathcal E_\partial=\{x_1=0\}\in\mathcal F_1$.

For a function $u$ on an interval $[a,b]$, we write $\|u\|_{\infty,[a,b]}=\sup_{x\in[a,b]}|u(x)|$ and $\|u\|_\infty=\|u\|_{\infty,[0,1]}$. The distance from a point $x$ to a finite collection $X$ of points is $\operatorname{dist}(x,X)=\min_{y\in X}|x-y|$. The letters $C$ and $c$ denote positive constants whose values may change from one occurrence to the next. All constants, named or not, may depend on $B$, $v$, $\chi$, $\kappa$, $C_\gamma$, and the two auxiliary functions $\psi$ and $\theta$ fixed in Appendix~\ref{app:lb-rkhs}. They never depend on $N$, on the round $n$, on the admissible $\varepsilon_N$, or on the realized history. Constants that carry a subscript or a star, such as $C_\psi$, $c_0$, or $b_*$, keep their values once they are introduced.

\subsection{Preliminary Propositions}\label{app:lb-preliminary-results}

\subsubsection{Deterministic Estimates in the Mat\'ern RKHS}\label{app:lb-rkhs}
Let $\mathcal H_k(\mathbb R)$ denote the RKHS of the stationary kernel $(x,y)\mapsto\varphi(x-y)$ on $\mathbb R$, and let $\mathcal H_k$ denote, as before, the RKHS of its restriction $k$ to $[0,1]$. For an open interval $(a,b)$, which may be all of $\mathbb R$, the Sobolev space $H^3(a,b)$ consists of the functions whose weak derivatives up to order three belong to $L^2(a,b)$. Every element of $H^3(a,b)$ has a twice continuously differentiable representative, which we always use, and its second derivative satisfies $u''(x)-u''(y)=\int_y^xu'''(t)\,dt$. We write $\|g\|_2$ for the norm of $L^2(\mathbb R)$ and use the Fourier transform $\widehat g(\omega)=\int_{\mathbb R}g(x)e^{-i\omega x}\,dx$. For $r\in\{0,1,2\}$ and $x\in[0,1]$, we set $k_x^{(r)}=\partial_1^rk(x,\cdot)$, the derivative of order $r$ of the kernel with respect to its first argument, so that $k_x^{(0)}=k_x$. At the endpoints of $[0,1]$, derivatives are taken from inside the interval.

\begin{lbproposition}[RKHS norm and derivative evaluation]\label{proposition:lb-norm}
The kernel \eqref{eq:lower-matern-kernel} has the following properties.
\begin{enumerate}
\item[(i)] The space $\mathcal H_k(\mathbb R)$ coincides with $H^3(\mathbb R)$, and for every $g\in\mathcal H_k(\mathbb R)$,
\begin{equation}\label{eq:lb-fourier-norm}
\|g\|_{\mathcal H_k(\mathbb R)}^2
=\frac3{16}\left(\|g\|_2^2+3\|g'\|_2^2+3\|g''\|_2^2+\|g'''\|_2^2\right).
\end{equation}
\item[(ii)] The space $\mathcal H_k$ consists of the restrictions to $[0,1]$ of the functions in $\mathcal H_k(\mathbb R)$, and
\begin{equation}\label{eq:lb-restriction-norm}
\|u\|_{\mathcal H_k}
=\min\bigl\{\|g\|_{\mathcal H_k(\mathbb R)}:
g\in\mathcal H_k(\mathbb R),\ g=u\text{ on }[0,1]\bigr\}.
\end{equation}
In particular, $\mathcal H_k\subseteq H^3(0,1)$ and
\begin{equation}\label{eq:lb-third-derivative}
\|u'''\|_{L^2(0,1)}\le\frac4{\sqrt3}\|u\|_{\mathcal H_k},
\qquad u\in\mathcal H_k.
\end{equation}
\item[(iii)] For $r\in\{0,1,2\}$ and $x\in[0,1]$, the function $k_x^{(r)}$ belongs to $\mathcal H_k$ and
\begin{equation}\label{eq:lb-derivative-reproducing}
u^{(r)}(x)=\langle u,k_x^{(r)}\rangle_{\mathcal H_k},
\qquad u\in\mathcal H_k.
\end{equation}
Moreover, for $x,y\in[0,1]$,
\begin{equation}\label{eq:lb-representer-norms}
\|k_x^{(1)}\|_{\mathcal H_k}^2=\frac13,\qquad
\|k_x^{(2)}\|_{\mathcal H_k}^2=1,\qquad
\|k_x^{(2)}-k_y^{(2)}\|_{\mathcal H_k}^2\le C_k|x-y|,
\end{equation}
with $C_k=16/3$. Consequently, $\|u''\|_\infty\le\|u\|_{\mathcal H_k}$ for every $u\in\mathcal H_k$.
\end{enumerate}
\end{lbproposition}
\begin{proof}
For (i), integrating the three terms of \eqref{eq:lb-profile} gives $\widehat\varphi(\omega)=16/(3(1+\omega^2)^3)$, which is positive and integrable. The RKHS of such a stationary kernel consists of the functions $g\in L^2(\mathbb R)$ for which $\|g\|_{\mathcal H_k(\mathbb R)}^2=(2\pi)^{-1}\int_{\mathbb R}|\widehat g(\omega)|^2/\widehat\varphi(\omega)\,d\omega$ is finite \citep[Lemma~1]{JMLR:v12:bull11a}. The constant can be checked directly, because the Fourier transform of $\varphi(\cdot-x)$ is $e^{-i\omega x}\widehat\varphi(\omega)$, so that the inner product of $g$ with this function equals $(2\pi)^{-1}\int\widehat g(\omega)e^{i\omega x}\,d\omega=g(x)$. Expanding $(1+\omega^2)^3$ and applying Plancherel's identity $\|g^{(j)}\|_2^2=(2\pi)^{-1}\int\omega^{2j}|\widehat g(\omega)|^2\,d\omega$ gives \eqref{eq:lb-fourier-norm}, whose right side is finite exactly when $g\in H^3(\mathbb R)$.

For (ii), the description of $\mathcal H_k$ as a space of restrictions with the norm \eqref{eq:lb-restriction-norm} is the restriction property of an RKHS \citep[Lemma~2]{JMLR:v12:bull11a}. Every extension $g$ of $u$ satisfies $\|u'''\|_{L^2(0,1)}\le\|g'''\|_2\le(4/\sqrt3)\|g\|_{\mathcal H_k(\mathbb R)}$ by \eqref{eq:lb-fourier-norm}, and minimizing over $g$ gives \eqref{eq:lb-third-derivative}.

For (iii), the profile $\varphi$ is four times continuously differentiable, and so is the kernel on $[0,1]^2$. The derivative reproducing property \citep[Theorem~1]{zhou2008derivative} then shows that $k_x^{(r)}$ belongs to $\mathcal H_k$ and satisfies \eqref{eq:lb-derivative-reproducing}. Applying \eqref{eq:lb-derivative-reproducing} to the function $k_y^{(r)}$ yields $\langle k_x^{(r)},k_y^{(r)}\rangle_{\mathcal H_k}=\partial_x^r\partial_y^rk(x,y)=(-1)^r\varphi^{(2r)}(x-y)$. Direct differentiation of \eqref{eq:lb-profile} gives, for $t\ge0$,
\begin{equation}\label{eq:lb-profile-derivatives}
\varphi''(t)=\frac{t^2-t-1}{3}e^{-t},\qquad
\varphi^{(4)}(t)=\frac{t^2-5t+3}{3}e^{-t},
\end{equation}
and both derivatives are even functions. Hence $\|k_x^{(1)}\|_{\mathcal H_k}^2=-\varphi''(0)=1/3$ and $\|k_x^{(2)}\|_{\mathcal H_k}^2=\varphi^{(4)}(0)=1$. Moreover, $\|k_x^{(2)}-k_y^{(2)}\|_{\mathcal H_k}^2=2[1-\varphi^{(4)}(|x-y|)]$, and the derivative of $\varphi^{(4)}$ is bounded by $8/3$ in absolute value on $[0,1]$, which gives the last bound in \eqref{eq:lb-representer-norms}. Finally, the Cauchy-Schwarz inequality gives $|u''(x)|\le\|u\|_{\mathcal H_k}\|k_x^{(2)}\|_{\mathcal H_k}=\|u\|_{\mathcal H_k}$.
\end{proof}

In order to bound a posterior uncertainty from below, we will exhibit a function of controlled norm that vanishes at all observations. For this purpose we fix, once and for all, two smooth functions on $\mathbb R$, namely a bump $\psi$ supported in $(-1/4,1/4)$ with $\psi(0)=1$ and a cutoff $\theta$ supported in $(-1,1)$ with $\theta(0)=1$. For $0<h\le1$, $z\in\mathbb R$, and $0<R\le1$, define
\begin{equation}\label{eq:lb-test-functions}
u_{h,z}(x)=h^{5/2}\psi\left(\frac{x-z}{h}\right),\qquad
\vartheta_R(x)=x\,\theta\left(\frac xR\right).
\end{equation}
A function defined on $\mathbb R$ is identified with its restriction to $[0,1]$ whenever it is regarded as an element of $\mathcal H_k$.

\begin{lbproposition}[Local test functions]\label{proposition:lb-test-functions}
Let
\[
C_\psi=\frac3{16}\left(\|\psi\|_2^2+3\|\psi'\|_2^2+3\|\psi''\|_2^2+\|\psi'''\|_2^2\right),
\]
and let $C_\theta$ denote the same expression with $\psi$ replaced by the function $t\mapsto t\,\theta(t)$.
\begin{enumerate}
\item[(i)] The bump $u_{h,z}$ is supported in $(z-h/4,z+h/4)$ and satisfies $u_{h,z}(z)=h^{5/2}$. For finitely many bumps $u_{h_i,z_i}$ with pairwise disjoint supports and real coefficients $a_i$,
\begin{equation}\label{eq:lb-bump-sum}
\Bigl\|\sum_ia_iu_{h_i,z_i}\Bigr\|_{\mathcal H_k}^2\le C_\psi\sum_ia_i^2.
\end{equation}
In particular, $\|u_{h,z}\|_{\mathcal H_k}\le C_\psi^{1/2}$.
\item[(ii)] The function $\vartheta_R$ is supported in $(-R,R)$ and satisfies $\vartheta_R(0)=0$, $\vartheta_R'(0)=1$, and $\|\vartheta_R\|_{\mathcal H_k}\le C_\theta^{1/2}R^{-3/2}$.
\end{enumerate}
\end{lbproposition}
\begin{proof}
Differentiation and the change of variables $t=(x-z)/h$ give $\|u_{h,z}^{(j)}\|_2^2=h^{6-2j}\|\psi^{(j)}\|_2^2$ for $0\le j\le3$. Since $h\le1$, the identity \eqref{eq:lb-fourier-norm} yields $\|u_{h,z}\|_{\mathcal H_k(\mathbb R)}^2\le C_\psi$. If the supports are pairwise disjoint, the squared norms in $L^2(\mathbb R)$ of each derivative add, so the same identity gives \eqref{eq:lb-bump-sum} with the norm of $\mathcal H_k(\mathbb R)$ on the left side. Restriction to $[0,1]$ does not increase the norm by \eqref{eq:lb-restriction-norm}. For (ii), write $\vartheta_R(x)=R\,q(x/R)$ with $q(t)=t\,\theta(t)$. The same computation gives $\|\vartheta_R^{(j)}\|_2^2=R^{3-2j}\|q^{(j)}\|_2^2$, and $R\le1$ yields $\|\vartheta_R\|_{\mathcal H_k(\mathbb R)}^2\le C_\theta R^{-3}$. The values $\vartheta_R(0)=0$ and $\vartheta_R'(0)=\theta(0)=1$ are immediate.
\end{proof}

Conversely, three zeros suffice to control a function and its first two derivatives by the third derivative.

\begin{lbproposition}[Estimate from three zeros]\label{proposition:lb-three-zeros}
Let $a<b$ and let $u\in H^3(a,b)$ vanish at three distinct points of $[a,b]$. Then
\begin{equation}\label{eq:lb-three-zeros}
\|u^{(r)}\|_{\infty,[a,b]}
\le(b-a)^{5/2-r}\|u'''\|_{L^2(a,b)},\qquad r=0,1,2.
\end{equation}
\end{lbproposition}
\begin{proof}
By Rolle's theorem, $u'$ has two distinct zeros in $[a,b]$, and $u''$ has a zero $x_0\in[a,b]$. For $x\in[a,b]$, the Cauchy-Schwarz inequality gives $|u''(x)|=\bigl|\int_{x_0}^xu'''(t)\,dt\bigr|\le(b-a)^{1/2}\|u'''\|_{L^2(a,b)}$. Integrating this bound from a zero of $u'$ gives the case $r=1$, and integrating once more from a zero of $u$ gives the case $r=0$.
\end{proof}

We now connect these estimates with the posterior. Let $X\subset[0,1]$ be a finite set of observation points. Repeated queries do not change the posterior, so only the set of distinct points matters. Extending the notation of Section~\ref{sec:background}, let $W_X=\operatorname{span}\{k_x:x\in X\}$, let $P_X$ be the orthogonal projection of $\mathcal H_k$ onto $W_X$, and define the residual space $\mathcal R_X=W_X^\perp=\{u\in\mathcal H_k:u(x)=0\text{ for every }x\in X\}$, where the second equality follows from the reproducing property. The posterior covariance and the posterior standard deviations of the function and of its first two derivatives are
\begin{equation}\label{eq:lb-derivative-uncertainty}
c_X(x,y)=\langle(I-P_X)k_x,(I-P_X)k_y\rangle_{\mathcal H_k},\quad
s_X^{(r)}(x)=\|(I-P_X)k_x^{(r)}\|_{\mathcal H_k},\ \ r\in\{0,1,2\},
\end{equation}
and we write $s_X=s_X^{(0)}$. Proposition~\ref{proposition:lb-paths} below justifies the interpretation of $s_X^{(r)}$ as a standard deviation. The posterior mean of an objective $f\in\mathcal H_k$ is $\mu_X=P_Xf$. When $X$ is the set of points of the history $X_{n-1}$, these objects are $P_{n-1}$, $c_{n-1}$, $s_{n-1}$, and $\mu_{n-1}$, and we write $s_{n-1}^{(r)}$ and $\mathcal R_{n-1}$ accordingly.

\begin{lbproposition}[Residual space and posterior uncertainty]\label{proposition:lb-residual}
Let $X\subset[0,1]$ be finite, and set $c_\psi=C_\psi^{-1/2}$ and $c_\theta=C_\theta^{-1/2}$.
\begin{enumerate}
\item[(i)] \emph{Variational identity.} For $x\in[0,1]$ and $r\in\{0,1,2\}$,
\begin{equation}\label{eq:lb-residual-variational}
s_X^{(r)}(x)
=\sup\bigl\{|u^{(r)}(x)|:u\in\mathcal R_X,\ \|u\|_{\mathcal H_k}\le1\bigr\}.
\end{equation}
\item[(ii)] \emph{Empty interval.} If $z\in[0,1]$, $0<h\le1$, and the interval $(z-h/4,z+h/4)$ contains no point of $X$, then $s_X(z)\ge c_\psi h^{5/2}$.
\item[(iii)] \emph{Empty boundary layer.} If $0<R\le1$ and the interval $(0,R)$ contains no point of $X$, then $s_X^{(1)}(0)\ge c_\theta R^{3/2}$.
\item[(iv)] \emph{Three observations.} If an interval $[a,a+h]\subseteq[0,1]$ contains three distinct points of $X$, then
\begin{equation}\label{eq:lb-three-point-variance}
\sup_{x\in[a,a+h]}s_X^{(2)}(x)^2\le\frac{16}{3}h.
\end{equation}
\item[(v)] \emph{Posterior mean.} If $f\in\mathcal H_k$ and $\|f\|_{\mathcal H_k}\le B$, then $\mu_X$ coincides with $f$ on $X$ and
\begin{equation}\label{eq:lb-mean-bounds}
\|\mu_X\|_{\mathcal H_k}\le B,\qquad
\|\mu_X''\|_\infty\le B,\qquad
|f(x)-\mu_X(x)|\le Bs_X(x)\quad\text{for }x\in[0,1].
\end{equation}
\end{enumerate}
\end{lbproposition}
\begin{proof}
For $u\in\mathcal R_X$, the identity \eqref{eq:lb-derivative-reproducing} and the orthogonality of $u$ to $W_X$ give $u^{(r)}(x)=\langle u,(I-P_X)k_x^{(r)}\rangle_{\mathcal H_k}$. The Cauchy-Schwarz inequality shows that the supremum in \eqref{eq:lb-residual-variational} is at most $s_X^{(r)}(x)$. The function $(I-P_X)k_x^{(r)}$ belongs to $\mathcal R_X$, and its normalization attains this value when it is nonzero. This proves (i). For (ii), the bump $u_{h,z}$ vanishes on $X$ and therefore belongs to $\mathcal R_X$. Applying (i) with $r=0$ to its normalization and using Proposition~\ref{proposition:lb-test-functions} gives $s_X(z)\ge u_{h,z}(z)/\|u_{h,z}\|_{\mathcal H_k}\ge c_\psi h^{5/2}$. For (iii), the function $\vartheta_R$ vanishes at zero and on $[R,1]$, so it belongs to $\mathcal R_X$, and (i) with $r=1$ gives $s_X^{(1)}(0)\ge \vartheta_R'(0)/\|\vartheta_R\|_{\mathcal H_k}\ge c_\theta R^{3/2}$. For (iv), every $u\in\mathcal R_X$ has three distinct zeros in $[a,a+h]$. Proposition~\ref{proposition:lb-three-zeros} and \eqref{eq:lb-third-derivative} give $\|u''\|_{\infty,[a,a+h]}\le h^{1/2}\|u'''\|_{L^2(a,a+h)}\le(4/\sqrt3)h^{1/2}\|u\|_{\mathcal H_k}$, and taking the supremum over the unit ball of $\mathcal R_X$ in (i) proves \eqref{eq:lb-three-point-variance}. For (v), the function $f-\mu_X=(I-P_X)f$ belongs to $\mathcal R_X$ and thus vanishes on $X$, and the bound $|f-\mu_X|\le Bs_X$ is \eqref{eq:deterministic-upper-bound}. Since $P_X$ is an orthogonal projection, $\|\mu_X\|_{\mathcal H_k}\le B$, and Proposition~\ref{proposition:lb-norm} then gives $\|\mu_X''\|_\infty\le B$.
\end{proof}

\subsubsection{Gaussian Estimates for a Posterior Sample}\label{app:lb-gaussian}
In this part the finite set $X$ of observation points is fixed. Its Gram matrix and kernel vector are denoted by $K_X$ and $\mathbf k_X(x)$, as in Section~\ref{sec:background}, and $\Xi\sim\mathcal{GP}(0,k)$ is a prior path. Following \eqref{eq:app-exact-posterior-path}, the posterior residual is
\begin{equation}\label{eq:lb-posterior-residual}
Z_X(x)=\Xi(x)-\mathbf k_X(x)^\top K_X^\dagger\,\Xi(X),
\end{equation}
where $\Xi(X)$ is the vector of the values of $\Xi$ at the points of $X$. It is a centered Gaussian process with covariance $c_X$. When $X$ is the set of points of $X_{n-1}$ and $\Xi$ is drawn independently of $\mathcal F_{n-1}$, the process $Z_X$ has the conditional law of $Z_n$ given $\mathcal F_{n-1}$. Every statement of this part therefore holds for $Z_n$ conditionally on $\mathcal F_{n-1}$.

\begin{lbproposition}[Regularity of posterior paths]\label{proposition:lb-paths}
The prior path $\Xi$ has a version whose paths are twice continuously differentiable on $[0,1]$. With this version, for every finite $X\subset[0,1]$, the paths of $Z_X$ are twice continuously differentiable, and for $r,s\in\{0,1,2\}$ and $x,y\in[0,1]$,
\begin{equation}\label{eq:lb-derivative-covariance}
\operatorname{Cov}\bigl(Z_X^{(r)}(x),Z_X^{(s)}(y)\bigr)
=\bigl\langle(I-P_X)k_x^{(r)},(I-P_X)k_y^{(s)}\bigr\rangle_{\mathcal H_k}.
\end{equation}
In particular, $\operatorname{Var}(Z_X^{(r)}(x))=s_X^{(r)}(x)^2$, and
\begin{equation}\label{eq:lb-derivative-increments}
\operatorname{Var}(Z_X''(x))\le1,\qquad
\operatorname{Var}(Z_X''(x)-Z_X''(y))\le C_k|x-y|.
\end{equation}
\end{lbproposition}
\begin{proof}
Since $\varphi$ is four times continuously differentiable, $\Xi$ has mean square derivatives $\Xi_1$ and $\Xi_2$ of orders one and two. The processes $\Xi$, $\Xi_1$, and $\Xi_2$ are jointly Gaussian, and their covariances are obtained by differentiating the kernel \citep[Sections~4.1 and~9.4]{DBLP:books/lib/RasmussenW06}. In particular, $\mathbb E|\Xi_2(x)-\Xi_2(y)|^2=2[1-\varphi^{(4)}(|x-y|)]\le C_k|x-y|$, so the Kolmogorov-Chentsov theorem provides a version of $\Xi_2$ with continuous paths. Integrating this version twice yields a version of $\Xi$ whose paths are twice continuously differentiable and whose derivatives are versions of $\Xi_1$ and $\Xi_2$ \citep{da2026sample}. From now on $\Xi$ denotes this version. The kernel sections $k(x_i,\cdot)$ are four times continuously differentiable, so the paths of $Z_X$ in \eqref{eq:lb-posterior-residual} are twice continuously differentiable as well.

For the covariances, let $\mathcal J$ be the linear isometry from $\mathcal H_k$ into the space of square integrable random variables that is determined by $\mathcal Jk_x=\Xi(x)$. The difference quotients of $x\mapsto k_x$ and of $x\mapsto k_x^{(1)}$ converge in $\mathcal H_k$ to $k_x^{(1)}$ and $k_x^{(2)}$, hence $\mathcal Jk_x^{(r)}=\Xi^{(r)}(x)$ almost surely. The identity $\mu_X=P_Xf$ and the formula \eqref{eq:gp-posterior} for the posterior mean, applied to the function $f=k_x^{(r)}$ whose values on $X$ form the vector $\partial_x^r\mathbf k_X(x)$, show that $P_Xk_x^{(r)}$ is the combination of the sections $k_{x_i}$, $x_i\in X$, with the coefficient vector $K_X^\dagger\partial_x^r\mathbf k_X(x)$. Comparing with the derivative of order $r$ of \eqref{eq:lb-posterior-residual} gives $\mathcal J(I-P_X)k_x^{(r)}=Z_X^{(r)}(x)$ almost surely, and \eqref{eq:lb-derivative-covariance} follows because $\mathcal J$ preserves inner products. Finally, $I-P_X$ is an orthogonal projection, so \eqref{eq:lb-derivative-increments} follows from \eqref{eq:lb-derivative-covariance} and \eqref{eq:lb-representer-norms}.
\end{proof}

\begin{lbproposition}[Supremum of the second derivative]\label{proposition:lb-derivative-supremum}
There is a numerical constant $C_{\sup}$ such that, for every finite $X\subset[0,1]$ and every $0<\delta<1$,
\begin{equation}\label{eq:lb-derivative-tail}
\mathbb P\Bigl(\|Z_X''\|_\infty
>C_{\sup}\bigl(1+\sqrt{\log(1/\delta)}\bigr)\Bigr)\le\delta.
\end{equation}
If, in addition, an interval $[a,a+h]\subseteq[0,1]$ contains three distinct points of $X$, then
\begin{equation}\label{eq:lb-local-derivative-tail}
\mathbb P\Bigl(\|Z_X''\|_{\infty,[a,a+h]}
>C_{\sup}\sqrt h\,\bigl(1+\sqrt{\log(1/\delta)}\bigr)\Bigr)\le\delta.
\end{equation}
\end{lbproposition}
\begin{proof}
Both bounds follow from the chaining bound of Lemma~\ref{lemma:app-chaining}. Let $U$ be a centered Gaussian process on $[0,1]$ with continuous paths such that $\operatorname{Var}(U(t))\le V_0^2$ and $\operatorname{Var}(U(t)-U(t'))\le V_1^2|t-t'|$ for all $t,t'\in[0,1]$. Lemma~\ref{lemma:app-chaining} with $d=1$, $\zeta=1/2$, $L=V_1$, and $J=1$ gives $\|U\|_\infty\le V_0u+V_1[C_1(1/2)\,u+C_2(1,1/2)]$ with probability at least $1-\delta$, where $u=\sqrt{2\log(8/\delta)}\le\sqrt{2\log8}+\sqrt{2\log(1/\delta)}$. Hence there is a numerical constant $C_{\mathrm{ch}}$ such that
\begin{equation}\label{eq:lb-chaining}
\mathbb P\Bigl(\|U\|_\infty
>C_{\mathrm{ch}}(V_0+V_1)\bigl(1+\sqrt{\log(1/\delta)}\bigr)\Bigr)\le\delta.
\end{equation}

By Proposition~\ref{proposition:lb-paths}, the process $U=Z_X''$ satisfies the assumptions with $V_0=1$ and $V_1=C_k^{1/2}$, which gives \eqref{eq:lb-derivative-tail}. Under the additional assumption, Proposition~\ref{proposition:lb-residual} gives $\operatorname{Var}(Z_X''(x))=s_X^{(2)}(x)^2\le16h/3$ on $[a,a+h]$. The rescaled process $U(t)=h^{-1/2}Z_X''(a+ht)$, $t\in[0,1]$, therefore satisfies the assumptions with $V_0=(16/3)^{1/2}$ and $V_1=C_k^{1/2}$, and \eqref{eq:lb-chaining} gives \eqref{eq:lb-local-derivative-tail}. Both bounds hold with $C_{\sup}=C_{\mathrm{ch}}(1+2C_k^{1/2})$.
\end{proof}

For symmetric matrices, $A\preceq A'$ means that $A'-A$ is positive semidefinite, and $I_K$ denotes the identity matrix of size $K$.

\begin{lbproposition}[Gaussian tail and maximum]\label{proposition:lb-gaussian-max}
Let $c_{\mathrm{tail}}=e^{-3/2}/\sqrt{2\pi}$.
\begin{enumerate}
\item[(i)] For every $t\ge1$,
\begin{equation}\label{eq:lb-normal-tail}
1-\Phi(t)\ge c_{\mathrm{tail}}\,t^{-1}e^{-t^2/2}.
\end{equation}
\item[(ii)] Let $K\ge3$ and let $V\sim\mathcal N(0,\Sigma)$ be a Gaussian vector in $\mathbb R^K$ such that $c_0I_K\preceq\Sigma\preceq C_0I_K$ for constants $0<c_0\le C_0$. Then
\begin{equation}\label{eq:lb-gaussian-max}
\mathbb P\Bigl(\max_{1\le i\le K}V_i
\le\sqrt{c_0\log K}-\sqrt{8C_0}\Bigr)
\le e^{-K/2}
+\exp\Bigl(-\frac{c_{\mathrm{tail}}}{2}\sqrt{\frac{K}{\log K}}\Bigr).
\end{equation}
\end{enumerate}
\end{lbproposition}
\begin{proof}
For (i), the standard normal density $\Phi'$ is decreasing on $[0,\infty)$, so $1-\Phi(t)\ge\int_t^{t+1/t}\Phi'(y)\,dy\ge t^{-1}\Phi'(t+1/t)=(t\sqrt{2\pi})^{-1}\exp\bigl(-t^2/2-1-1/(2t^2)\bigr)$, and $t\ge1$ gives \eqref{eq:lb-normal-tail}.

For (ii), the lower bound on $\Sigma$ allows us to split off independent coordinates. Let $\xi\sim\mathcal N(0,I_K)$ and $Y\sim\mathcal N(0,\Sigma-c_0I_K)$ be independent. The vector $\sqrt{c_0}\,\xi+Y$ has the law of $V$, so it suffices to prove the bound for this vector. Since $0\preceq\Sigma-c_0I_K\preceq C_0I_K$, diagonalizing this matrix and using $\mathbb E\exp(a\xi_1^2)=(1-2a)^{-1/2}$ for $a<1/2$ gives
\[
\mathbb E\exp\Bigl(\frac{\|Y\|_2^2}{4C_0}\Bigr)\le2^{K/2},
\qquad
\mathbb P\bigl(\|Y\|_2^2>4C_0K\bigr)\le e^{-K}2^{K/2}\le e^{-K/2},
\]
where the second bound is Markov's inequality. On the event $\{\|Y\|_2^2\le4C_0K\}$, fewer than $K/2$ coordinates satisfy $Y_i<-\sqrt{8C_0}$, so the set $I_Y=\{i:Y_i\ge-\sqrt{8C_0}\}$ has at least $K/2$ elements. Conditionally on $Y$, the variables $\xi_i$ with $i\in I_Y$ are independent standard normal variables. By (i) with $t=\sqrt{\log K}\ge1$,
\[
\mathbb P\Bigl(\max_{i\in I_Y}\xi_i\le\sqrt{\log K}\,\Big|\,Y\Bigr)
=\Phi\bigl(\sqrt{\log K}\bigr)^{|I_Y|}
\le\Bigl(1-\frac{c_{\mathrm{tail}}}{\sqrt{K\log K}}\Bigr)^{K/2}
\le\exp\Bigl(-\frac{c_{\mathrm{tail}}}{2}\sqrt{\frac{K}{\log K}}\Bigr)
\]
on this event. Outside the two exceptional events, some index $i\in I_Y$ satisfies $\xi_i>\sqrt{\log K}$, and then $\sqrt{c_0}\,\xi_i+Y_i>\sqrt{c_0\log K}-\sqrt{8C_0}$.
\end{proof}

\begin{lbproposition}[Boundary event]\label{proposition:lb-boundary}
For both implementations,
\begin{equation}\label{eq:lb-boundary-probability}
\mathbb P(\mathcal E_\partial)\ge q_\partial
:=\tfrac12\Phi\bigl(-3e\sqrt3\bigr)>0.
\end{equation}
On $\mathcal E_\partial$, every $n\ge2$ satisfies $\mu_{n-1}(0)=0$, $s_{n-1}(0)=0$, and $G_n(0)=0$ almost surely.
\end{lbproposition}
\begin{proof}
In the first round, $G_1=v\,\Xi$ for a prior path $\Xi\sim\mathcal{GP}(0,k)$ with twice continuously differentiable paths. Let $S=\Xi'(0)$ and regress the derivative of the path on this variable,
\begin{equation}\label{eq:lb-boundary-regression}
\Xi'(x)=\lambda(x)S+R(x),\qquad
\lambda(x)=\frac{\operatorname{Cov}(\Xi'(x),S)}{\operatorname{Var}(S)}
=-3\varphi''(x)=(1+x-x^2)e^{-x},
\end{equation}
where the covariance is computed from Proposition~\ref{proposition:lb-paths} and \eqref{eq:lb-profile-derivatives}. The residual process $R$ is centered, jointly Gaussian with $S$, and uncorrelated with $S$ at every point. It is therefore independent of $S\sim\mathcal N(0,1/3)$, and so is $\|R\|_\infty$, which is a measurable function of countably many values of the continuous process $R$. Two facts about this decomposition are needed. First, $\lambda'(x)=x(x-3)e^{-x}\le0$ on $[0,1]$, so $\lambda(x)\ge\lambda(1)=e^{-1}$. Second, $R(0)=0$, and $R'(x)=\Xi''(x)-\lambda'(x)S$ is the residual of the regression of $\Xi''(x)$ on $S$, because $\lambda'(x)=\operatorname{Cov}(\Xi''(x),S)/\operatorname{Var}(S)$. Hence $\operatorname{Var}(R'(x))\le\operatorname{Var}(\Xi''(x))=1$ and $\mathbb E\|R\|_\infty\le\int_0^1\mathbb E|R'(x)|\,dx\le1$. Markov's inequality gives $\mathbb P(\|R\|_\infty\le2)\ge1/2$. On the intersection of the independent events $\{S\le-3e\}$ and $\{\|R\|_\infty\le2\}$, we have $\Xi'(x)\le-3+2=-1$ for every $x\in[0,1]$. The path $G_1$ is then strictly decreasing, and its only maximizer over either decision domain is zero. Since $\mathbb P(S\le-3e)=\Phi(-3e\sqrt3)$, this proves \eqref{eq:lb-boundary-probability}. On $\mathcal E_\partial$, the point zero is observed with the value $f_N(0)=0$. Interpolation gives $\mu_{n-1}(0)=0$ and $s_{n-1}(0)=0$ for $n\ge2$, so $Z_n(0)=0$ and $G_n(0)=0$ almost surely.
\end{proof}

\subsubsection{Tools for Adaptive Histories}\label{app:lb-adaptive}
The estimates above hold for a fixed set of observations, whereas the algorithm chooses its queries adaptively. A random point $z_n$ that is determined by the history before round $n$, that is, measurable with respect to $\mathcal F_{n-1}$, can be treated as fixed when the sample of round $n$ is drawn. Hence
\begin{equation}\label{eq:lb-past-marginal}
G_n(z_n)\mid\mathcal F_{n-1}
\sim\mathcal N\bigl(\mu_{n-1}(z_n),\,v^2s_{n-1}(z_n)^2\bigr).
\end{equation}
The query $x_n$, in contrast, depends on the sample of round $n$, and \eqref{eq:lb-past-marginal} does not apply to it. Sampled values at $x_n$ are instead controlled by an envelope that holds simultaneously at all points of the domain. For a deterministic integer $J\ge1$ and $0<\delta<1$, define the coefficients
\begin{equation}\label{eq:lb-envelope-coefficients}
u_{\mathrm{env}}(J,\delta)=\sqrt{2\log\frac{8(J+1)}{\delta}},\qquad
d_{\mathrm{env}}(J,\delta)=\frac{L_k}{J}
\bigl[2u_{\mathrm{env}}(J,\delta)+C_2(1,1)\bigr],
\end{equation}
where $L_k$ is the constant of Assumption~\ref{asmp:canonical-metric} for $\zeta=1$ and $C_2(1,1)$ is defined in \eqref{eq:continuous-envelope-constants}. Define also the envelope event
\begin{equation}\label{eq:lb-envelope-event}
\mathcal E^{\mathrm{env}}_n(J,\delta)
=\bigl\{|Z_n(x)|\le u_{\mathrm{env}}(J,\delta)\,s_{n-1}(x)
+d_{\mathrm{env}}(J,\delta)\text{ for every }x\in[0,1]\bigr\}.
\end{equation}
The integer $J$ is the resolution of a grid that is used in the analysis only. It is unrelated to the decision grid.

\begin{lbproposition}[Envelope for one round]\label{proposition:lb-envelope}
Let $n\ge1$, let $J\ge1$ be a deterministic integer, and let $0<\delta<1$.
\begin{enumerate}
\item[(i)] Almost surely, $\mathbb P\bigl(\mathcal E^{\mathrm{env}}_n(J,\delta)^c\mid\mathcal F_{n-1}\bigr)\le\delta/2$.
\item[(ii)] On $\mathcal E^{\mathrm{env}}_n(J,\delta)$, if some $z\in\mathcal D_N$ satisfies $G_n(z)\ge t$ for a real number $t$, then
\begin{equation}\label{eq:lb-excursion-innovation}
t\le\Delta_N+\bigl(B+vu_{\mathrm{env}}(J,\delta)\bigr)s_{n-1}(x_n)
+vd_{\mathrm{env}}(J,\delta).
\end{equation}
\end{enumerate}
\end{lbproposition}
\begin{proof}
Part (i) follows from Lemma~\ref{lemma:app-chaining}, applied conditionally on $\mathcal F_{n-1}$ to the process $Z_n$ with $d=1$, $\zeta=1$, $L=L_k$, and with $\delta/2$ in place of $\delta$. The assumption on the increments holds by \eqref{eq:app-increment-contraction} and Assumption~\ref{asmp:canonical-metric}. The coefficient of the lemma is then $u_{\mathrm{env}}(J,\delta)$, and its remainder equals $d_{\mathrm{env}}(J,\delta)$ because $h_J=1/J$ and $C_1(1)=2$. For (ii), the envelope and \eqref{eq:lb-mean-bounds} give, for every $x\in[0,1]$,
\[
G_n(x)=\mu_{n-1}(x)+vZ_n(x)
\le f_N(x)+\bigl(B+vu_{\mathrm{env}}(J,\delta)\bigr)s_{n-1}(x)
+vd_{\mathrm{env}}(J,\delta).
\]
Since $x_n$ maximizes $G_n$ over $\mathcal D_N$, we have $t\le G_n(z)\le G_n(x_n)$, and \eqref{eq:lb-excursion-innovation} follows from the display at $x=x_n$ and $f_N(x_n)\le\Delta_N$.
\end{proof}

\begin{lbproposition}[Counting along the trajectory]\label{proposition:lb-counting}
The following statements hold for both implementations.
\begin{enumerate}
\item[(i)] \emph{Large innovations are rare.} For every realized trajectory and every nonempty set $I\subseteq\{1,\ldots,N\}$ of rounds, some $n\in I$ satisfies $s_{n-1}(x_n)\le C_{\mathrm{sep}}|I|^{-5/2}$, where $C_{\mathrm{sep}}$ is the constant of Lemma~\ref{lemma:deterministic-bound-arbitrary-subseq}. Consequently, $\#\{1\le n\le N:s_{n-1}(x_n)\ge t\}\le(C_{\mathrm{sep}}/t)^{2/5}$ for every $t>0$.
\item[(ii)] \emph{Conditional successes accumulate.} Let $2\le m\le N$ and $p\in(0,1]$. For $2\le n\le m$, let $Q_n\in\{0,1\}$ be measurable with respect to $\mathcal F_n$ and satisfy $\mathbb E(Q_n\mid\mathcal F_{n-1})\ge p$ almost surely. Then $\mathbb P\bigl(\sum_{n=2}^mQ_n<p(m-1)/2\mid\mathcal E_\partial\bigr)\le e^{-p(m-1)/8}$.
\item[(iii)] \emph{Failures on a past event.} Let $n\ge2$, let $A\in\mathcal F_{n-1}$, and let $F$ be an event such that $\mathbb P(F\mid\mathcal F_{n-1})\le q$ on $A\cap\mathcal E_\partial$ for a deterministic number $q$. Then $\mathbb P(A\cap F\mid\mathcal E_\partial)\le q$.
\end{enumerate}
\end{lbproposition}
\begin{proof}
Part (i) is Lemma~\ref{lemma:deterministic-bound-arbitrary-subseq} with $\alpha=5/2$, the exponent supplied by Proposition~\ref{proposition:Matern-kernel-asmps} for $\nu=5/2$ and $d=1$. The counting bound follows by taking $I=\{n\le N:s_{n-1}(x_n)\ge t\}$. For (ii), the argument of Appendix~\ref{app:lemma-conditional}, started after the first round and carried out conditionally on $\mathcal F_1$, gives the stated bound with $\mathcal F_1$ in place of $\mathcal E_\partial$ almost surely, and averaging over the event $\mathcal E_\partial\in\mathcal F_1$ proves (ii). For (iii), the event $A\cap\mathcal E_\partial$ belongs to $\mathcal F_{n-1}$, so $\mathbb P(A\cap F\cap\mathcal E_\partial)=\mathbb E\bigl[\mathbf 1_{A\cap\mathcal E_\partial}\,\mathbb P(F\mid\mathcal F_{n-1})\bigr]\le q\,\mathbb P(\mathcal E_\partial)$, and dividing by $\mathbb P(\mathcal E_\partial)$ gives (iii).
\end{proof}

\subsection{Lemma~\ref{lemma:lower-plateau}: Persistent Overshoot on the Plateau}\label{app:lemma-lower-plateau}
Two quantities measure the coverage of the plateau after $m\ge1$ rounds, namely the largest posterior standard deviation on the plateau and the fill distance of the queries,
\begin{equation}\label{eq:lb-coverage-quantities}
\bar s_m=\max_{z\in\mathcal D_N\cap\mathcal P}s_m(z),\qquad
\rho_m=\sup_{x\in\mathcal P}\operatorname{dist}(x,X_m).
\end{equation}

\begin{proof}
We carry out the three steps in turn.

\runinlabel{Step 1. Coverage of the plateau.}
We show that there are constants $C_H$ and $c$ such that, with $H_m=C_Hm^{-1}[\log(m+1)]^{1/5}$,
\begin{equation}\label{eq:lb-actual-coverage}
\mathbb P(\rho_m>H_m\mid\mathcal E_\partial)\le e^{-c(m-1)},
\qquad2\le m\le N.
\end{equation}
Fix $2\le m\le N$. Before round $n$, let $z_n$ be a maximizer of $s_{n-1}$ over $\mathcal D_N\cap\mathcal P$, selected by a fixed measurable rule. It is determined by the history and satisfies $s_{n-1}(z_n)=\bar s_{n-1}$. Since $f_N(z_n)=0$, the bound \eqref{eq:lb-mean-bounds} gives $\mu_{n-1}(z_n)\ge-B\bar s_{n-1}$, and the marginal \eqref{eq:lb-past-marginal} yields
\begin{equation}\label{eq:lb-variance-optimism}
\mathbb P(G_n(z_n)\ge\bar s_{n-1}\mid\mathcal F_{n-1})
\ge p_*:=\Phi\bigl(-(B+1)/v\bigr).
\end{equation}
When $\bar s_{n-1}=0$, the event holds almost surely. Consider the envelope event $\mathcal E^{\mathrm{env}}_n(J,\delta)$ with $J=\lceil m^{7/2}\rceil$ and $\delta=p_*$, and abbreviate its coefficients by $u_{\mathrm{env}}$ and $d_{\mathrm{env}}$. By \eqref{eq:lb-envelope-coefficients}, they satisfy $u_{\mathrm{env}}\le C\sqrt{\log(m+1)}$ and $d_{\mathrm{env}}\le Cm^{-7/2}\sqrt{\log(m+1)}$. For $2\le n\le m$, define the indicator $Q_n^{\mathrm{cov}}=\mathbf 1\{G_n(z_n)\ge\bar s_{n-1}\}\,\mathbf 1_{\mathcal E^{\mathrm{env}}_n(J,\delta)}$.
By \eqref{eq:lb-variance-optimism} and Proposition~\ref{proposition:lb-envelope}, $\mathbb E(Q_n^{\mathrm{cov}}\mid\mathcal F_{n-1})\ge p_*-p_*/2=p_*/2$. Proposition~\ref{proposition:lb-counting} with $p=p_*/2$ shows that, outside an event of conditional probability at most $e^{-p_*(m-1)/16}$ given $\mathcal E_\partial$, the set $I_m=\{2\le n\le m:Q_n^{\mathrm{cov}}=1\}$ has at least $p_*(m-1)/4$ elements. For every $n\in I_m$, the monotonicity \eqref{eq:app-projection-monotonicity} of the posterior standard deviation and Proposition~\ref{proposition:lb-envelope} with $z=z_n$ and $t=\bar s_{n-1}$ give
\begin{equation}\label{eq:lb-variance-comparison}
\bar s_m\le\bar s_{n-1}
\le\Delta_N+(B+vu_{\mathrm{env}})s_{n-1}(x_n)+vd_{\mathrm{env}}.
\end{equation}
By Proposition~\ref{proposition:lb-counting}, some $n\in I_m$ satisfies $s_{n-1}(x_n)\le C_{\mathrm{sep}}(p_*(m-1)/4)^{-5/2}$. Substituting this round into \eqref{eq:lb-variance-comparison} and using the bounds on $u_{\mathrm{env}}$ and $d_{\mathrm{env}}$ gives
\begin{equation}\label{eq:lb-plateau-variance}
\mathbb P\Bigl(\bar s_m>\Delta_N+Cm^{-5/2}\sqrt{\log(m+1)}
\,\Big|\,\mathcal E_\partial\Bigr)\le e^{-c(m-1)},
\qquad c=\frac{p_*}{16}.
\end{equation}

Next we convert a small value of $\bar s_m$ into a small fill distance. Let $x\in\mathcal P$ and $r=\operatorname{dist}(x,X_m)$, and let $z$ be a point of $\mathcal D_N\cap\mathcal P$ with $|z-x|\le\gamma_N$, which exists because $1\in\mathcal D_N$. If $r>0$ and $r\ge2\gamma_N$, then $\operatorname{dist}(z,X_m)\ge r/2$, so the interval $(z-r/8,z+r/8)$ contains no observation, and Proposition~\ref{proposition:lb-residual} with $h=r/2$ gives $\bar s_m\ge s_m(z)\ge c_\psi(r/2)^{5/2}$. In every case $r\le2(\bar s_m/c_\psi)^{2/5}+2\gamma_N$, and taking the supremum over $x\in\mathcal P$ yields
\begin{equation}\label{eq:lb-fill-from-variance}
\rho_m\le2(\bar s_m/c_\psi)^{2/5}+2\gamma_N.
\end{equation}
On the complement of the event in \eqref{eq:lb-plateau-variance}, the height condition $\Delta_N\le N^{-5/2}\sqrt{\log N}$ and the monotonicity of $t\mapsto t^{-5/2}\sqrt{\log t}$ on $[2,\infty)$ give $\bar s_m\le Cm^{-5/2}\sqrt{\log(m+1)}$ for $2\le m\le N$. Moreover, $2\gamma_N\le2C_\gamma N^{-5/4}\le2C_\gamma m^{-1}$. Inserting both bounds into \eqref{eq:lb-fill-from-variance} proves \eqref{eq:lb-actual-coverage} for a suitable constant $C_H$.

\runinlabel{Step 2. Comparison points on a covered plateau.}
This step is deterministic. Fix a round $n$ and a history before round $n$ such that $\rho_{n-1}\le H$ for some $0<H\le1/80$. We construct an integer $K$, points $z_1,\ldots,z_K\in\mathcal D_N\cap\mathcal P$, and scales $h_1,\ldots,h_K$, all determined by the history, with the following three properties for all $N\ge N_1$, where $N_1$ depends only on $C_\gamma$.
\begin{enumerate}
\item[(a)] The number of points satisfies $1/(32H)\le K\le n$, and the scales satisfy $1/(64n)\le h_i\le2H$.
\item[(b)] The normalized covariance matrix $\Sigma=\bigl[h_i^{-5/2}h_j^{-5/2}c_{n-1}(z_i,z_j)\bigr]_{i,j=1}^K$ satisfies
\begin{equation}\label{eq:lb-normalized-spectrum}
c_0I_K\preceq\Sigma\preceq C_0I_K,\qquad
c_0=C_\psi^{-1},\quad C_0=\frac{512}{3}.
\end{equation}
\item[(c)] The posterior mean satisfies $|\mu_{n-1}(z_i)|\le C_\mu Bh_i^{5/2}$ with $C_\mu=2^{5/2}\cdot4/\sqrt3$.
\end{enumerate}

\emph{Construction.} Let $y_1<\cdots<y_\ell$ be the distinct observations in $\mathcal P$, and let $d_j=y_{j+1}-y_j$ be the gaps between consecutive ones. The points $3/4+H$ and $1$ and the midpoints of the gaps belong to $\mathcal P$, so each of them lies within distance $H$ of an observation. Since no observation lies in the interior of a gap, this implies
\[
y_1\le\frac34+2H,\qquad y_\ell\ge1-H,\qquad d_j\le2H,
\qquad\text{and hence}\qquad\sum_jd_j=y_\ell-y_1\ge\frac14-3H.
\]
Group the gaps into consecutive pairs $(d_1,d_2),(d_3,d_4),\ldots$ and discard the last gap if their number is odd. For each pair, let $\mathcal I_i$ be the closed interval formed by its two gaps, let $(a_i,b_i)$ be the longer of the two gaps, and let $h_i=b_i-a_i$ be its length. Then $\mathcal I_i$ contains three distinct observations and has length at most $2h_i$, the intervals $\mathcal I_i$ have disjoint interiors, and the selected gaps have total length at least $\frac12(\frac14-5H)$. Finally, discard the pairs with $h_i<1/(64n)$. There are fewer than $n$ pairs, so the discarded gaps have total length less than $1/64$, and since $H\le1/80$ the remaining gaps have total length at least $\frac12(\frac14-5H)-\frac1{64}\ge\frac5{64}>\frac1{16}$.
Relabel the remaining pairs by $i=1,\ldots,K$. Each $h_i$ is at most $2H$, which gives $K\ge1/(32H)$, and $K\le n$ is clear. This is property (a). Let $z_i$ be a point of $\mathcal D_N$ nearest to the midpoint of $(a_i,b_i)$, so that $z_i$ lies within distance $\gamma_N/2$ of this midpoint. Since $\gamma_N\le C_\gamma N^{-5/4}$ and $h_i\ge1/(64N)$, we have $\gamma_N\le h_i/2$ for all $N\ge N_1:=(128C_\gamma)^4$. The point $z_i$ then lies in the gap $(a_i,b_i)$ at distance at least $h_i/4$ from its endpoints. In particular, $z_i\in\mathcal D_N\cap\mathcal P\cap\mathcal I_i$, and the support of the bump $u_{h_i,z_i}$ is contained in $(a_i,b_i)$, which contains no observation.

\emph{Covariance bounds.} For $w\in\mathbb R^K$, let $r_w=\sum_{i=1}^Kw_ih_i^{-5/2}(I-P_{n-1})k_{z_i}\in\mathcal R_{n-1}$. The definition \eqref{eq:lb-derivative-uncertainty} of the posterior covariance and the reproducing property give
\begin{equation}\label{eq:lb-quadratic-form}
w^\top\Sigma w=\|r_w\|_{\mathcal H_k}^2,\qquad
\langle r_w,u\rangle_{\mathcal H_k}
=\sum_{i=1}^K\frac{w_iu(z_i)}{h_i^{5/2}}
\quad\text{for }u\in\mathcal R_{n-1}.
\end{equation}
For the lower bound, consider the test function $u_w=\sum_iw_iu_{h_i,z_i}$. Its bumps have disjoint supports inside gaps without observations, so $u_w\in\mathcal R_{n-1}$ and $u_w(z_i)=w_ih_i^{5/2}$, and Proposition~\ref{proposition:lb-test-functions} gives $\|u_w\|_{\mathcal H_k}^2\le C_\psi\|w\|_2^2$. Hence $\|w\|_2^2=\langle r_w,u_w\rangle_{\mathcal H_k}\le C_\psi^{1/2}\|r_w\|_{\mathcal H_k}\|w\|_2$, which is the lower bound in \eqref{eq:lb-normalized-spectrum}. For the upper bound, every $u\in\mathcal R_{n-1}$ vanishes at the three observations in $\mathcal I_i$, and $z_i\in\mathcal I_i$. Proposition~\ref{proposition:lb-three-zeros} on the interval $\mathcal I_i$ of length at most $2h_i$, the disjointness of the interiors, and \eqref{eq:lb-third-derivative} give
\begin{equation}\label{eq:lb-patch-energy}
\sum_{i=1}^K\frac{|u(z_i)|^2}{h_i^5}
\le32\sum_{i=1}^K\|u'''\|_{L^2(\mathcal I_i)}^2
\le32\|u'''\|_{L^2(0,1)}^2
\le\frac{512}{3}\|u\|_{\mathcal H_k}^2.
\end{equation}
Taking $u=r_w$ in \eqref{eq:lb-quadratic-form} and applying the Cauchy-Schwarz inequality in $\mathbb R^K$ and then \eqref{eq:lb-patch-energy} gives
\[
\|r_w\|_{\mathcal H_k}^2
\le\|w\|_2\Bigl(\sum_{i=1}^K\frac{|r_w(z_i)|^2}{h_i^5}\Bigr)^{1/2}
\le\Bigl(\frac{512}{3}\Bigr)^{1/2}\|w\|_2\|r_w\|_{\mathcal H_k},
\]
which is the upper bound in \eqref{eq:lb-normalized-spectrum}. This proves property (b).

\emph{Mean bound.} The posterior mean interpolates the observations, and $f_N$ vanishes on $\mathcal P$. Hence $\mu_{n-1}$ vanishes at the three observations in $\mathcal I_i$. Proposition~\ref{proposition:lb-three-zeros}, the bound \eqref{eq:lb-third-derivative}, and $\|\mu_{n-1}\|_{\mathcal H_k}\le B$ give $|\mu_{n-1}(z_i)|\le(2h_i)^{5/2}(4/\sqrt3)B$, which is property (c).

\runinlabel{Step 3. The maximum in one round and the sum over rounds.}
Define the coverage event $\mathcal E^{\mathrm{cov}}_{n-1}=\{\rho_{n-1}\le H_{n-1}\}\in\mathcal F_{n-1}$ and the constant $A_0=(C_\mu B/v+\sqrt{8C_0})/\sqrt{c_0}$. Choose the integer $m_*\ge2$ so large that every $n>m_*$ satisfies $H_{n-1}\le1/80$ and
\begin{equation}\label{eq:lb-K-range}
\frac{1}{32H_{n-1}}\ge n^{1/2}\ge3,\qquad
\sqrt{\tfrac12\log n}-A_0\ge\tfrac18\sqrt{\log n}.
\end{equation}
This is possible because $1/(32H_{n-1})$ grows like $n/(\log n)^{1/5}$. Let $n>m_*$ and consider a history in $\mathcal E^{\mathrm{cov}}_{n-1}$. Step 2 with $H=H_{n-1}$ provides $K$, $z_i$, and $h_i$ with $n^{1/2}\le K\le n$. Conditionally on $\mathcal F_{n-1}$, the vector $V=(Z_n(z_i)/h_i^{5/2})_{1\le i\le K}$ is centered Gaussian with covariance matrix $\Sigma$, because the points and the scales are determined by the history. By property (b) and Proposition~\ref{proposition:lb-gaussian-max}, outside an event of conditional probability at most $q_n:=e^{-n^{1/2}/2}+\exp\bigl(-\frac{c_{\mathrm{tail}}}{2}n^{1/4}/\sqrt{\log n}\bigr)$, some index $i$ satisfies $V_i>\sqrt{c_0\log K}-\sqrt{8C_0}$. Here we used $K\ge n^{1/2}$ and the fact that $K\mapsto K/\log K$ is increasing on $[3,\infty)$. For this index, properties (c) and (a) and \eqref{eq:lb-K-range} give
\[
G_n(z_i)=\mu_{n-1}(z_i)+vh_i^{5/2}V_i
>v\sqrt{c_0}\,h_i^{5/2}\bigl(\sqrt{\log K}-A_0\bigr)
\ge\frac{v\sqrt{c_0}}{8}(64n)^{-5/2}\sqrt{\log n}
\ge\beta_n,
\]
where $\beta_n=b_*n^{-5/2}\sqrt{\log n}$ with $b_*=\min\{1,2^{-18}v\sqrt{c_0}\}$. Since $z_i\in\mathcal D_N\cap\mathcal P$, the failure event $\mathcal B_n=\{\max_{z\in\mathcal D_N\cap\mathcal P}G_n(z)\le\beta_n\}$ satisfies $\mathbb P(\mathcal B_n\mid\mathcal F_{n-1})\le q_n$ on $\mathcal E^{\mathrm{cov}}_{n-1}$, for every $n>m_*$. Proposition~\ref{proposition:lb-counting} with $A=\mathcal E^{\mathrm{cov}}_{n-1}$ and the coverage bound \eqref{eq:lb-actual-coverage} then give
\[
\mathbb P(\mathcal B_n\mid\mathcal E_\partial)
\le\mathbb P\bigl((\mathcal E^{\mathrm{cov}}_{n-1})^c\mid\mathcal E_\partial\bigr)
+\mathbb P\bigl(\mathcal E^{\mathrm{cov}}_{n-1}\cap\mathcal B_n\mid\mathcal E_\partial\bigr)
\le e^{-c(n-2)}+q_n.
\]
The right side is summable in $n$. Enlarging $m_*$ so that $\sum_{n>m_*}(e^{-c(n-2)}+q_n)\le1/8$ and taking a union bound over $m_*<n\le N$ proves \eqref{eq:lower-plateau-event} for all $N\ge\max\{N_1,m_*+1\}$.
\end{proof}

\subsection{Lemma~\ref{lemma:lower-local}: Local Observations and Curvature}\label{app:lemma-lower-local}
For $0<h<1/8$, let $\tau_h$ denote the first round $n$ such that the queries $x_1,\ldots,x_n$ include two distinct points of $(0,h)$, with $\tau_h=\infty$ if there is no such round. Thus $\tau_N^{\mathrm{loc}}=\tau_{h_N}$. The event $\{\tau_h>n-1\}$, on which the history before round $n$ contains at most one point of $(0,h)$, belongs to $\mathcal F_{n-1}$.

\begin{proof}
We carry out the four steps in turn.

\runinlabel{Step 1. A test point with a large sampled value.}
Fix $h\in(0,h_0]$, where $h_0<1/8$ is a constant that will be reduced a finite number of times below, and assume that
\begin{equation}\label{eq:lb-local-assumptions}
\Delta_N\le\frac{h^2}{4}\qquad\text{and}\qquad\gamma_N\le\frac{h}{128}.
\end{equation}
Let $n\ge2$ and consider a history in $\{\tau_h>n-1\}$. At least two of the three intervals $(h/8,h/4)$, $(3h/8,h/2)$, and $(5h/8,3h/4)$ contain no observation. Let $z_n$ be a point of $\mathcal D_N$ nearest to the midpoint of the first such interval. This point is determined by the history. By \eqref{eq:lb-local-assumptions}, it lies at distance at least $h/16-h/256>h/64$ from the endpoints of the chosen interval, so the interval $(z_n-h/64,z_n+h/64)$ contains no observation. Proposition~\ref{proposition:lb-residual} with the scale $h/16$ gives $s_{n-1}(z_n)\ge c_bh^{5/2}$ with $c_b=2^{-10}c_\psi$. Moreover, $z_n\in(0,h)\subset(0,1/8)$, where $f_N(x)=\kappa(2\varepsilon_Nx-x^2)$. Hence $f_N(z_n)\ge-\kappa h^2$, and \eqref{eq:lb-mean-bounds} gives $\mu_{n-1}(z_n)\ge-\kappa h^2-Bs_{n-1}(z_n)$. By the marginal \eqref{eq:lb-past-marginal}, the standardized threshold of the event $\{G_n(z_n)\ge h^2\}$ is at most
\[
\frac{(1+\kappa)h^2+Bs_{n-1}(z_n)}{vs_{n-1}(z_n)}
\le t_h:=\frac Bv+\frac{1+\kappa}{vc_b\sqrt h},
\]
and therefore
\begin{equation}\label{eq:lb-local-excursion}
\mathbb P(G_n(z_n)\ge h^2\mid\mathcal F_{n-1})\ge p_h:=1-\Phi(t_h)
\qquad\text{on }\{\tau_h>n-1\}.
\end{equation}
For small $h$ we have $t_h\ge1$ and $t_h^2\le C/h$. Proposition~\ref{proposition:lb-gaussian-max} then gives, after reducing $h_0$,
\begin{equation}\label{eq:lb-local-excursion-probability}
\log\frac1{p_h}\le\log\frac{t_h}{c_{\mathrm{tail}}}+\frac{t_h^2}{2}
\le\frac{C_p}{h},\qquad0<h\le h_0.
\end{equation}
On the complement of $\{\tau_h>n-1\}$, where no test point is needed, we set $z_n=0$.

\runinlabel{Step 2. A large sampled value forces a large innovation.}
Consider the envelope event $\mathcal E^{\mathrm{env}}_n(J_h,p_h)$ with $J_h=\lceil h^{-3}\rceil$, and abbreviate its coefficients by $u_{\mathrm{env}}$ and $d_{\mathrm{env}}$. By \eqref{eq:lb-envelope-coefficients} and \eqref{eq:lb-local-excursion-probability},
\[
u_{\mathrm{env}}^2=2\log\bigl(8(J_h+1)\bigr)+2\log\frac1{p_h}\le\frac{C_u^2}{h},
\qquad
d_{\mathrm{env}}\le L_kh^3\bigl[2u_{\mathrm{env}}+C_2(1,1)\bigr]\le C_dh^{5/2}.
\]
Reduce $h_0$ so that $vC_dh_0^{1/2}\le1/4$, which gives $vd_{\mathrm{env}}\le h^2/4$. Define the event $\mathcal S_{n,h}=\{G_n(z_n)\ge h^2\}\cap\mathcal E^{\mathrm{env}}_n(J_h,p_h)$. By \eqref{eq:lb-local-excursion} and Proposition~\ref{proposition:lb-envelope},
\begin{equation}\label{eq:lb-local-joint-success}
\mathbb P(\mathcal S_{n,h}\mid\mathcal F_{n-1})\ge p_h-\frac{p_h}{2}=\frac{p_h}{2}
\qquad\text{on }\{\tau_h>n-1\}.
\end{equation}
On $\mathcal S_{n,h}\cap\{\tau_h>n-1\}$, Proposition~\ref{proposition:lb-envelope} with $z=z_n$ and $t=h^2$ and the assumption \eqref{eq:lb-local-assumptions} give
\[
h^2\le\Delta_N+(B+vu_{\mathrm{env}})s_{n-1}(x_n)+vd_{\mathrm{env}}
\le\frac{h^2}{2}+(B+vC_u)h^{-1/2}s_{n-1}(x_n),
\]
that is,
\begin{equation}\label{eq:lb-local-innovation}
s_{n-1}(x_n)\ge c_*h^{5/2},\qquad c_*=\frac{1}{2(B+vC_u)}.
\end{equation}
This conclusion holds wherever the query $x_n$ lies. By Proposition~\ref{proposition:lb-counting}, every trajectory satisfies
\begin{equation}\label{eq:lb-large-innovation-count}
\#\{1\le n\le N:s_{n-1}(x_n)\ge c_*h^{5/2}\}\le D_0h^{-1},
\qquad D_0=\Bigl(\frac{C_{\mathrm{sep}}}{c_*}\Bigr)^{2/5}.
\end{equation}

\runinlabel{Step 3. Arrival of two local observations.}
For $2\le n\le N$, define the indicator
\begin{equation}\label{eq:lb-stopped-indicator}
Q_n^{\mathrm{loc}}=
\begin{cases}
\mathbf 1_{\mathcal S_{n,h}}&\text{if }\tau_h>n-1,\\
1&\text{if }\tau_h\le n-1.
\end{cases}
\end{equation}
The case distinction is determined by $\mathcal F_{n-1}$, and \eqref{eq:lb-local-joint-success} gives $\mathbb E(Q_n^{\mathrm{loc}}\mid\mathcal F_{n-1})\ge p_h/2$ in both cases. Let $2\le m\le N$. On the event $\{\tau_h>m\}$, every round $2\le n\le m$ with $Q_n^{\mathrm{loc}}=1$ satisfies \eqref{eq:lb-local-innovation}, so $\sum_{n=2}^mQ_n^{\mathrm{loc}}\le D_0h^{-1}$ by \eqref{eq:lb-large-innovation-count}. If
\begin{equation}\label{eq:lb-local-budget-condition}
p_h(m-1)>4D_0h^{-1},
\end{equation}
then $\{\tau_h>m\}$ is contained in the event $\{\sum_{n=2}^mQ_n^{\mathrm{loc}}<p_h(m-1)/4\}$, and Proposition~\ref{proposition:lb-counting} with $p=p_h/2$ yields
\begin{equation}\label{eq:lb-local-hitting-time}
\mathbb P(\tau_h>m\mid\mathcal E_\partial)\le\exp\Bigl(-\frac{p_h(m-1)}{16}\Bigr).
\end{equation}
Now choose $A_{\mathrm{loc}}=\max\{16C_p,1\}$ and let $h_N$ and $m_N$ be as in \eqref{eq:lower-local-scales}. For all sufficiently large $N$, we have $h_N\le h_0$, and the assumptions \eqref{eq:lb-local-assumptions} hold with $h=h_N$ because $\Delta_N\le N^{-5/2}\sqrt{\log N}$ and $\gamma_N\le C_\gamma N^{-5/4}$. Moreover, \eqref{eq:lb-local-excursion-probability} gives $p_{h_N}\ge N^{-C_p/A_{\mathrm{loc}}}\ge N^{-1/16}$, hence $p_{h_N}(m_N-1)\ge\tfrac12N^{3/16}>4D_0\log N/A_{\mathrm{loc}}=4D_0h_N^{-1}$.
The condition \eqref{eq:lb-local-budget-condition} therefore holds for $h=h_N$ and $m=m_N$, and \eqref{eq:lb-local-hitting-time} gives \eqref{eq:lower-local-visits} with $c=1/32$.

\runinlabel{Step 4. Curvature bounds.}
Since $G_n=\mu_{n-1}+vZ_n$ and $\|\mu_{n-1}''\|_\infty\le B$ by \eqref{eq:lb-mean-bounds}, it suffices to bound the second derivative of $Z_n$. Proposition~\ref{proposition:lb-derivative-supremum}, applied to the conditional law of $Z_n$ given $\mathcal F_{n-1}$ with $\delta=N^{-3}$, gives $\mathbb P\bigl(\|G_n''\|_\infty>B+vC_{\sup}(1+\sqrt{3\log N})\mid\mathcal F_{n-1}\bigr)\le N^{-3}$ for every $n$.
Choose $C_g$ such that $B+vC_{\sup}(1+\sqrt{3\log N})\le C_g\sqrt{\log N}$ for all $N\ge2$. Proposition~\ref{proposition:lb-counting}, with $A$ equal to the whole probability space, and a union bound over $2\le n\le N$ prove \eqref{eq:lower-global-curvature}.

For the local bound, let $n>m_N$ and consider a history in $\mathcal E_\partial\cap\{\tau_N^{\mathrm{loc}}\le m_N\}$. The interval $[0,h_N]$ then contains three distinct observations, namely zero and two points of $(0,h_N)$. The local bound of Proposition~\ref{proposition:lb-derivative-supremum} with $\delta=N^{-3}$ and the identity $\sqrt{h_N}(1+\sqrt{3\log N})=\sqrt{A_{\mathrm{loc}}}(1/\sqrt{\log N}+\sqrt3)$ give
\[
\mathbb P\bigl(\|G_n''\|_{\infty,[0,h_N]}>L\mid\mathcal F_{n-1}\bigr)\le N^{-3},
\qquad
L:=\max\Bigl\{1,\ \kappa,\ B+vC_{\sup}\sqrt{A_{\mathrm{loc}}}
\Bigl(\frac{1}{\sqrt{\log2}}+\sqrt3\Bigr)\Bigr\}.
\]
Since the event $\{\tau_N^{\mathrm{loc}}\le m_N\}$ belongs to $\mathcal F_{n-1}$ for $n>m_N$, Proposition~\ref{proposition:lb-counting} with $A=\{\tau_N^{\mathrm{loc}}\le m_N\}$ and a union bound over $m_N<n\le N$ prove \eqref{eq:lower-local-curvature}.
\end{proof}

\subsection{Common Steps for Theorem~\ref{thm:matern-log-lower}}\label{app:lb-common}
This subsection shows that, with probability bounded away from zero, no query falls in the neighborhood $\mathcal A_N$ of the maximizer defined in \eqref{eq:lower-near-optimal-set}. Proposition~\ref{proposition:lb-first-order} proves the comparison \eqref{eq:lower-discrete-comparison}, Proposition~\ref{proposition:lb-initial-rounds} handles the initial rounds, and Proposition~\ref{proposition:lb-avoidance} combines both with Lemmas~\ref{lemma:lower-plateau} and~\ref{lemma:lower-local}. Recall the convention $\gamma_N=0$ for the continuous implementation.

\begin{lbproposition}[First order comparison]\label{proposition:lb-first-order}
Let $g$ be twice continuously differentiable on $[0,1]$, and let $x\in\mathcal D_N$ with $0<x<1$ maximize $g$ over $\mathcal D_N$. If $|g''|\le\Lambda$ on $[0,x+\gamma_N]$ for some $\Lambda\ge0$, then
\begin{equation}\label{eq:lb-grid-stationarity}
|g'(x)|\le\tfrac12\Lambda\gamma_N
\qquad\text{and}\qquad
|g'(0)|\le\Lambda\bigl(x+\tfrac12\gamma_N\bigr).
\end{equation}
If, in addition, $g(0)=0$, then
\begin{equation}\label{eq:lb-anchor-comparison}
g(x)\le\tfrac12\Lambda\bigl(x^2+\gamma_Nx\bigr).
\end{equation}
\end{lbproposition}
\begin{proof}
For the continuous implementation, $x$ is an interior maximizer, so $g'(x)=0$. For the discrete implementation, the neighbors $x-\gamma_N$ and $x+\gamma_N$ belong to the grid, and $g$ is at most $g(x)$ at both. Taylor's formula gives $g(x\pm\gamma_N)\ge g(x)\pm\gamma_Ng'(x)-\frac12\Lambda\gamma_N^2$, hence $\pm g'(x)\le\frac12\Lambda\gamma_N$. In both cases the first bound in \eqref{eq:lb-grid-stationarity} holds, and the second bound follows from $|g'(0)-g'(x)|\le\Lambda x$. If $g(0)=0$, Taylor's formula at $x$ gives $0=g(0)\ge g(x)-xg'(x)-\frac12\Lambda x^2$, so $g(x)\le x|g'(x)|+\frac12\Lambda x^2$, which is \eqref{eq:lb-anchor-comparison}.
\end{proof}

A positive query $x_n$ close to zero forces $|G_n'(0)|$ to be small by Proposition~\ref{proposition:lb-first-order}, whereas the conditional variance of $G_n'(0)$ is bounded below as long as no previous query lies close to zero. A sequence of shrinking radii makes these two facts compatible over a fixed number of rounds.

\begin{lbproposition}[Initial rounds]\label{proposition:lb-initial-rounds}
Let $m_*\ge2$ be a fixed integer. Under the assumptions of Lemma~\ref{lemma:lower-local},
\begin{equation}\label{eq:lb-prefix-avoidance}
\mathbb P\bigl(0<x_n<N^{-1/4}\text{ for some }2\le n\le m_*
\,\big|\,\mathcal E_\partial\bigr)\to 0
\qquad\text{as }N\to\infty.
\end{equation}
\end{lbproposition}
\begin{proof}
For $1\le n\le m_*$, define the radii $R_n=N^{-2^{\,n-m_*-2}}$. They decrease in $n$ and satisfy $R_n=R_{n-1}^2$ and $R_{m_*}=N^{-1/4}$. For $2\le n\le m_*$, let
\[
\mathcal E^{\mathrm{far}}_{n-1}=\bigcap_{j=2}^{n-1}\{x_j\notin(0,R_j)\},\qquad
F_n=\mathcal E^{\mathrm{far}}_{n-1}\cap\{0<x_n<R_n\},
\]
where the empty intersection is the whole probability space. Thus $F_n$ is the event that round $n$ is the first round with a positive query below its radius. Let also $\mathcal E^{\mathrm{glob}}_N=\bigl\{\max_{2\le n\le N}\|G_n''\|_\infty\le C_g\sqrt{\log N}\bigr\}$ be the event of the global curvature bound \eqref{eq:lower-global-curvature}.

On $\mathcal E_\partial\cap\mathcal E^{\mathrm{far}}_{n-1}$, the history before round $n$ contains no point of $(0,R_{n-1})$, because $R_j\ge R_{n-1}$ for $j\le n-1$. Propositions~\ref{proposition:lb-residual} and~\ref{proposition:lb-paths} give $\operatorname{Var}(G_n'(0)\mid\mathcal F_{n-1})=v^2s_{n-1}^{(1)}(0)^2\ge v^2c_\theta^2R_{n-1}^3$. The conditional density of the Gaussian variable $G_n'(0)$ is therefore at most $(2\pi)^{-1/2}(vc_\theta)^{-1}R_{n-1}^{-3/2}$, whatever its mean, and
\begin{equation}\label{eq:lb-derivative-density}
\mathbb P\bigl(|G_n'(0)|\le t\mid\mathcal F_{n-1}\bigr)\le Ct\,R_{n-1}^{-3/2},
\qquad t>0.
\end{equation}
On the other hand, on $\{0<x_n<R_n\}\cap\mathcal E^{\mathrm{glob}}_N$, Proposition~\ref{proposition:lb-first-order} with $\Lambda=C_g\sqrt{\log N}$ gives $|G_n'(0)|\le C_g\sqrt{\log N}\bigl(R_n+\tfrac12\gamma_N\bigr)\le2C_g\sqrt{\log N}\,R_n$, where the last step uses $\gamma_N\le C_\gamma N^{-5/4}\le N^{-1/4}\le R_n$ for large $N$. Hence $F_n\cap\mathcal E^{\mathrm{glob}}_N\subseteq\mathcal E^{\mathrm{far}}_{n-1}\cap\bigl\{|G_n'(0)|\le2C_g\sqrt{\log N}\,R_n\bigr\}$, and Proposition~\ref{proposition:lb-counting} with $A=\mathcal E^{\mathrm{far}}_{n-1}$ and \eqref{eq:lb-derivative-density} give
\begin{equation}\label{eq:lb-prefix-density}
\mathbb P\bigl(F_n\cap\mathcal E^{\mathrm{glob}}_N\mid\mathcal E_\partial\bigr)
\le C\sqrt{\log N}\,R_nR_{n-1}^{-3/2}
=C\sqrt{\log N}\,R_{n-1}^{1/2}
\le C\sqrt{\log N}\,N^{-2^{-m_*-2}}.
\end{equation}
If some $2\le n\le m_*$ satisfies $0<x_n<R_n$, then the first such round belongs to $F_n$. A union bound and \eqref{eq:lower-global-curvature} therefore give
\[
\mathbb P\bigl(0<x_n<R_n\text{ for some }2\le n\le m_*\mid\mathcal E_\partial\bigr)
\le N^{-2}+Cm_*\sqrt{\log N}\,N^{-2^{-m_*-2}},
\]
which tends to zero because $m_*$ is fixed. Since $R_n\ge N^{-1/4}$ for every $n\le m_*$, this proves \eqref{eq:lb-prefix-avoidance}.
\end{proof}

\begin{lbproposition}[No query near the maximizer]\label{proposition:lb-avoidance}
Let $b_*$ and $m_*$ be the constants of Lemma~\ref{lemma:lower-plateau}, let $L$ be the constant of Lemma~\ref{lemma:lower-local}, and let $\varepsilon_N$, $f_N$, and $\mathcal A_N$ be defined by \eqref{eq:lower-epsilon-choice} and \eqref{eq:lower-near-optimal-set}. For both implementations and all sufficiently large $N$,
\begin{equation}\label{eq:lb-avoidance}
\mathbb P\bigl(x_n\notin\mathcal A_N\text{ for every }1\le n\le N\bigr)
\ge\frac{q_\partial}{2},
\end{equation}
and on this event $r_N^{\mathrm{best}}\ge\kappa\varepsilon_N^2/4$.
\end{lbproposition}
\begin{proof}
We record the orders of the parameters and the objective gap, define a good event, and show that it excludes every query from $\mathcal A_N$.

\runinlabel{Parameters.}
By \eqref{eq:lower-epsilon-choice}, $\varepsilon_N^2=\beta_N/(16L)$ with $\beta_N=b_*N^{-5/2}\sqrt{\log N}$. Since $b_*\le1$ and $L\ge\max\{1,\kappa\}$, we have $\Delta_N=\kappa\varepsilon_N^2\le\beta_N/16\le N^{-5/2}\sqrt{\log N}$, so both lemmas apply to $f_N$. Moreover, with $h_N$ from \eqref{eq:lower-local-scales},
\begin{equation}\label{eq:lb-epsilon-mesh-scales}
\varepsilon_N=\Bigl(\frac{b_*}{16L}\Bigr)^{1/2}N^{-5/4}(\log N)^{1/4},
\qquad
\frac{\gamma_N}{\varepsilon_N}\to0,\qquad
\frac{\varepsilon_N}{h_N}\to0,\qquad
\varepsilon_NN^{1/4}\to0.
\end{equation}
In particular, for all sufficiently large $N$ we have $\varepsilon_N<1/16$, $\mathcal A_N\subset(0,N^{-1/4})$, and $\frac32\varepsilon_N+\gamma_N\le h_N$.

\runinlabel{Objective gap.}
On $[0,1/8]$ we have $f_N^*-f_N(x)=\kappa(x-\varepsilon_N)^2$, which is at least $\kappa\varepsilon_N^2/4$ outside $\mathcal A_N$, and on $[1/8,1]$ we have $f_N\le0$. Hence $f_N^*-f_N(x)\ge\kappa\varepsilon_N^2/4$ for every $x\in[0,1]\setminus\mathcal A_N$, which proves the last claim of the proposition once \eqref{eq:lb-avoidance} is established.

\runinlabel{Good event.}
Let $C_g$, $h_N$, $m_N$, and $\tau_N^{\mathrm{loc}}$ be as in Lemma~\ref{lemma:lower-local}. Consider the five events
\begin{align*}
\mathcal E^{\mathrm{init}}_N&=\{x_n\notin\mathcal A_N\text{ for every }2\le n\le m_*\},\\
\mathcal E^{\mathrm{over}}_N&=\Bigl\{\max_{z\in\mathcal D_N\cap\mathcal P}G_n(z)>\beta_n\text{ for every }m_*<n\le N\Bigr\},\\
\mathcal E^{\mathrm{glob}}_N&=\Bigl\{\max_{2\le n\le N}\|G_n''\|_\infty\le C_g\sqrt{\log N}\Bigr\},\qquad \mathcal E^{\mathrm{loc}}_N=\{\tau_N^{\mathrm{loc}}\le m_N\},\\
\mathcal E^{\mathrm{curv}}_N&=\Bigl\{\max_{m_N<n\le N}\|G_n''\|_{\infty,[0,h_N]}\le L\Bigr\},
\end{align*}
and let $\mathcal E^{\mathrm{good}}_N$ be their intersection. Proposition~\ref{proposition:lb-initial-rounds} together with $\mathcal A_N\subset(0,N^{-1/4})$, Lemma~\ref{lemma:lower-plateau}, and the three bounds of Lemma~\ref{lemma:lower-local} give
\begin{equation}\label{eq:lb-joint-trajectory}
\begin{aligned}
\mathbb P\bigl((\mathcal E^{\mathrm{good}}_N)^c\mid\mathcal E_\partial\bigr)
&\le\mathbb P\bigl((\mathcal E^{\mathrm{init}}_N)^c\mid\mathcal E_\partial\bigr)
+\mathbb P\bigl((\mathcal E^{\mathrm{over}}_N)^c\mid\mathcal E_\partial\bigr)
+\mathbb P\bigl((\mathcal E^{\mathrm{glob}}_N)^c\mid\mathcal E_\partial\bigr)\\
&\quad+\mathbb P\bigl((\mathcal E^{\mathrm{loc}}_N)^c\mid\mathcal E_\partial\bigr)
+\mathbb P\bigl(\mathcal E^{\mathrm{loc}}_N\cap(\mathcal E^{\mathrm{curv}}_N)^c\mid\mathcal E_\partial\bigr)\\
&\le o(1)+\frac18+N^{-2}+e^{-cN^{3/16}}+N^{-2}<\frac12
\end{aligned}
\end{equation}
for all sufficiently large $N$. Since $\mathbb P(\mathcal E_\partial)\ge q_\partial$ by Proposition~\ref{proposition:lb-boundary}, the event $\mathcal E_\partial\cap\mathcal E^{\mathrm{good}}_N$ has probability at least $q_\partial/2$. It remains to show that no query lies in $\mathcal A_N$ on this event.

\runinlabel{Exclusion of all queries.}
Work on $\mathcal E_\partial\cap\mathcal E^{\mathrm{good}}_N$. The first query is $x_1=0\notin\mathcal A_N$, and the rounds $2\le n\le m_*$ are covered by $\mathcal E^{\mathrm{init}}_N$. Let $n>m_*$ and suppose that $x_n\in\mathcal A_N$. Then $0<x_n<\frac32\varepsilon_N$, and $G_n(0)=0$ by Proposition~\ref{proposition:lb-boundary}. If $n\le m_N$, Proposition~\ref{proposition:lb-first-order} with $g=G_n$ and $\Lambda=C_g\sqrt{\log N}$ gives $G_n(x_n)\le C_g\sqrt{\log N}\,\bigl(\frac98\varepsilon_N^2+\frac34\gamma_N\varepsilon_N\bigr)\le CN^{-5/2}\log N$.
The levels $\beta_n$ decrease in $n\ge2$, so $\beta_n\ge\beta_{m_N}\ge cN^{-5/8}\sqrt{\log N}$ in this range, and therefore $G_n(x_n)<\beta_n$ for all sufficiently large $N$. If $n>m_N$, the interval $[0,x_n+\gamma_N]$ is contained in $[0,h_N]$, and Proposition~\ref{proposition:lb-first-order} with $\Lambda=L$ gives
\begin{equation}\label{eq:lb-late-exclusion}
G_n(x_n)\le L\Bigl(\frac98\varepsilon_N^2+\frac34\gamma_N\varepsilon_N\Bigr)
=\beta_N\Bigl(\frac{9}{128}+\frac{3\gamma_N}{64\varepsilon_N}\Bigr)
<\beta_N\le\beta_n
\end{equation}
for all sufficiently large $N$. In both cases $G_n(x_n)<\beta_n<\max_{z\in\mathcal D_N\cap\mathcal P}G_n(z)$ on $\mathcal E^{\mathrm{over}}_N$, which contradicts the maximality of $x_n$ over $\mathcal D_N$. Hence no query lies in $\mathcal A_N$, and \eqref{eq:lb-avoidance} follows.
\end{proof}

\subsection{Theorem~\ref{thm:matern-log-lower}: Mat\'ern Logarithmic Lower Bound}\label{app:thm-lower}
\begin{proof}
The objective $f_N$ satisfies $\|f_N\|_{\mathcal H_k}\le B/2$ by the construction of the family \eqref{eq:lower-objective-family}. For either implementation, Proposition~\ref{proposition:lb-avoidance}, with the constants $b_*$ and $L$ of Lemmas~\ref{lemma:lower-plateau} and~\ref{lemma:lower-local} for this implementation, gives, for all sufficiently large $N$,
\[
\mathbb P_{f_N}\Bigl(r_N^{\mathrm{best}}\ge\frac{\kappa\varepsilon_N^2}{4}\Bigr)
\ge\frac{q_\partial}{2},
\qquad
\frac{\kappa\varepsilon_N^2}{4}=\frac{\kappa b_*}{64L}N^{-5/2}\sqrt{\log N}.
\]
This is \eqref{eq:matern-log-lower-hp} with $c_{\mathrm{lb}}=\kappa b_*/(64L)$ and $q_{\mathrm{lb}}=q_\partial/2$. Since the regret is nonnegative, the worst-case expected regret is at least $c_{\mathrm{lb}}q_{\mathrm{lb}}N^{-5/2}\sqrt{\log N}$. For the continuous implementation, the matching upper bound is Corollary~\ref{corollary:Matern-continuous} with $\nu=5/2$ and $d=1$.

For the discrete implementation, we first verify \eqref{eq:lower-grid-bias}. Let $\|f\|_{\mathcal H_k}\le B$, so that $\|f''\|_\infty\le B$ by Proposition~\ref{proposition:lb-norm}. If a boundary point maximizes $f$, it belongs to the grid and the bias vanishes. Otherwise, let $x^*$ be an interior maximizer and let $z$ be a nearest grid point, so that $f'(x^*)=0$ and $|x^*-z|\le\gamma_N/2$. Taylor's formula gives
\begin{equation}\label{eq:lb-quadratic-grid-error}
0\le b_{\mathcal G_N^{\mathrm{lb}}}(f)
\le f(x^*)-f(z)
\le\frac B2|x^*-z|^2
\le\frac B8\gamma_N^2,
\end{equation}
which is $O(N^{-5/2})$ under \eqref{eq:lower-resolution}. The grid has $M_N=J_N^{\mathrm{lb}}+1$ points with $\log M_N=O(\log N)$. We apply Theorem~\ref{thm:simple-regret-discrete-general} with $\alpha=5/2$, exact maximization, and $\delta_N=N^{-7/2}$. Its budget condition holds for all sufficiently large $N$, and $u_{N,M_N,\delta_N}^2=2\log(4NM_N/\delta_N)=O(\log N)$. Using \eqref{eq:lb-quadratic-grid-error} and the bound $r_N^{\mathrm{best}}\le2B$ on the exceptional event, as in the proof of Corollary~\ref{corollary:Matern-discrete}, we obtain
\[
\sup_{\|f\|_{\mathcal H_k}\le B}\mathbb E_f[r_N^{\mathrm{best}}]
\le\frac B8\gamma_N^2+CN^{-5/2}\sqrt{\log N}+2BN^{-7/2}
=O\bigl(N^{-5/2}\sqrt{\log N}\bigr).
\]
Together with the lower bound, this proves \eqref{eq:matern-tight-rate}. The condition on the cardinality of the grids is used only in this upper bound, whereas the lower bound uses only the condition on the mesh.
\end{proof}

\section{Proof of Corollary~\ref{corollary:Matern-capped}}\label{app:gp-cts}
\begin{proof}
Proposition~\ref{proposition:Matern-kernel-asmps} gives $\alpha=\nu/d$. For $\mathcal D=\mathcal G_N$, the bound \eqref{eq:grid-bias-size} holds uniformly over the RKHS ball, so $\overline b_{\mathcal G_N}\le BL_kN^{-\nu/d}$. For $\mathcal D=\mathcal X$, we have $\overline b_{\mathcal X}=0$, and the posterior samples have continuous paths by Appendix~\ref{app:background}. With exact maximization, \eqref{eq:cts-expected} gives in both cases
\begin{equation}\label{eq:app-cts-matern}
\sup_{\|f\|_{\mathcal H_k}\le B}\mathbb E_f[r_N^{\mathrm{best}}]
\le(BL_k+C_{\mathrm{cap}})N^{-\nu/d}+2Be^{-pN/8}
=O(N^{-\nu/d}),
\end{equation}
because $p>0$ is fixed. The same substitution covers every deterministic sequence of finite decision sets with $\overline b_{\mathcal D}=O(N^{-\nu/d})$. Finally, GP-CTS with a fixed $\overline B$ is a randomized sequential strategy that returns a queried point. The lower bound of \citet[Theorem~1]{JMLR:v12:bull11a}, recalled in Appendix~\ref{app:cor-matern-discrete}, therefore applies to it, and the rate $N^{-\nu/d}$ is minimax optimal over the fixed Mat\'ern RKHS ball.
\end{proof}

\section{Experimental Details}\label{app:experiments}

\subsection{Benchmark Objectives and Candidate Sets}\label{app:exp-objectives}
All objectives are maximized, so each benchmark $f_0$ is the negative of the usual minimization version \citep{surjanovic2013virtual}, and its original domain is mapped affinely onto $[0,1]^d$. Eggholder oscillates over the whole domain, is not differentiable everywhere, has several hundred local maxima, and attains its maximum at the boundary point $(512,404.23)$. Styblinski-Tang has one local maximum in each of the $16$ orthants, with values between $100.1$ and the global maximum $156.66$. Hartmann exceeds half of its maximum on less than $2\%$ of the domain, and its two highest local maxima, $3.32$ and $3.20$, are attained at distant points.

The discrete implementations and the normalization in Appendix~\ref{app:exp-model} use the same finite candidate set $\mathcal G\subset[0,1]^d$. It is a regular grid with $161$ points per coordinate for Eggholder, and it consists of $2^{14}$ and $2^{15}$ scrambled Sobol points \citep{sobol1967distribution} for the other two tasks, since a regular grid of comparable resolution is not feasible for $d=4$ and $d=6$. The candidate sets are the same for all methods and macro-replications, and no known maximizer is inserted. Table~\ref{tab:exp-settings} lists the resulting discretization bias $b_{\mathcal G}$ in the original units, which is below the regret levels in Figure~\ref{fig:regret}. The reference maximum $f_0^*$ over the continuous domain is computed numerically and is used only to report the regret.

\subsection{Kernel, Lengthscale, and Normalization}\label{app:exp-model}
\runinlabel{Kernel and lengthscale.}
All methods use the Mat\'ern-$5/2$ kernel with unit amplitude on the unit cube, $k_0(x,x')=(1+q+q^2/3)e^{-q}$ with $q=\sqrt5\,\|x-x'\|_2/\ell$. For each task, the lengthscale $\ell$ in Table~\ref{tab:exp-settings} is the offline maximum likelihood estimate described in Section~\ref{sec:experiments}. It is fixed before the runs and shared by the four methods, since our theoretical analysis does not cover a kernel that changes with the observations.

\runinlabel{Normalization.}
Our theoretical analysis ties $v$ and $\overline B$ to the norm bound $B$, which is not available for a benchmark objective. Adding a constant to the objective or multiplying it by a positive constant does not change the optimization problem, so we choose the offset and the scale such that the norm becomes known on a finite set. If the Gram matrix $K$ of a kernel on $\mathcal G$ is nonsingular, a function on $\mathcal G$ with the vector of values $g\in\mathbb R^M$ has the squared RKHS norm $g^\top K^{-1}g$. The Gram matrix of $k_0$ on a fine candidate set is nearly singular. For numerical stability of Cholesky decomposition in the discrete implementation, we use the kernel $k_\delta(x,x')=k_0(x,x')+\delta\,\mathbf 1\{x=x'\}$ with $\delta=10^{-10}$, which is part of the model and not observation noise. No further jitter is added during a run. Let $K$ be the Gram matrix of $k_\delta$ on $\mathcal G$, and let $f_{0,\mathcal G}$ be the vector of the values of $f_0$ on $\mathcal G$. We set
\begin{equation}\label{eq:exp-normalization}
c_{\mathcal G}=\frac{\mathbf 1^\top K^{-1}f_{0,\mathcal G}}{\mathbf 1^\top K^{-1}\mathbf 1},\qquad
a_{\mathcal G}=\sqrt{(f_{0,\mathcal G}-c_{\mathcal G}\mathbf 1)^\top K^{-1}(f_{0,\mathcal G}-c_{\mathcal G}\mathbf 1)},\qquad
f=\frac{f_0-c_{\mathcal G}}{a_{\mathcal G}}.
\end{equation}
The shift $c_{\mathcal G}$ minimizes the norm of $f_{0,\mathcal G}-c\mathbf 1$ over $c\in\mathbb R$, and $a_{\mathcal G}$ is the minimal norm. The restriction of $f$ to $\mathcal G$ therefore has unit norm in the RKHS of $k_\delta$ on $\mathcal G$. With $\mathcal G$ regarded as the domain, $f$ satisfies the norm condition of Assumption~\ref{asmp:kernel-model} with $B=1$, so $v=\overline B=1$ realizes $v=B$ together with a valid cap input, and $U_{n-1}$ dominates $f$ on $\mathcal G$ in every round. This construction has one limitation. The continuous implementations use the same $f$ together with $k_0$ on $[0,1]^d$, and the unit norm on $\mathcal G$ does not bound the norm on $[0,1]^d$, so the cap with $\overline B=1$ is not certified for the continuous implementations. However, this is still a reasonable empirical choice for practical implementations.

\begin{table}[tbp]
\centering
\small
\setlength{\tabcolsep}{2pt}
\begin{tabular}{@{}lclclccccc@{}}
\toprule
Task & $d$ & Domain & $N$
& \begin{tabular}[c]{@{}l@{}}Candidate set\\$\mathcal G$\end{tabular}
& $\ell$ & $f_0^*$ & $b_{\mathcal G}$ & $c_{\mathcal G}$ & $a_{\mathcal G}$ \\
\midrule
Eggholder & $2$ & $[-512,512]^2$ & $200$
& \begin{tabular}[c]{@{}l@{}}regular grid,\\$M=161^2$\end{tabular}
& $0.0338$ & $959.64$ & $1.20$ & $1.016$ & $5.685\times10^4$ \\
\addlinespace[2pt]
Styblinski-Tang & $4$ & $[-5,5]^4$ & $500$
& \begin{tabular}[c]{@{}l@{}}Sobol points,\\$M=2^{14}$\end{tabular}
& $2.7$ & $156.66$ & $3.38$ & $-2.075\times10^4$ & $1.375\times10^5$ \\
\addlinespace[2pt]
Hartmann & $6$ & $[0,1]^6$ & $1000$
& \begin{tabular}[c]{@{}l@{}}Sobol points,\\$M=2^{15}$\end{tabular}
& $0.38$ & $3.3224$ & $0.055$ & $0.0243$ & $16.12$ \\
\bottomrule
\end{tabular}
\caption{Settings of the three tasks. The lengthscale $\ell$ refers to the unit cube, and $c_{\mathcal G}$ and $a_{\mathcal G}$ are the normalization constants in \eqref{eq:exp-normalization}.}
\label{tab:exp-settings}
\end{table}

\subsection{Discrete Implementations}\label{app:exp-discrete}
The discrete implementations realize Algorithms~\ref{alg:gp-ts} and~\ref{alg:gp-cts} without approximation, because the model on $\mathcal G$ is a Gaussian vector of dimension $M$. The Gram matrix of $k_\delta$ on $\mathcal G$ is factorized once by a Cholesky decomposition. In every round, a fresh Gaussian vector with covariance $K$ is drawn and conditioned on the observations by the pathwise update \citep{wilson2020efficiently}, which gives an exact joint posterior draw $G_n$ on $\mathcal G$. The criterion, which is $G_n$ for GP-TS and $H_n=\min\{G_n,U_{n-1}\}$ for GP-CTS, is maximized exactly over $\mathcal G$, with ties resolved by a fixed permutation of $\mathcal G$. The selection rules \eqref{eq:TS-selection} and \eqref{eq:CTS-selection} hence hold with $\eta_n=0$.

\subsection{Continuous Implementations}\label{app:exp-continuous}
The exact continuous implementation requires a posterior sample path on all of $[0,1]^d$ and a global maximizer of the criterion, and neither can be computed exactly. The continuous implementations therefore use approximate sample paths and a multi-start local optimizer. Following the pathwise update \eqref{eq:app-exact-posterior-path}, the sample in round $n$ is
\begin{equation}\label{eq:exp-continuous-sample}
G_n(x)=\mu_{n-1}(x)+v\bigl[p_n(x)-\mathbf k_{n-1}(x)^\top(K_{n-1}+\lambda I)^{-1}p_n(X_{n-1})\bigr],
\end{equation}
where $p_n$ is an approximate prior path drawn afresh in round $n$ from $4096$ random Fourier features \citep{rahimi2007random}, and $\mu_{n-1}$ and $s_{n-1}$ are computed from \eqref{eq:gp-posterior} for the kernel $k_0$, with $K_{n-1}^\dagger$ replaced by $(K_{n-1}+\lambda I)^{-1}$ and $\lambda=10^{-10}$. Since the covariance of $p_n$ only approximates $k_0$, the function $G_n$ is an approximate posterior sample.

In every round, the criterion is evaluated at $256$ scrambled Sobol points that are fixed within a run, the $2^d$ corners of the domain, $32$ fresh Sobol points, and all previous queries. Up to four local searches with L-BFGS-B \citep{byrd1995limited} and analytic gradients start from the best of these candidates, and the query is the best point among the candidates and the end points of the searches. GP-CTS performs these searches for both $G_n$ and $H_n$, and hence uses up to eight searches per round. This procedure does not guarantee a global maximizer, which is why our theoretical analysis allows the tolerance $\eta_n$.

\subsection{Evaluation and Detailed Results}\label{app:exp-evaluation}
To compare the four methods on one scale, we measure the best-query simple regret in the original units of $f_0$ and relative to the maximum over the continuous domain, $r_n=f_0^*-\max_{1\le t\le n}f_0(x_t)$. The discrete implementations are thus charged with the discretization bias $b_{\mathcal G}$, as in \eqref{eq:ts-discrete-bound}. The bootstrap intervals in Figure~\ref{fig:regret} are obtained from $1000$ resamples of the $50$ trajectories. Table~\ref{tab:exp-results} lists the mean regret at the final budget together with its interval. The mean regret of the discrete GP-CTS is $79\%$, $50\%$, and $59\%$ of that of the discrete GP-TS on Eggholder, Styblinski-Tang, and Hartmann, and the corresponding values for the continuous implementations are $76\%$, $64\%$, and $74\%$.

\begin{table}[tbp]
\centering
\small
\begin{tabular}{@{}llcccc@{}}
\toprule
 & & \multicolumn{2}{c}{GP-TS} & \multicolumn{2}{c}{GP-CTS} \\
\cmidrule(lr){3-4}\cmidrule(l){5-6}
Task & & discrete & continuous & discrete & continuous \\
\midrule
Eggholder & mean & $124.2$ & $149.3$ & $97.9$ & $112.9$ \\
 & interval & $[108.7,\,140.9]$ & $[129.7,\,168.8]$ & $[81.4,\,115.6]$ & $[95.9,\,133.2]$ \\
\midrule
Styblinski-Tang & mean & $24.5$ & $22.4$ & $12.3$ & $14.4$ \\
 & interval & $[22.3,\,26.5]$ & $[20.6,\,24.4]$ & $[11.1,\,13.7]$ & $[12.6,\,16.2]$ \\
\midrule
Hartmann & mean & $0.619$ & $0.685$ & $0.368$ & $0.508$ \\
 & interval & $[0.568,\,0.671]$ & $[0.627,\,0.744]$ & $[0.334,\,0.402]$ & $[0.465,\,0.547]$ \\
\bottomrule
\end{tabular}
\caption{Mean best-query simple regret at the final budget, in the original units, with the pointwise $95\%$ bootstrap interval for the mean.}
\label{tab:exp-results}
\end{table}

%

\bibliographystyle{plainnat}
\bibliography{references}

\end{document}